\documentclass[letterpaper,journal]{IEEEtran}

\usepackage{amsmath,amsfonts,amssymb}
\usepackage{bm}
\usepackage{graphicx}
\usepackage[table]{xcolor}
\usepackage{booktabs}
\usepackage{multirow}
\usepackage{array}
\usepackage{capt-of}
\usepackage{algorithm}
\usepackage{algorithmicx}
\usepackage{algpseudocode}
\usepackage{stfloats}
\usepackage{textcomp}
\usepackage{url}
\usepackage{pifont}
\usepackage[normalem]{ulem}
\usepackage{cite}
\usepackage[hidelinks,hypertexnames=false]{hyperref}

\providecommand{\Description}[1]{}

\begin{document}

\title{ScaleResfusion: Residual Rectified Flow based on Residual Vector Field}

\author{Zhenning~Shi,
        Chen~Xu,
        Junhao~Zhang,
        Kefei~Zhang,
        Linjie~Liu,
        Zhedong~Zheng,~\IEEEmembership{Senior Member,~IEEE},
        and~Tao~Li%
\thanks{Zhenning Shi and Chen Xu contributed equally. Tao Li and Zhedong Zheng are the corresponding authors.}%
\thanks{Zhenning Shi, Kefei Zhang, and Tao Li are with College of Computer Science, Nankai University, Tianjin, China (e-mail: litao@nankai.edu.cn).}%
\thanks{Chen Xu and Zhedong Zheng are with Faculty of Information Science and Computing, University of Macau, Macau, China (e-mail: zhedongzheng@um.edu.mo).}%
\thanks{Junhao Zhang is with CSIRO Data61, Australia.}%
\thanks{Linjie Liu is with Beihang University, Beijing, China.}}

\hypersetup{%
  pdftitle={ScaleResfusion: Residual Rectified Flow based on Residual Vector Field},
  pdfauthor={Zhenning Shi, Chen Xu, Junhao Zhang, Kefei Zhang, Linjie Liu, Zhedong Zheng, Tao Li},
  pdfsubject={Manuscript prepared for IEEE Transactions on Image Processing},
  pdfkeywords={diffusion models, image restoration, image super-resolution, rectified flow}%
}

\markboth{IEEE Transactions on Image Processing}%
{Shi \MakeLowercase{\textit{et al.}}: ScaleResfusion}

\maketitle

\begin{abstract}
Real-world Image Restoration (Real-IR) aims to recover high-quality (HQ) images from complex and unknown degradations. Recent diffusion-based methods have substantially improved perceptual quality, yet two obstacles remain: methods that sample from Gaussian noise require many steps and are often less faithful to the degraded input, whereas residual-based methods that start from the low-quality (LQ) image typically train task-specific models from scratch, with optimization objectives coupled to a particular noise scheduler, and therefore cannot reuse modern pre-trained generative priors. We present \textbf{ScaleResfusion}, which rewrites residual restoration as a scheduler-independent adaptation interface for pre-trained text-to-image rectified-flow models. Its core, \textbf{Residual Rectified Flow} (RRF), inserts the residual term $R$ into the linear transport path of Rectified Flow, so that sampling starts from noisy LQ at an exact acceleration point, where the signal-to-noise ratio of the starting state is continuously controlled by the residual ratio $\gamma$. The resulting optimization target, the \textbf{residual vector field}, contains no scheduler-specific coefficients and differs from the pre-trained rectified-flow target only by the residual offset $\gamma R$; adapting a frozen billion-scale backbone therefore reduces to fitting this compact residual correction with LoRA-only training. A knowledge-distillation pipeline built around RRF further reduces sampling to as few as 4 steps. Experiments on real-world super-resolution across multiple benchmarks show that ScaleResfusion achieves state-of-the-art restoration quality and transfers consistently across pre-trained rectified-flow backbones from 2B to 9B parameters.
\end{abstract}


\section{Introduction}
\label{sec:intro}
\IEEEPARstart{I}{mage} restoration (IR) aims to recover a high-quality (HQ) image from low-quality (LQ) observations ~\cite{dong2014learning, zhang2018image, zhang2017learning, chen2021pre, liang2021swinir, zhang2022efficient, chen2023activating, ledig2017photo, wang2018esrgan, wang2021real}.
Traditional image restoration methods~\cite{dong2014learning, zhang2018image, liang2021swinir, zhang2022efficient, chen2023activating} typically simplify the degradation process from HQ to LQ images as bicubic downsampling or Gaussian blur.
However, the degradation process in real-world image restoration (Real-IR) is typically complex, unknown, and often compound, and remains a highly ill-posed problem in practical scenarios~\cite{zhang2021designing, wang2021real, fan2024toward, 10616018}.

To address this problem, researchers have explored complex synthetic degradation pipelines~\cite{wang2018esrgan, wang2021real}, perceptual losses~\cite{johnson2016perceptual}, and end-to-end restoration models trained on paired LQ-HQ data~\cite{dong2014learning, zhang2017learning, lim2017enhanced, zhang2018image, chen2021pre, liang2021swinir, zamir2022restormer, zhang2022efficient, chen2023activating}. However, directly learning an LQ-to-HQ mapping with pixel-wise supervision often produces over-smoothed results or texture hacking~\cite{ledig2017photo, wang2018esrgan}. Generative adversarial network (GAN) methods~\cite{goodfellow2014generative, ledig2017photo, wang2018esrgan, wang2021real, zhang2021designing, liang2022details} improve perceptual realism~\cite{wang2021towards}, but they tend to generate uncontrollable artifacts and suffer from unstable adversarial training.

Recently, diffusion-based models~\cite{ho2020denoising, song2020denoising, song2019generative, song2020score, lipman2022flow, liu2022flow} have shown remarkable ability in synthesizing high-fidelity restored images~\cite{saharia2022image, ozdenizci2023restoring, whang2022deblurring, he2025diffusion}. 
To leverage the powerful prior knowledge of pre-trained text-to-image (T2I) diffusion models~\cite{rombach2022high, esser2024scaling}, recent works~\cite{lin2024diffbir, wang2024exploiting, yang2024pixel, wu2024seesr, yu2024scaling} introduce the LQ image as a conditional input~\cite{zhang2023adding}, producing more realistic images than GAN-based methods with more stable training. However, these methods usually consider only LQ images as the condition and start diffusion from Gaussian noise, resulting in lengthy inference and weak consistency between the generated output and the LQ input. Although some works~\cite{wu2024one, dong2025tsd, li2025one} reduce diffusion steps through knowledge distillation to achieve single-step inference, they tend to learn a LQ-HQ mapping while abandoning the multi-step sampling characteristic of diffusion models, leading to reduced diversity and fidelity.

Another line of diffusion-based restoration methods~\cite{delbracio2023inversion, luo2023image, luo2023refusion, liu2023i2sb, yue2023resshift, shi2024resfusion, wang2025residual} starts the diffusion process from noisy LQ images rather than Gaussian noise, shortening the sampling trajectory while preserving the stochastic sampling capability of diffusion models. However, these methods typically train task-specific diffusion models from scratch, with training objectives and customized noise schedules that deviate from those of pre-trained models, making direct reuse difficult. Resfusion~\cite{shi2024resfusion} partially mitigates this within Denoising Diffusion Probabilistic Models (DDPM)~\cite{ho2020denoising} by learning the residual noise (resnoise) and unifying training and inference through a smooth equivalence transformation. Nevertheless, its resnoise target is tied to the DDPM marginal coefficients ($\alpha_t$, $\bar{\alpha}_t$, $\beta_t$), so changing the noise scheduler or the sampling interval requires retraining; it's model is likewise trained from scratch rather than adapted from a pre-trained T2I model; and its acceleration point is a discrete step hard-coded by the schedule, without explicit control over the signal-to-noise ratio (SNR) of the starting state. Since modern large-scale pre-trained models~\cite{esser2024scaling, cai2025z, flux-2-2025} are built on Rectified Flow~\cite{liu2022flow}, these couplings leave a concrete open problem: expressing residual transport as a scheduler-independent offset on the pre-trained rectified-flow objective, so that a frozen billion-scale backbone can be adapted to restoration by fitting only a compact residual correction.

To address this problem, we propose \textbf{ScaleResfusion}, which rewrites residual restoration as a residual-vector-field adaptation interface for pre-trained T2I Rectified Flow models. Its core is \textbf{Residual Rectified Flow} (RRF), which inserts the weighted residual $\gamma R$ into the Rectified Flow trajectory while preserving its linear transport structure, supported by a measure-transport analysis showing that, under standard stability assumptions, the terminal discrepancy of the reverse flow is controlled by that of its initialization, so an observation-anchored start closer to the task-aware initial measure admits a tighter terminal-discrepancy bound than a task-agnostic Gaussian one (Sec.~\ref{sec: residual modeling}). Sampling starts from a noisy LQ state at the exact acceleration point $t^\star = 1/(1+\gamma)$, where the residual ratio $\gamma$ continuously controls the SNR of the initialization---replacing Resfusion's hard-coded discrete approximation with explicit, continuous control. Crucially, the optimization target, the \textbf{residual vector field} $resv=v^{RF}+\gamma R$, contains no scheduler-specific marginal coefficients and differs from the pre-trained Rectified Flow target only by the residual offset $\gamma R$; adapting a frozen backbone therefore reduces to fitting a compact residual correction, which we realize with \textbf{Low-Rank Adaptation (LoRA)}~\cite{hu2022lora} alone. Around this core, we build a knowledge-distillation based parameter-efficient fine-tuning (PEFT) pipeline that reduces sampling to as few as 4 steps. Experiments on real-world super-resolution across multiple benchmarks show that ScaleResfusion achieves state-of-the-art performance and transfers consistently across pre-trained rectified-flow backbones from 2B to 9B parameters. Moreover, when the LQ and HQ images share a representation in which the residual $R$ is well-defined and the two endpoints remain close, RRF is in principle applicable to other restoration settings.

Our contributions can be summarized as follows:

(1) We formulate \textbf{Residual Rectified Flow} for Real-IR by incorporating image residuals into the Rectified Flow trajectory, enabling the diffusion process to start directly from noisy LQ images within the Rectified Flow framework. The formulation is supported by a measure-transport analysis and admits an exact acceleration point $t^\star = 1/(1+\gamma)$, whose starting-state SNR is continuously controlled by the residual ratio $\gamma$ rather than hard-coded by a discrete noise schedule.

(2) We derive the \textbf{residual vector field} as the optimization target and show that it serves as a scheduler-independent adaptation interface: it contains no scheduler-specific marginal coefficients and differs from the pre-trained Rectified Flow target only by the residual offset $\gamma R$, making LoRA-only PEFT of frozen pre-trained rectified-flow backbones principled and interpretable.

(3) Around the RRF core, we construct an efficient training pipeline with knowledge distillation, and demonstrate on real-world super-resolution across multiple benchmarks that residual-based diffusion modeling extends to billion-scale pre-trained models (2B--9B) \textbf{with only 4 sampling steps} and $\sim$60M trainable parameters, transferring consistently across rectified-flow backbones.

\section{Related Work}
Traditional Real-IR methods typically learn a direct LQ-to-HQ mapping under simplified or synthetic degradations~\cite{dong2014learning, zhang2017learning, zhang2018image, chen2021pre, liang2021swinir, zhang2022efficient, chen2023activating, wang2021real}, while GAN-based methods improve perceptual realism by matching the natural image distribution~\cite{goodfellow2014generative, ledig2017photo, wang2018esrgan, zhang2021designing, liang2022details}. However, these methods often suffer from over-smoothed details, unstable adversarial training, or uncontrollable artifacts. These limitations have shifted recent Real-IR research toward diffusion-based restoration, which offers stronger generative priors and a more stable training paradigm. In this context, we summarize current diffusion-based methods from two perspectives: diffusion models initialized from Gaussian noise and those initialized from noisy LQ images. 

\textbf{Diffusion models initialized from Gaussian noise.}
Recent Real-IR methods often leverage pre-trained T2I diffusion priors by injecting the LQ image as a condition while keeping the original Gaussian noise initialization.
StableSR~\cite{wang2024exploiting} balances fidelity and perceptual quality by fine-tuning a time-aware encoder and using controllable feature wrapping.
DiffBIR~\cite{lin2024diffbir} restores the LR images with a reconstruction network and then uses Stable Diffusion~\cite{rombach2022high} to synthesize realistic details.
SeeSR~\cite{wu2024seesr} extracts semantic features from the input image to better activate the generative prior of Stable Diffusion.
PASD~\cite{yang2024pixel} combines pixel-aware cross-attention with degradation-robust guidance to preserve local structure during generation.
SUPIR~\cite{yu2024scaling} further improves generative fidelity through negative-quality prompts, restoration-guided sampling, and larger-scale pre-training.
Overall, despite promising perceptual quality, these methods still sample from Gaussian white noise with the LQ image acting only as a condition, which entails 20--50 inference steps and weakens fidelity to the degraded input. This motivates the noisy-LQ initialization of our Residual Rectified Flow, which anchors the starting state to the degraded input itself.

\begin{figure*}
  \includegraphics[width=\textwidth]{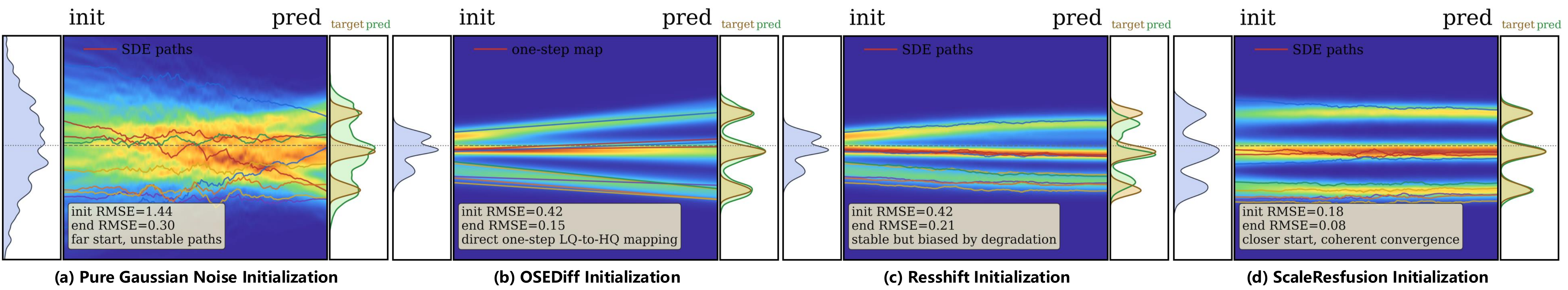}
  \caption{
  Initialization comparison from a transport perspective. 
  (a) Gaussian-noise initialization starts far from the restoration manifold and induces unstable paths. 
  (b) One-step restoration starts from the LQ observation but collapses the generative restoration process into a deterministic LQ-to-HQ mapping, losing the sampling capability of diffusion models. 
  (c) Noisy-LQ-initialized DDPM sampling injects stochasticity, yet it biases the reverse path toward the input degradation without an explicit residual correction direction. 
  (d) ScaleResfusion starts from a noisy LQ state and follows a residual-oriented transport, producing a closer initialization and more coherent convergence toward the target distribution.
  }
  \label{fig:flow}
\end{figure*}

\textbf{Diffusion models initialized from noisy LQ images.}
To reduce inference cost, OSEDiff~\cite{wu2024one}, TSD-SR~\cite{dong2025tsd}, and FluxSR~\cite{li2025one} distill pre-trained diffusion priors into a direct LQ-to-HQ mapping for one-step restoration. While efficient, they largely replace iterative sampling with a deterministic mapping, which may reduce diversity and fidelity.
Other works preserve iterative sampling by initializing diffusion from noisy LQ images.
IR-SDE~\cite{luo2023image} models degradation and restoration with a mean-reverting stochastic differential equation.
Refusion~\cite{luo2023refusion} performs realistic large-size restoration through latent-space diffusion.
DriftRec~\cite{welker2024driftrec} adapts the forward stochastic differential equation to blind JPEG restoration by introducing a drift toward the degraded observation.
ResShift~\cite{yue2023resshift} accelerates super-resolution by shifting the residual between LQ and HQ images.
Residual Denoising Diffusion Models (RDDM)~\cite{liu2024residual} jointly model the residual and noise terms.
Resfusion~\cite{shi2024resfusion} learns resnoise to unify training and inference through a smooth equivalence transformation, preserving consistency with the original diffusion process. However, its resnoise target is coupled to the DDPM~\cite{ho2020denoising} marginal coefficients ($\alpha_t$, $\bar{\alpha}_t$, $\beta_t$), so changing the noise scheduler or the sampling interval requires retraining, and its acceleration point is a discrete step hard-coded by the schedule without explicit SNR control.
Overall, noisy-LQ initialization shortens the sampling trajectory while preserving the diffusion sampling nature, but most existing methods train task-specific diffusion models from scratch with customized objectives or scheduler-coupled targets, leaving no direct route to reusing pre-trained T2I priors. This motivates our ScaleResfusion, whose Residual Rectified Flow preserves the linear transport structure of Rectified Flow and whose training target differs from the pre-trained rectified-flow objective only by the residual offset $\gamma R$, without any scheduler-specific coefficients, so that a frozen pre-trained backbone can be adapted by fitting a compact residual correction.

\section{Method}
\label{sec: method}
\subsection{Preliminaries}
\label{sec: preliminaries}
Denoising Diffusion Probabilistic Models (DDPM) \cite{ho2020denoising} aim
to approximate the real data distribution $p(x_0)$ with the learned model distribution $p_\theta (x_0)$ through a discrete Markov chain. Flow-based generative models \cite{lipman2022flow, liu2022flow} improve sampling efficiency by replacing the discrete Markovian process with a continuous transport path between a data sample $x_0 \sim p(x_0)$ and Gaussian noise $\epsilon \sim \mathcal{N}(0,I)$, formulated as
\begin{equation}
\label{eq:flow_path}
x_t=a_t x_0+b_t\epsilon, t\in[0,1].
\end{equation}
The goal is to learn a vector field $v_\theta(x_t,t)$ that approximates the target vector field along this path. Rectified Flow (RF)~\cite{liu2022flow} adopts the linear path $x_t=(1-t)x_0+t\epsilon$ and optimizes
\begin{equation}
\label{eq:rf_flow_loss}
\mathcal{L}_{RF}
=\int_{0}^{1}\mathbb{E}_{x_0,\epsilon}
\left[
\left\|v_\theta(x_t,t)-(\epsilon-x_0)\right\|^2
\right]dt.
\end{equation}
Sampling is performed by solving the reverse ODE
\begin{equation}
\label{eq:rf_reverse_ode}
x_0=x_1+\int_{1}^{0}v_\theta(x_\tau,\tau)d\tau,
\end{equation}
This Gaussian-to-image transport is effective for generic generation, but it is redundant for image restoration since the LQ image itself already carries effective information.
\subsection{Residual Modeling for Image Restoration}
\label{sec: residual modeling}
Given an LQ--HQ image pair, we denote the LQ observation and the HQ target as $\hat{x}_0$ and $x_0$ respectively, and define the residual term as
\begin{equation}
\label{eq:residual_definition}
R=\hat{x}_0-x_0.
\end{equation}
Conditioned on the observed LQ image $\hat{x}_0$, recovering $x_0$ is therefore equivalent to estimating the correction $R$, since $x_0=\hat{x}_0-R$. This residual formulation is particularly suitable for restoration: the LQ image already preserves most semantic layout and coarse structures, while the residual mainly captures local degradation artifacts, high-frequency details, and texture corrections. The formulation itself is not specific to super-resolution: in principle, it applies whenever the LQ observation and the HQ target share a representation in which $R$ is well-defined and the two endpoints remain close. In this paper, we instantiate and validate it on real-world super-resolution.

\begin{figure}[ht]
    \centering
    \includegraphics[width=0.45\textwidth]{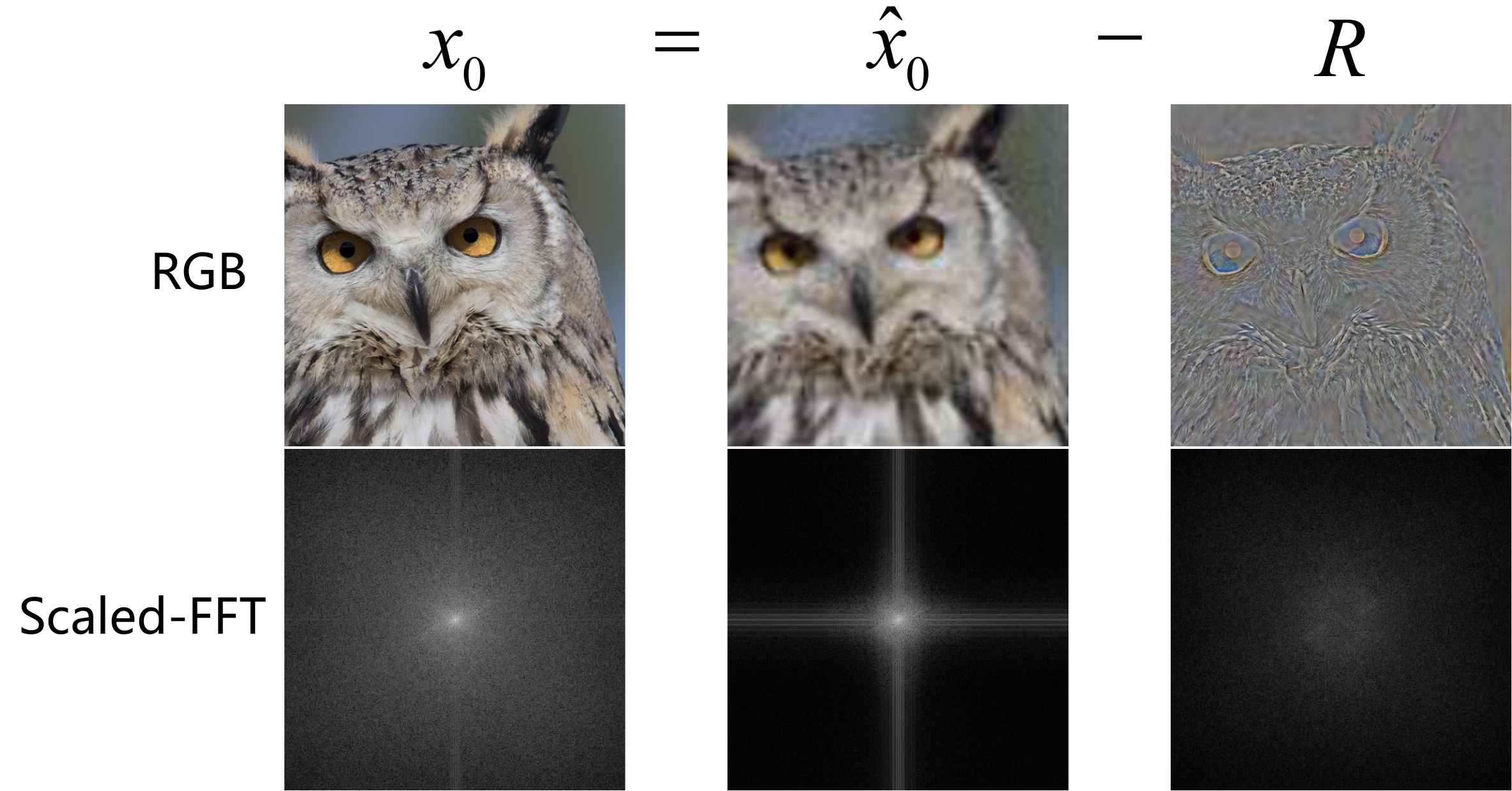}
    \caption{The residual term $R$ exhibits consistent low-energy, edged patterns in both the spatial and frequency domains, suggesting that RRF only needs to learn a compact residual update around the pre-trained transport.}
    \label{fig:residual}
\end{figure}

As shown in Fig.~\ref{fig:residual}, the residual is much more compact than either the LQ or HQ image. In the spatial domain, $R$ appears as a low-amplitude and localized correction around edges, textures, and degradation artifacts. After the \textbf{Scaled Fourier Transform}, it also exhibits substantially lower spectral energy than the image signals. This observation suggests that the task-specific transformation for restoration is concentrated in a compact residual subspace, rather than being distributed across the full image-generation process. Such a localized and energy-limited update is naturally compatible with \textbf{LoRA}~\cite{hu2022lora}, which is designed to capture low-rank adaptations around a pre-trained model.

\begin{figure*}[ht]
    \centering
    \includegraphics[width=0.85\textwidth]{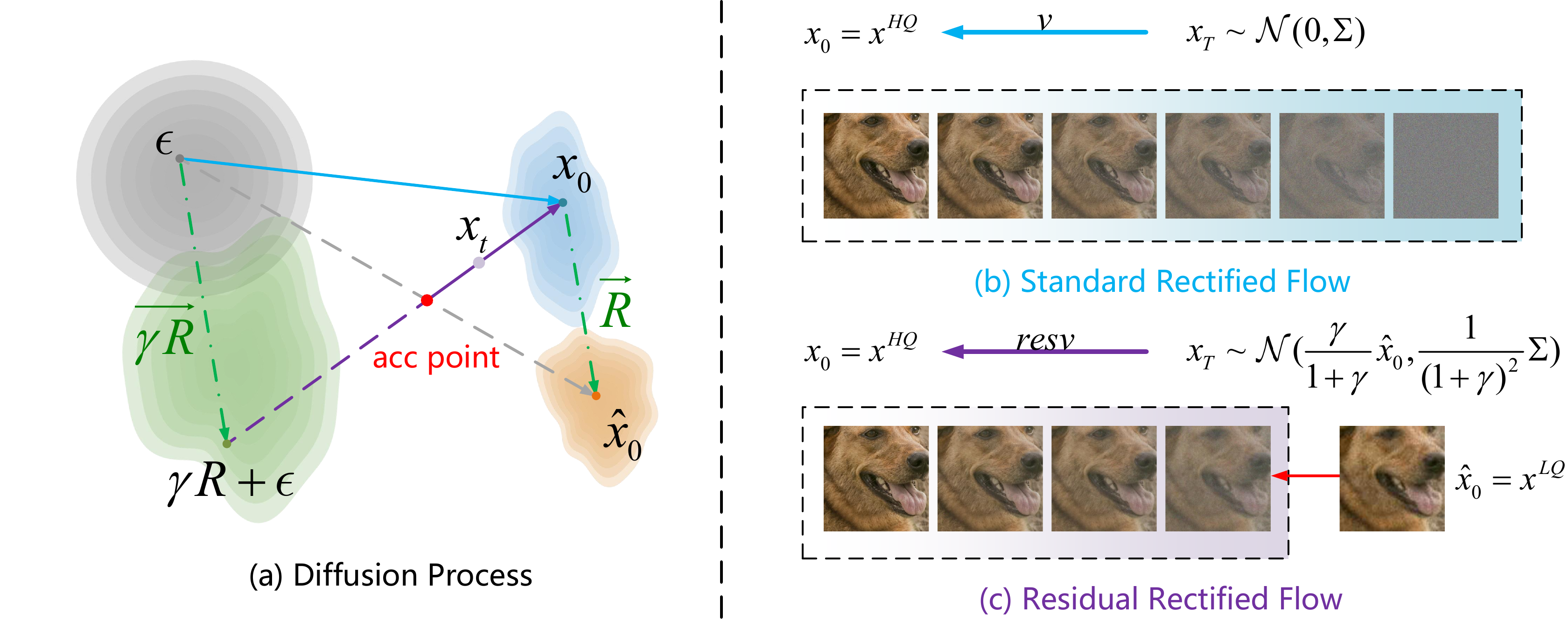}
    \caption{Geometric Interpretation of Residual Rectified Flow (RRF). (a) \textcolor{red!50!blue}{RRF} introduces the weighted residual term $\gamma R$ into the diffusion path, yielding a straight trajectory that intersects the implicit noise-to-LQ path at the acceleration point. (b) \textcolor{cyan!50!blue}{Standard Rectified Flow} transports Gaussian noise to the HQ image distribution through a linear ODE. (c) Starting from the noisy-LQ state $x_{t^\star}\sim\mathcal{N}\left(\frac{\gamma}{1+\gamma}\hat{x}_0,\frac{1}{(1+\gamma)^2}\Sigma\right)$, \textcolor{red!50!blue}{RRF} learns the residual vector field $resv$ to recover the HQ image $x_0$ over a shortened sampling interval.}
    \label{fig:theory}
\end{figure*}

The residual formulation is also preferable from a transport perspective. The reverse ODE is initialization-sensitive: under standard stability assumptions, an initial state closer to the task-relevant restoration manifold is more likely to be transported to a terminal state close to the desired HQ distribution. Pure Gaussian initialization is task-agnostic and far from the observed content, whereas residual modeling preserves the structural prior in $\hat{x}_0$ and makes the transport correction-oriented through $R$. Fig.~\ref{fig:flow} compares different strategies: Gaussian initialization suffers from large initial discrepancy; one-step restoration (OSEDiff) loses the sampling capability of diffusion models by using a deterministic LQ-to-HQ mapping; and Noisy-LQ-initialized DDPM sampling (ResShift) remains degradation-biased without an explicit residual direction. In contrast, ScaleResfusion starts from a noisy LQ state and follows a residual-oriented path, yielding more coherent trajectories and lower endpoint error. A formal measure-transport analysis is provided in Appendix~\ref{sec:residual_transport_theoretical_justification}.
Overall, the above analysis shows that residual modeling provides both a compact learning target and a better-aligned transport perspective. This motivates a residual-oriented flow that starts from a noisy LQ state and transports along an explicit correction direction. The next subsection formalizes this idea as a continuous Residual Rectified Flow trajectory.
\subsection{Residual Rectified Flow}
\label{sec: residual rectified flow}
\begin{figure*}[ht]
    \centering
    \includegraphics[width=0.75\textwidth]{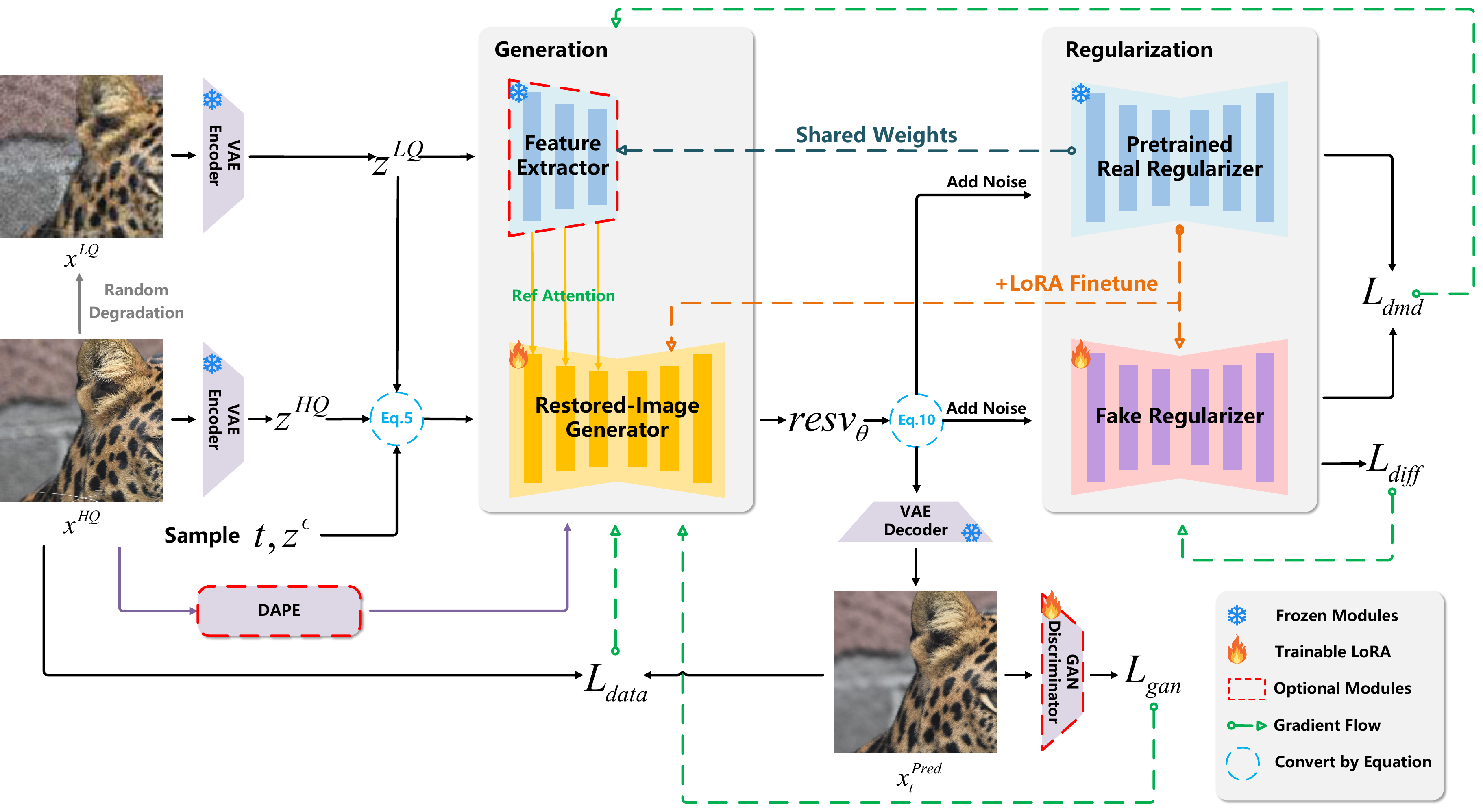}
    \caption{Overview of our Training Pipeline, which consists of a generation module and a regularization module. In the generation module, the restored-image generator predicts the residual vector field under Residual Rectified Flow, with LQ features extracted by the Feature Extractor and injected through Ref Attention, and DAPE providing textual constraints. The generation and regularization modules share the same pre-trained frozen backbone weights, while LoRA adapters enable efficient adaptation. In the regularization module, DMD and optional GAN losses regularize the restored distribution toward natural HQ images.}
  \label{fig:ScaleResfusion}
\end{figure*}

Following the residual definition in Eq.~\eqref{eq:residual_definition}, the proposed \textbf{Residual Rectified Flow} (RRF) introduces the weighted residual term into the RF trajectory in Eq.~\eqref{eq:flow_path}. Given a residual ratio $\gamma > 0$, $t\in[0,1]$, and $\epsilon \sim \mathcal{N}(0,I)$, RRF constructs
\begin{equation}
\label{eq:rrf_forward_path}
x_t = (1-t)x_0 + t\gamma R + t\epsilon.
\end{equation}
The trajectory is a straight line connecting the HQ image $x_0$ and the endpoint $\gamma R+\epsilon$, whose target velocity is constant with respect to
$t$:
\begin{equation}
\label{eq:rrf_velocity}
u_t(x_t \mid x_0,\epsilon,R) = \frac{d x_t}{dt} = resv
= \gamma R + \epsilon - x_0.
\end{equation}

As shown in Fig.~\ref{fig:theory}, the \textcolor{cyan!50!blue}{blue} arrow represents the linear diffusion process of Standard Rectified Flow (RF), which is an ODE from the Gaussian distribution to the HQ image distribution.
While the \textcolor{red!50!blue}{purple} arrow represents the proposed Residual Rectified Flow (RRF), which constructs a continuous path between $x_0$ and a noisy weighted residual term $\gamma R+\epsilon$. In the diffusion process of RRF, it implicitly reduces the residual term $R$ between $x_0$ and $\hat{x_0}$ with the residual vector field. By the definition of similar triangles, we can obtain the expression for $x_t$ as in Eq.~\eqref{eq:rrf_forward_path}, where the coefficient of $\gamma R$ is $t$.

When we define $\hat{x_t}=(1-t)\hat{x_0}+t\epsilon$ with the same Gaussian noise $\epsilon$ as in $x_t$, the \textcolor{gray}{gray} arrow represents an implicit Rectified Flow diffusion process from the Gaussian distribution to the LQ image $\hat{x_0}$ distribution. Due to the residual term R, the trajectories of \textcolor{red!50!blue}{RRF} and the \textcolor{gray}{implicit Rectified Flow} are constrained on the same hyperplane, and their \textbf{intersection} is the acceleration point~\cite{shi2024resfusion}. 
The proposed Residual Rectified Flow admits an exact acceleration point due to its continuous ODE trajectory. Substituting $R=\hat{x}_0-x_0$ into Eq.~\eqref{eq:rrf_forward_path} gives:
\begin{equation}
\label{eq:rrf_expand}
x_t = (1-t)x_0 + t\gamma(\hat{x}_0-x_0) + t\epsilon
= (1-(1+\gamma)t)x_0 + t\gamma \hat{x}_0 + t\epsilon.
\end{equation}
The coefficient of $x_0$ vanishes when $1-(1+\gamma)t^\star = 0$, which leads to the exact acceleration point $t^\star = \frac{1}{1+\gamma}$. At this point, the state of RRF can be completely represented by the LQ image $\hat{x_0}$ and noise $\epsilon$, no longer depending on the unknown HQ image $x_0$, thereby enabling accelerated sampling. Accordingly, RRF trains a neural \textbf{residual velocity field} $resv_\theta(x_t,t)$ only over the effective interval $[0,t^\star]$:
\begin{equation}
\label{eq:rrf_flow_loss}
\mathcal{L}_{RRF}
= \int_{0}^{t^\star}
\mathbb{E}\left[
\left\|resv_\theta(x_t,t)-(\gamma R + \epsilon - x_0)\right\|^{2}
\right]dt.
\end{equation}

\begin{figure}[t]
\center
\begin{minipage}{\columnwidth}
\begin{algorithm}[H]
	\footnotesize
	\caption{Residual Rectified Flow: Main Algorithm. Modifications relative to Standard Rectified Flow are highlighted in \textcolor{red}{red}.}
	\label{alg: main algorithm}
	\begin{algorithmic}
		\State \textbf{Training:} 		
		\State \textbf{Require:} Paired LQ-HQ samples $(\hat{x}_0, x_0)\sim D$, residual ratio $\gamma$.
		\State $\textcolor{red}{t^\star = \frac{1}{1+\gamma}}$
		\Repeat
    	\State Sample $(\hat{x}_{0}^i$, $x_{0}^i)\sim D,\epsilon\sim N(0,I)$, $t \sim \mathcal{U}(0, t^\star)$
            \State $\textcolor{red}{R = \hat{x}_{0} - x_{0}}$ 
            \State ${x}_{t}= (1-t) x_{0} + 
            \textcolor{red}{t \gamma R} 
            + t \epsilon$
            \State $\textcolor{red}{resv} = \textcolor{red}{\gamma R} + \epsilon - x_{0}$
            \State take gradient step on $\nabla_{\theta} ||\textcolor{red}{resv}-\textcolor{red}{resv_{\theta}}({x}_{t},t)||^2$
    	\Until{convergence} 

		\Statex \hrulefill

		\State \textbf{Inference:} 
		\State \textbf{Require:} LQ image $\hat{x}_{0}$, residual velocity field $resv_{\theta}$, residual ratio $\gamma$.
		\State $\textcolor{red}{t^\star = \frac{1}{1+\gamma}}$
		\State Sample $\epsilon\sim N(0,I)$
		\State $\textcolor{red}{x_{t^\star} = t^\star \gamma \hat{x_0} + t^\star \epsilon= \frac{\gamma}{1+\gamma} \hat{x}_{0} + \frac{1}{1+\gamma} \epsilon}$
		\State Solve the ODE $\frac{d x_t}{dt} = resv_{\theta}(x_t, t)$ from $t=t^\star$ to $0$
		\State
		\Return $x_0$
	\end{algorithmic}
\end{algorithm}
\end{minipage}
\end{figure}

For a given LQ observation $\hat{x}_0$, RRF is initialized at $t^\star$ from $\mathcal{N}\left(\frac{\gamma}{1+\gamma}\hat{x}_0,\frac{1}{(1+\gamma)^2}I\right)$. We can obtain its signal-to-noise ratio (SNR)~\cite{kingma2021variational} as:
\begin{equation}
    \operatorname{SNR}(x_{t^\star})\propto
    \frac{\frac{\gamma}{1+\gamma}}{\frac{1}{1+\gamma}}=\gamma.
\end{equation}
Thus, unlike the discrete acceleration point of Resfusion~\cite{shi2024resfusion}, which is hard-coded by the noise schedule, the residual ratio $\gamma$ continuously and explicitly controls the SNR level of the starting point: larger $\gamma$ yields an initialization closer to the LQ image, while smaller $\gamma$ injects stronger Gaussian randomness. During sampling, RRF starts from $x_{t^\star}$ and integrates the learned ODE $x_0 = x_{t^\star} + \int_{t^\star}^{0}resv_\theta(x_\tau,\tau)d\tau$ to the data endpoint. 
Compared with Resfusion~\cite{shi2024resfusion}, whose residual-noise parameterization is tied to discrete DDPM transition coefficients, we formulate residual adaptation within continuous-time Rectified Flow. As formalized in Eq.~\eqref{eq:rrf_forward_path} and Eq.~\eqref{eq:rrf_velocity}, and detailed in Algorithm~\ref{alg: main algorithm}, RRF preserves the Standard Rectified Flow structure and only introduces residual offsets: $x_t=x_t^{RF}+t\gamma R$ and $resv=v^{RF}+\gamma R$, neither of which involves scheduler-specific marginal coefficients. Therefore, the residual vector field acts as a scheduler-independent adaptation interface: adapting a pre-trained rectified-flow model to RRF does not require relearning the image-noise transport, but only fitting the compact residual correction $\gamma R$. Given a sufficiently expressive pre-trained RF backbone, we hypothesize that low-rank parameter updates can capture both the restoration-specific information carried by the residual term and the corresponding adjustment of the velocity predictor. This motivates using \textbf{LoRA}~\cite{hu2022lora} to adapt the backbone jointly to the residual-augmented velocity target and the RRF state distribution.
\begin{table*}[!t]\tiny
\setlength{\abovecaptionskip}{0.10cm}
\caption{Quantitative comparison on real-world benchmarks with our default FLUX2-Klein (4B) variants. Higher is better for PSNR, SSIM, MUSIQ, and MANIQA, while lower is better for LPIPS, DISTS, FID, and NIQE. Within each benchmark, the first group lists GAN-based methods and the second group diffusion-based methods. The best and second-best results of each metric are highlighted in \textcolor{red}{red} and \textcolor{blue}{blue}, respectively. Results for the other ScaleResfusion backbones are consolidated in Table~\ref{tab:ablation_backbone}.}
\label{tab:multi-step-all-real}
\centering
\renewcommand{\arraystretch}{1.08}
\setlength{\tabcolsep}{3.8pt}
\resizebox{\textwidth}{!}{%
\begin{tabular}{@{}c|c|cccccccc@{}}
\toprule
\textbf{Datasets} & \textbf{Method} & \textbf{PSNR $\uparrow$} & \textbf{SSIM $\uparrow$} & \textbf{LPIPS $\downarrow$} & \textbf{DISTS $\downarrow$} & \textbf{FID $\downarrow$} & \textbf{NIQE $\downarrow$} & \textbf{MUSIQ $\uparrow$} & \textbf{MANIQA $\uparrow$} \\ \midrule
 & BSRGAN & {\color[HTML]{4E83FD} \textbf{28.70}} & 0.80 & 0.29 & {\color[HTML]{4E83FD} \textbf{0.21}} & 155.61 & 6.54 & 57.15 & 0.48 \\
 & Real-ESRGAN & 28.61 & {\color[HTML]{4E83FD} \textbf{0.81}} & {\color[HTML]{4E83FD} \textbf{0.28}} & {\color[HTML]{4E83FD} \textbf{0.21}} & 147.66 & 6.70 & 54.27 & 0.49 \\
 & LDL & 28.20 & {\color[HTML]{4E83FD} \textbf{0.81}} & {\color[HTML]{4E83FD} \textbf{0.28}} & {\color[HTML]{4E83FD} \textbf{0.21}} & 155.51 & 7.14 & 53.94 & 0.49 \\
 & FeMaSR & 26.87 & 0.76 & 0.32 & 0.22 & 157.72 & {\color[HTML]{D83931} \textbf{5.91}} & 53.70 & 0.44 \\ \cline{2-10}
 & StableSR & 28.04 & 0.75 & 0.33 & 0.23 & 144.15 & 6.60 & 58.53 & 0.56 \\
 & SUPIR & 25.09 & 0.65 & 0.42 & 0.28 & 169.48 & 7.39 & 58.79 & 0.55 \\
 & TSD-SR & 27.77 & 0.76 & 0.30 & {\color[HTML]{4E83FD} \textbf{0.21}} & 134.98 & {\color[HTML]{D83931} \textbf{5.91}} & {\color[HTML]{D83931} \textbf{66.62}} & 0.59 \\
 & AddSR & 26.68 & 0.74 & 0.37 & 0.26 & 164.82 & 7.80 & 65.36 & 0.60 \\
 & CCSR & 28.24 & 0.78 & 0.32 & 0.23 & 157.30 & 6.81 & 66.28 & 0.61 \\
 & DiffBIR & 25.90 & 0.62 & 0.47 & 0.29 & 180.33 & 6.33 & 66.13 & {\color[HTML]{4E83FD} \textbf{0.62}} \\
 & OSEDiff & 27.92 & 0.78 & 0.30 & 0.22 & 135.41 & 6.46 & 64.69 & 0.59 \\
 & PASD & 28.02 & 0.78 & 0.32 & 0.23 & 174.76 & 6.72 & 57.23 & 0.51 \\
 & ResShift & 27.05 & 0.74 & 0.39 & 0.26 & 159.90 & 8.65 & 51.24 & 0.47 \\
 & SeeSR & 28.07 & 0.77 & 0.32 & 0.23 & 147.37 & 6.41 & 65.09 & 0.61 \\
 & FluxSR & 20.88 & 0.61 & 0.39 & 0.25 & 144.66 & 7.11 & {\color[HTML]{4E83FD} \textbf{66.38}} & 0.57 \\ \cline{2-10}
\rowcolor[gray]{0.95}
 & \textbf{Ours (w/o GAN)} & {\color[HTML]{D83931} \textbf{29.77}} & {\color[HTML]{D83931} \textbf{0.82}} & {\color[HTML]{D83931} \textbf{0.25}} & {\color[HTML]{D83931} \textbf{0.20}} & {\color[HTML]{D83931} \textbf{118.18}} & 6.95 & 62.42 & 0.61 \\
\rowcolor[gray]{0.95}
\multirow{-17}{*}{\textbf{DRealSR}} & \textbf{Ours (w/ GAN)} & 28.26 & 0.78 & 0.29 & {\color[HTML]{4E83FD} \textbf{0.21}} & {\color[HTML]{4E83FD} \textbf{124.03}} & {\color[HTML]{4E83FD} \textbf{6.21}} & 65.16 & {\color[HTML]{D83931} \textbf{0.64}} \\ \midrule
 & BSRGAN & 26.38 & {\color[HTML]{4E83FD} \textbf{0.77}} & 0.27 & {\color[HTML]{4E83FD} \textbf{0.21}} & 141.24 & 5.64 & 63.28 & 0.54 \\
 & Real-ESRGAN & {\color[HTML]{4E83FD} \textbf{26.65}} & 0.76 & 0.27 & {\color[HTML]{4E83FD} \textbf{0.21}} & 136.29 & 5.85 & 60.45 & 0.55 \\
 & LDL & 25.28 & 0.76 & 0.28 & {\color[HTML]{4E83FD} \textbf{0.21}} & 142.74 & 5.99 & 60.92 & 0.55 \\
 & FeMaSR & 25.06 & 0.74 & 0.29 & 0.23 & 141.01 & 5.77 & 59.05 & 0.49 \\ \cline{2-10}
 & StableSR & 24.62 & 0.70 & 0.31 & 0.22 & 128.54 & 5.78 & 65.48 & 0.62 \\
 & SUPIR & 23.65 & 0.66 & 0.35 & 0.25 & 130.38 & 6.11 & 62.09 & 0.58 \\
 & TSD-SR & 24.81 & 0.72 & 0.27 & {\color[HTML]{4E83FD} \textbf{0.21}} & 114.45 & {\color[HTML]{D83931} \textbf{5.13}} & {\color[HTML]{4E83FD} \textbf{71.19}} & 0.63 \\
 & AddSR & 22.65 & 0.65 & 0.38 & 0.27 & 154.18 & 6.62 & {\color[HTML]{D83931} \textbf{71.41}} & {\color[HTML]{4E83FD} \textbf{0.67}} \\
 & CCSR & 25.92 & 0.75 & 0.28 & {\color[HTML]{4E83FD} \textbf{0.21}} & 122.84 & 5.73 & 69.18 & 0.64 \\
 & DiffBIR & 24.83 & 0.65 & 0.36 & 0.24 & 130.75 & 5.84 & 69.28 & 0.65 \\
 & OSEDiff & 25.15 & 0.73 & 0.29 & {\color[HTML]{4E83FD} \textbf{0.21}} & 123.53 & 5.65 & 69.08 & 0.63 \\
 & PASD & 26.04 & 0.74 & 0.28 & {\color[HTML]{4E83FD} \textbf{0.21}} & 135.48 & 5.71 & 60.03 & 0.56 \\
 & ResShift & 25.66 & 0.74 & 0.33 & 0.25 & 128.03 & 8.07 & 56.89 & 0.51 \\
 & SeeSR & 25.15 & 0.72 & 0.30 & 0.22 & 125.30 & 5.40 & 69.81 & 0.65 \\
 & FluxSR & 23.83 & 0.69 & 0.32 & 0.23 & 120.63 & 6.57 & 70.07 & 0.61 \\ \cline{2-10}
\rowcolor[gray]{0.95}
 & \textbf{Ours (w/o GAN)} & {\color[HTML]{D83931} \textbf{27.05}} & {\color[HTML]{D83931} \textbf{0.78}} & {\color[HTML]{D83931} \textbf{0.24}} & {\color[HTML]{D83931} \textbf{0.20}} & {\color[HTML]{D83931} \textbf{104.74}} & 6.21 & 67.25 & 0.64 \\
\rowcolor[gray]{0.95}
\multirow{-17}{*}{\textbf{RealSR}} & \textbf{Ours (w/ GAN)} & 25.78 & 0.75 & {\color[HTML]{4E83FD} \textbf{0.26}} & {\color[HTML]{D83931} \textbf{0.20}} & {\color[HTML]{4E83FD} \textbf{106.22}} & {\color[HTML]{4E83FD} \textbf{5.29}} & 69.86 & {\color[HTML]{D83931} \textbf{0.68}} \\ \bottomrule
\end{tabular}%
}
\end{table*}

\begin{table*}[!t]\tiny
\setlength{\abovecaptionskip}{0.10cm}
\caption{Quantitative comparison on synthetic benchmarks with our default FLUX2-Klein (4B) variants. Higher is better for PSNR, SSIM, MUSIQ, and MANIQA, while lower is better for LPIPS, DISTS, FID, and NIQE. Within each benchmark, the first group lists GAN-based methods and the second group diffusion-based methods. The best and second-best results of each metric are highlighted in \textcolor{red}{red} and \textcolor{blue}{blue}, respectively. Results for the other ScaleResfusion backbones are consolidated in Appendix Table~\ref{tab:ablation_backbone_full}.}
\label{tab:multi-step-all-syn}
\centering
\renewcommand{\arraystretch}{1.08}
\setlength{\tabcolsep}{3.8pt}
\resizebox{\textwidth}{!}{%
\begin{tabular}{@{}c|c|cccccccc@{}}
\toprule
\textbf{Datasets} & \textbf{Method} & \textbf{PSNR $\uparrow$} & \textbf{SSIM $\uparrow$} & \textbf{LPIPS $\downarrow$} & \textbf{DISTS $\downarrow$} & \textbf{FID $\downarrow$} & \textbf{NIQE $\downarrow$} & \textbf{MUSIQ $\uparrow$} & \textbf{MANIQA $\uparrow$} \\ \midrule
 & BSRGAN & 24.58 & {\color[HTML]{4E83FD} \textbf{0.63}} & 0.35 & 0.23 & 49.55 & 4.75 & 61.68 & 0.50 \\
 & Real-ESRGAN & 24.02 & {\color[HTML]{D83931} \textbf{0.64}} & 0.32 & 0.21 & 38.87 & 4.83 & 60.38 & 0.54 \\
 & LDL & 23.83 & {\color[HTML]{4E83FD} \textbf{0.63}} & 0.33 & 0.22 & 42.28 & 4.86 & 60.04 & 0.53 \\
 & FeMaSR & 22.45 & 0.59 & 0.34 & 0.22 & 41.97 & 4.87 & 57.94 & 0.48 \\ \cline{2-10}
 & StableSR & 23.27 & 0.57 & 0.31 & 0.20 & 24.95 & 4.77 & 65.78 & 0.62 \\
 & SUPIR & 22.13 & 0.53 & 0.39 & 0.23 & 31.40 & 5.68 & 63.86 & 0.59 \\
 & TSD-SR & 23.02 & 0.58 & {\color[HTML]{4E83FD} \textbf{0.27}} & {\color[HTML]{4E83FD} \textbf{0.18}} & 29.16 & {\color[HTML]{D83931} \textbf{4.32}} & {\color[HTML]{D83931} \textbf{71.69}} & 0.62 \\
 & AddSR & 22.37 & 0.56 & 0.38 & 0.23 & 34.91 & 5.84 & 69.15 & 0.63 \\
 & CCSR & 24.30 & {\color[HTML]{4E83FD} \textbf{0.63}} & 0.30 & 0.20 & 30.84 & 5.34 & 69.53 & 0.61 \\
 & DiffBIR & 23.14 & 0.54 & 0.37 & 0.22 & 32.71 & 4.99 & {\color[HTML]{4E83FD} \textbf{69.87}} & {\color[HTML]{4E83FD} \textbf{0.64}} \\
 & OSEDiff & 23.72 & 0.61 & 0.29 & 0.20 & 26.34 & 4.71 & 67.96 & 0.61 \\
 & PASD & 24.01 & 0.61 & 0.38 & 0.22 & 37.06 & 4.98 & 63.75 & 0.55 \\
 & ResShift & {\color[HTML]{4E83FD} \textbf{24.59}} & 0.62 & 0.31 & 0.21 & 30.81 & 6.92 & 58.90 & 0.53 \\
 & SeeSR & 23.68 & 0.60 & 0.32 & 0.20 & 25.89 & 4.81 & 68.66 & 0.62 \\ \cline{2-10}
\rowcolor[gray]{0.95}
 & \textbf{Ours (w/o GAN)} & {\color[HTML]{D83931} \textbf{24.90}} & {\color[HTML]{D83931} \textbf{0.64}} & {\color[HTML]{4E83FD} \textbf{0.27}} & {\color[HTML]{4E83FD} \textbf{0.18}} & {\color[HTML]{4E83FD} \textbf{22.90}} & 5.09 & 66.10 & 0.63 \\
\rowcolor[gray]{0.95}
\multirow{-16}{*}{\textbf{DIV2K-Val}} & \textbf{Ours (w/ GAN)} & 23.63 & 0.60 & {\color[HTML]{D83931} \textbf{0.26}} & {\color[HTML]{D83931} \textbf{0.17}} & {\color[HTML]{D83931} \textbf{19.47}} & {\color[HTML]{4E83FD} \textbf{4.44}} & 69.30 & {\color[HTML]{D83931} \textbf{0.68}} \\ \midrule
 & BSRGAN & 20.82 & 0.54 & 0.25 & 0.16 & 46.37 & 4.21 & 68.94 & 0.63 \\
 & Real-ESRGAN & 20.58 & {\color[HTML]{4E83FD} \textbf{0.55}} & 0.24 & {\color[HTML]{4E83FD} \textbf{0.15}} & {\color[HTML]{4E83FD} \textbf{41.28}} & 4.18 & 69.52 & 0.64 \\
 & LDL & 20.31 & 0.53 & 0.25 & 0.16 & 44.75 & 4.36 & 68.61 & 0.64 \\
 & FeMaSR & 19.87 & 0.51 & 0.27 & 0.17 & 48.63 & 4.09 & 67.85 & 0.61 \\ \cline{2-10}
 & StableSR & 20.31 & {\color[HTML]{4E83FD} \textbf{0.55}} & 0.31 & 0.18 & 54.76 & 5.07 & 62.96 & 0.61 \\
 & SUPIR & 20.35 & 0.50 & 0.24 & {\color[HTML]{4E83FD} \textbf{0.15}} & 43.81 & 4.81 & 71.47 & 0.67 \\
 & TSD-SR & 19.05 & 0.49 & {\color[HTML]{D83931} \textbf{0.21}} & {\color[HTML]{D83931} \textbf{0.14}} & 45.66 & 3.86 & {\color[HTML]{D83931} \textbf{74.45}} & 0.68 \\
 & AddSR & 19.20 & 0.45 & 0.34 & 0.20 & 79.80 & 4.99 & {\color[HTML]{4E83FD} \textbf{74.20}} & {\color[HTML]{D83931} \textbf{0.70}} \\
 & CCSR & 20.76 & 0.53 & 0.26 & 0.16 & 56.28 & 4.25 & 72.57 & 0.66 \\
 & DiffBIR & 20.51 & 0.49 & 0.27 & 0.16 & 58.45 & 4.44 & 73.26 & 0.68 \\
 & OSEDiff & 20.39 & 0.52 & 0.27 & 0.16 & 59.57 & 4.03 & 72.34 & 0.66 \\
 & PASD & 20.93 & 0.52 & 0.31 & 0.17 & 59.11 & {\color[HTML]{D83931} \textbf{3.80}} & 69.29 & 0.62 \\
 & ResShift & {\color[HTML]{4E83FD} \textbf{21.23}} & {\color[HTML]{4E83FD} \textbf{0.55}} & {\color[HTML]{4E83FD} \textbf{0.23}} & {\color[HTML]{D83931} \textbf{0.14}} & {\color[HTML]{D83931} \textbf{38.98}} & 5.32 & 68.56 & 0.61 \\
 & SeeSR & 20.69 & 0.52 & 0.25 & {\color[HTML]{4E83FD} \textbf{0.15}} & 52.06 & 4.10 & 73.27 & 0.68 \\ \cline{2-10}
\rowcolor[gray]{0.95}
 & \textbf{Ours (w/o GAN)} & {\color[HTML]{D83931} \textbf{21.35}} & {\color[HTML]{D83931} \textbf{0.57}} & {\color[HTML]{4E83FD} \textbf{0.23}} & {\color[HTML]{D83931} \textbf{0.14}} & 41.69 & 4.21 & 70.61 & 0.67 \\
\rowcolor[gray]{0.95}
\multirow{-16}{*}{\textbf{LSDIR-Val}} & \textbf{Ours (w/ GAN)} & 20.64 & {\color[HTML]{4E83FD} \textbf{0.55}} & {\color[HTML]{D83931} \textbf{0.21}} & {\color[HTML]{D83931} \textbf{0.14}} & 43.56 & {\color[HTML]{4E83FD} \textbf{3.82}} & 72.59 & {\color[HTML]{4E83FD} \textbf{0.69}} \\ \bottomrule
\end{tabular}%
}
\end{table*}

\begin{table*}[!t]\tiny
\setlength{\abovecaptionskip}{0.1cm}
\caption{Efficiency comparison. We report the number of function evaluations (NFE) and the average inference time per $512\times512$ image, measured on a single NVIDIA RTX A6000 GPU (measurement protocol in Appendix~\ref{sec:appendix_experimental_setting_details}).}
\label{tab:efficiency}
\centering
\renewcommand{\arraystretch}{1.08}
\setlength{\tabcolsep}{13pt}
\resizebox{\textwidth}{!}{%
\begin{tabular}{@{}l|cccccc|c@{}}
\toprule
\textbf{Method} & StableSR & SUPIR & SeeSR & ResShift & CCSR & OSEDiff & \textbf{Ours} \\ \midrule
\textbf{NFE} & 200 & 50 & 50 & 15 & 6 & 1 & \textbf{4} \\
\textbf{Inf. Time (ms)} & 12,036 & 25,252 & 4,445 & 848 & 516 & 266 & \textbf{646} \\ \bottomrule
\end{tabular}%
}
\end{table*}

\subsection{Knowledge-Distilled Parameter-Efficient Training Pipeline}
\label{sec: training pipeline}
Based on the residual theory above, we design \textbf{ScaleResfusion}, a parameter-efficient training pipeline that distills pre-trained diffusion priors into an RRF generator. All components of this pipeline are implementation choices organized around the RRF transport structure: LoRA realizes the compact residual correction, the Feature Extractor and DAPE inject observation-derived features and text prompts as conditioning, and the distillation losses regularize few-step training; none of them modifies the RRF formulation itself. As shown in Fig.~\ref{fig:ScaleResfusion}, ScaleResfusion contains a generation branch and a regularization branch. Both branches share the same frozen pre-trained backbone weights, while trainable LoRA adapters learn the residual task update.

Given an LQ--HQ pair $(\hat{x}_0,x_0)$, we encode images into the latent space with a frozen VAE and train a LoRA-adapted generator $G_\theta$ to predict the residual velocity $resv_\theta=G_\theta(x_t,t)$. The LQ features are provided by a \textbf{Feature Extractor}: essentially a ReferenceNet~\cite{hu2024animate} that extracts reference features from the LQ input and injects them into the restored-image generator through Ref Attention~\cite{shi2025ultra}. We further use DAPE~\cite{wu2024seesr} to provide additional textual constraints. For notation simplicity, $x_t$ denotes the latent RRF state, and image-space losses are computed after VAE decoding. We convert the residual velocity output into an equivalent prediction to compute losses on the predicted clean restoration, where the VAE decoder is omitted in notation for image-space losses:
\begin{equation}
\label{eq:x0_pred}
x_0^{pred}=x_t-t\,resv_\theta(x_t,t).
\end{equation}

\textbf{Data term.}
The data objective enforces pairwise fidelity and perceptual consistency:
\begin{equation}
\mathcal{L}_{data}
=\lambda_{mse}\mathcal{L}_{MSE}(x_0^{pred},x_0)
+\lambda_{lpips}\mathcal{L}_{LPIPS}(x_0^{pred},x_0).
\end{equation}
Here $\mathcal{L}_{MSE}$ penalizes pixel-wise reconstruction errors and preserves the pairwise fidelity to the HQ target, while $\mathcal{L}_{LPIPS}$ measures perceptual similarity in a deep feature space~\cite{zhang2018unreasonable} to encourage visually consistent structures and textures.

\textbf{Regularization term.}
To align the restored distribution with natural HQ images, we introduce a regularization term that combines a DMD loss and an optional GAN loss:
\begin{equation}
\mathcal{L}_{reg}
=\lambda_{dmd}\mathcal{L}_{DMD}(x_0^{pred})
+\lambda_{gan}\mathcal{L}_{GAN}(x_0^{pred}).
\end{equation}
For one-step or few-step generators adapted from pre-trained diffusion models, such distribution-level regularization is a generic training requirement: without it, few-step outputs tend to be over-smoothed and to overfit perceptual similarity metrics. We adopt DMD as the default regularizer since it is the overall strongest choice in our experiments, while a GAN-only regularizer also yields competitive results (Sec.~\ref{sec:ablation}), indicating that RRF is not tied to a specific regularizer. The DMD term serves as a knowledge-distillation objective following the real/fake regularizer design~\cite{yin2024one}. Specifically, the frozen pre-trained regularizer acts as the teacher that provides the real diffusion prior, and the RRF generator is the student that learns to produce samples aligned with this prior. The LoRA-adapted fake regularizer is trained with $\mathcal{L}_{diff}$ on generated samples to estimate the generator-induced distribution, providing the fake score needed by DMD. Through this teacher--student distillation, $\mathcal{L}_{DMD}$ transfers the natural-image prior of the pre-trained diffusion model to the RRF generator and encourages restored images to stay on the HQ image manifold. Notably, $\mathcal{L}_{reg}$ constrains only the model output rather than the sampling trajectory: both terms act on the clean prediction $x_0^{pred}$ in Eq.~\eqref{eq:x0_pred}, so the teacher perturbs and scores ordinary clean-image samples along its native Gaussian noise-to-image trajectory. The optional GAN term further sharpens local details by adversarially matching the restored distribution to real HQ images.

\section{Experiment}
\label{sec:experiment}
\noindent
{\bf Datasets.}
Following~\cite{wang2024exploiting, dong2025tsd, wu2024one}, we use LSDIR~\cite{li2023lsdir} and the first 10K face images from FFHQ \cite{karras2019style} for training.
LR-HR pairs are synthesized with the Real-ESRGAN degradation pipeline~\cite{wang2021real, dong2025tsd, wu2024one}.
We use RealSR~\cite{cai2019toward}, DRealSR~\cite{wei2020component}, and DIV2K-Val~\cite{agustsson2017ntire} for evaluation, following the same setup as StableSR~\cite{wang2024exploiting}. Additionally, we employ a more challenging benchmark constructed by center-cropping $512\times512$ patches from the original LSDIR-Val split~\cite{li2023lsdir} to further compare the performance across models.

\noindent
{\bf Evaluation Metrics.}
To verify the performance of different models, we employ structural metrics, perceptual metrics, distribution consistency metrics, and no-reference metrics. Specifically, we report PSNR and SSIM \cite{wang2004image} (computed on the Y channel of the YCbCr color space) for distortion fidelity, LPIPS \cite{zhang2018unreasonable} and DISTS \cite{ding2020image} for perceptual similarity, FID \cite{heusel2017gans} for distribution alignment, and NIQE \cite{mittal2012making}, MUSIQ \cite{ke2021musiq}, and MANIQA \cite{yang2022maniqa} as no-reference quality indicators.

\noindent
{\bf Implementation Details.}
We train ScaleResfusion models with SD3 (2B)~\cite{esser2024scaling}, FLUX2-Klein (4B)~\cite{flux-2-2025}, Z-Image (6B)~\cite{cai2025z}, and FLUX2-Klein (9B) as backbones. By default, we use 4-step sampling (we discuss the effects of different sampling steps in Sec.~\ref{sec:ablation}) and the residual ratio $\gamma=1$ (we discuss the impact of different noise-to-signal ratio initializations in Appendix~D-A). We employ RAM-based~\cite{zhang2024recognize} DAPE following OSEDiff~\cite{wu2024one} (we discuss the effects of the DAPE in the Appendix~D-B). We use a DINOv2-based~\cite{oquab2023dinov2} GAN discriminator identical to AddSR~\cite{xie2024addsr}. Detailed experimental settings are provided in Appendix~B.

\begin{figure*}[t]
\center
\includegraphics[width=0.95\textwidth]{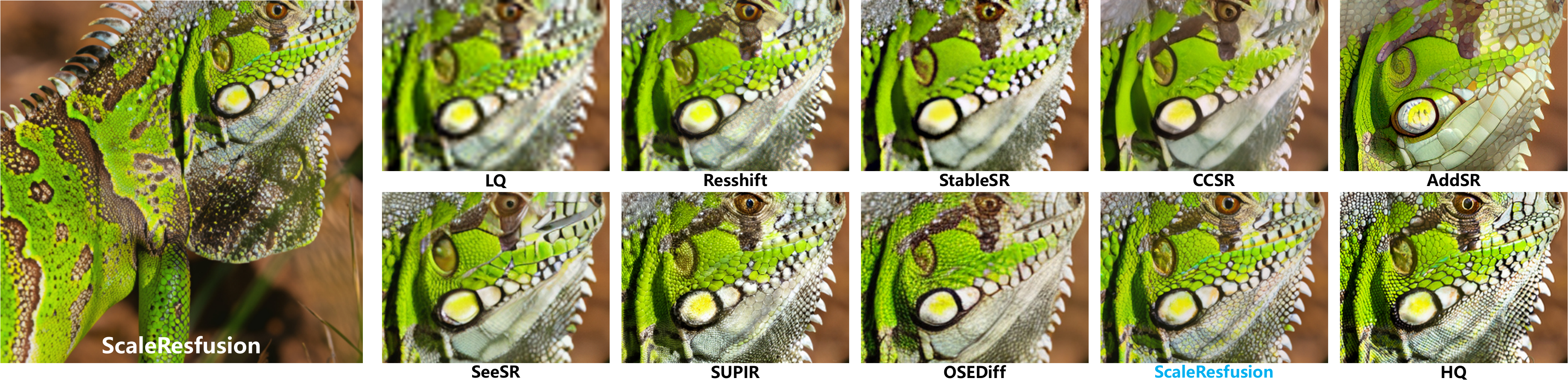}
\caption{Visual comparison with existing diffusion-based restoration methods. The left side shows the result of ScaleResfusion, while the right side compares local crops. Overall, ScaleResfusion better preserves the global structure while recovering more natural details without the over-smoothing or hallucinated patterns observed in competing methods.}
\label{fig:lsdir_1}
\end{figure*}

\noindent
{\bf Comparison with Existing Methods.}
Table~\ref{tab:multi-step-all-real} and Table~\ref{tab:multi-step-all-syn} report the quantitative comparison on real-world benchmarks (DRealSR and RealSR) and synthetic benchmarks (DIV2K-Val and LSDIR-Val). We compare ScaleResfusion with GAN-based methods, including BSRGAN~\cite{zhang2021designing}, Real-ESRGAN~\cite{wang2021real}, LDL~\cite{liang2022details}, and FeMaSR~\cite{chen2022real}, as well as diffusion-based methods, including StableSR~\cite{wang2024exploiting}, SUPIR~\cite{yu2024scaling}, TSD-SR~\cite{dong2025tsd}, AddSR~\cite{xie2024addsr}, CCSR~\cite{sun2023ccsr}, DiffBIR~\cite{lin2024diffbir}, OSEDiff~\cite{wu2024one}, PASD~\cite{yang2024pixel}, ResShift~\cite{yue2023resshift}, SeeSR~\cite{wu2024seesr}, and FluxSR~\cite{li2025one}. These methods cover adversarial restoration models, iterative diffusion models, variants that start diffusion from noisy LQ images, and efficient one-step or few-step generators, providing a broad comparison for real-world super-resolution. Note that FluxSR is not open-source and synthesizes its DIV2K-Val inputs with its own degradation pipeline; we therefore evaluate it only on the real-world benchmarks, using its released restoration results under standard protocol. We report our default FLUX2-Klein (4B) variants with and without GAN fine-tuning, and consolidate the results of the other backbones in Sec.~\ref{sec:ablation} (Table~\ref{tab:ablation_backbone}). We summarize the main observations as follows:

(1) On real-world benchmarks, ScaleResfusion achieves the strongest overall fidelity and distribution alignment. On DRealSR, the \emph{w/o GAN} variant simultaneously achieves the best PSNR, SSIM, LPIPS, DISTS, and FID, e.g., 29.77 vs. 28.70 in PSNR and 118.18 vs. 134.98 in FID against the best competing method. On RealSR, it likewise achieves the best PSNR, SSIM, LPIPS, DISTS, and FID. Scaling the backbone to FLUX2-Klein (9B) further improves PSNR, DISTS, and FID, reaching 30.17 PSNR and 109.28 FID on DRealSR (Table~\ref{tab:ablation_backbone}). These results indicate that the proposed residual rectified-flow formulation does not merely improve perceptual appearance, but also preserves the degraded input content more faithfully than previous restoration-oriented diffusion models.

(2) On synthetic benchmarks, ScaleResfusion remains consistently competitive under both DIV2K-Val and LSDIR-Val. On DIV2K-Val, the \emph{w/o GAN} variant achieves the best PSNR and tied-best SSIM, and the \emph{w/ GAN} variant obtains the best LPIPS, DISTS, and FID. On the more challenging LSDIR-Val benchmark, the \emph{w/o GAN} variant provides the best PSNR and SSIM, the \emph{w/ GAN} variant attains tied-best LPIPS and DISTS. In contrast, competing methods usually perform well only on a subset of metrics. For example, some diffusion-based methods obtain strong no-reference scores but suffer from weaker distortion fidelity, whereas GAN-based methods can preserve local structure but often show inferior distribution alignment.

(3) ScaleResfusion variants with and without GAN fine-tuning exhibit a clear perception--distortion trade-off~\cite{blau2018perception,zhang2022perception,luo2024skipdiff}. On all four benchmarks, the \emph{w/o GAN} variant delivers stronger reconstruction fidelity, whereas GAN fine-tuning consistently improves the no-reference quality scores at the cost of a moderate decrease in pixel-level fidelity. This behavior is consistent with prior findings~\cite{xie2024addsr,yu2024scaling} and explains why we report both variants: users can select the more faithful or the more perceptual model depending on the target application. Consistent transfer across the four pre-trained backbones is analyzed in Sec.~\ref{sec:ablation}.

\noindent
{\bf Efficiency Comparison.}
Table~\ref{tab:efficiency} compares the number of function evaluations (NFE) and the average per-image inference time. With only 4 sampling steps, ScaleResfusion runs at 646 ms per $512\times512$ image, about $6.9\times$ faster than SeeSR and $18.6\times$ faster than StableSR, while SUPIR requires more than 25 seconds per image. OSEDiff (266 ms) and CCSR (516 ms) are faster, but OSEDiff learns a deterministic single-step LQ-to-HQ mapping that abandons multi-step sampling, and CCSR trails on perceptual and distribution metrics (Table~\ref{tab:multi-step-all-real}); ScaleResfusion instead retains the stochastic sampling capability of diffusion models at sub-second latency.

\begin{table*}[!t]\tiny
\setlength{\abovecaptionskip}{0.1cm}
\caption{Effectiveness of Residual Rectified Flow. Under identical FLUX2-Klein (4B) settings, the three variants differ only in the forward transport path: Standard RF ($x_t=(1-t)x_0+t\epsilon$), a direct LQ$\rightarrow$HQ flow ($x_t=(1-t)x_0+t\hat{x}_0$), and the proposed RRF ($x_t=(1-t)x_0+t\gamma R+t\epsilon$). Within each benchmark, the best and second-best results of each metric are highlighted in \textcolor{red}{red} and \textcolor{blue}{blue}.}
\label{tab:ablation_rrf}
\centering
\renewcommand{\arraystretch}{1.08}
\setlength{\tabcolsep}{9pt}
\resizebox{\textwidth}{!}{%
\begin{tabular}{@{}c|l|cccccccc@{}}
\toprule
\textbf{Dataset} & \textbf{Strategy} & \textbf{PSNR $\uparrow$} & \textbf{SSIM $\uparrow$} & \textbf{LPIPS $\downarrow$} & \textbf{DISTS $\downarrow$} & \textbf{FID $\downarrow$} & \textbf{NIQE $\downarrow$} & \textbf{MUSIQ $\uparrow$} & \textbf{MANIQA $\uparrow$} \\ \midrule
 & Standard RF & 26.83 & 0.72 & 0.35 & 0.25 & 137.58 & 7.82 & {\color[HTML]{4E83FD} \textbf{58.40}} & 0.56 \\
 & LQ$\rightarrow$HQ Flow & {\color[HTML]{4E83FD} \textbf{28.03}} & {\color[HTML]{4E83FD} \textbf{0.74}} & {\color[HTML]{4E83FD} \textbf{0.31}} & {\color[HTML]{4E83FD} \textbf{0.23}} & {\color[HTML]{4E83FD} \textbf{131.61}} & {\color[HTML]{D83931} \textbf{6.16}} & 55.64 & {\color[HTML]{4E83FD} \textbf{0.57}} \\
\multirow{-3}{*}{DRealSR} & RRF (ours) & {\color[HTML]{D83931} \textbf{29.77}} & {\color[HTML]{D83931} \textbf{0.82}} & {\color[HTML]{D83931} \textbf{0.25}} & {\color[HTML]{D83931} \textbf{0.20}} & {\color[HTML]{D83931} \textbf{118.18}} & {\color[HTML]{4E83FD} \textbf{6.95}} & {\color[HTML]{D83931} \textbf{62.42}} & {\color[HTML]{D83931} \textbf{0.61}} \\ \midrule
 & Standard RF & {\color[HTML]{4E83FD} \textbf{25.61}} & {\color[HTML]{4E83FD} \textbf{0.75}} & {\color[HTML]{4E83FD} \textbf{0.25}} & {\color[HTML]{D83931} \textbf{0.20}} & {\color[HTML]{4E83FD} \textbf{109.64}} & 6.41 & {\color[HTML]{4E83FD} \textbf{65.22}} & {\color[HTML]{D83931} \textbf{0.64}} \\
 & LQ$\rightarrow$HQ Flow & 25.28 & 0.69 & 0.30 & 0.22 & 113.83 & {\color[HTML]{D83931} \textbf{5.42}} & 60.96 & 0.61 \\
\multirow{-3}{*}{RealSR} & RRF (ours) & {\color[HTML]{D83931} \textbf{27.05}} & {\color[HTML]{D83931} \textbf{0.78}} & {\color[HTML]{D83931} \textbf{0.24}} & {\color[HTML]{D83931} \textbf{0.20}} & {\color[HTML]{D83931} \textbf{104.74}} & {\color[HTML]{4E83FD} \textbf{6.21}} & {\color[HTML]{D83931} \textbf{67.25}} & {\color[HTML]{D83931} \textbf{0.64}} \\ \midrule
 & Standard RF & 22.60 & 0.55 & 0.35 & {\color[HTML]{4E83FD} \textbf{0.23}} & {\color[HTML]{4E83FD} \textbf{37.58}} & {\color[HTML]{4E83FD} \textbf{4.96}} & {\color[HTML]{4E83FD} \textbf{64.48}} & {\color[HTML]{4E83FD} \textbf{0.61}} \\
 & LQ$\rightarrow$HQ Flow & {\color[HTML]{4E83FD} \textbf{23.29}} & {\color[HTML]{4E83FD} \textbf{0.56}} & {\color[HTML]{4E83FD} \textbf{0.33}} & 0.24 & 38.17 & {\color[HTML]{D83931} \textbf{4.74}} & 58.70 & 0.59 \\
\multirow{-3}{*}{DIV2K-Val} & RRF (ours) & {\color[HTML]{D83931} \textbf{24.90}} & {\color[HTML]{D83931} \textbf{0.64}} & {\color[HTML]{D83931} \textbf{0.27}} & {\color[HTML]{D83931} \textbf{0.18}} & {\color[HTML]{D83931} \textbf{22.90}} & 5.09 & {\color[HTML]{D83931} \textbf{66.10}} & {\color[HTML]{D83931} \textbf{0.63}} \\ \bottomrule
\end{tabular}%
}
\end{table*}

\begin{figure}[t]
    \centering
    \includegraphics[width=\columnwidth]{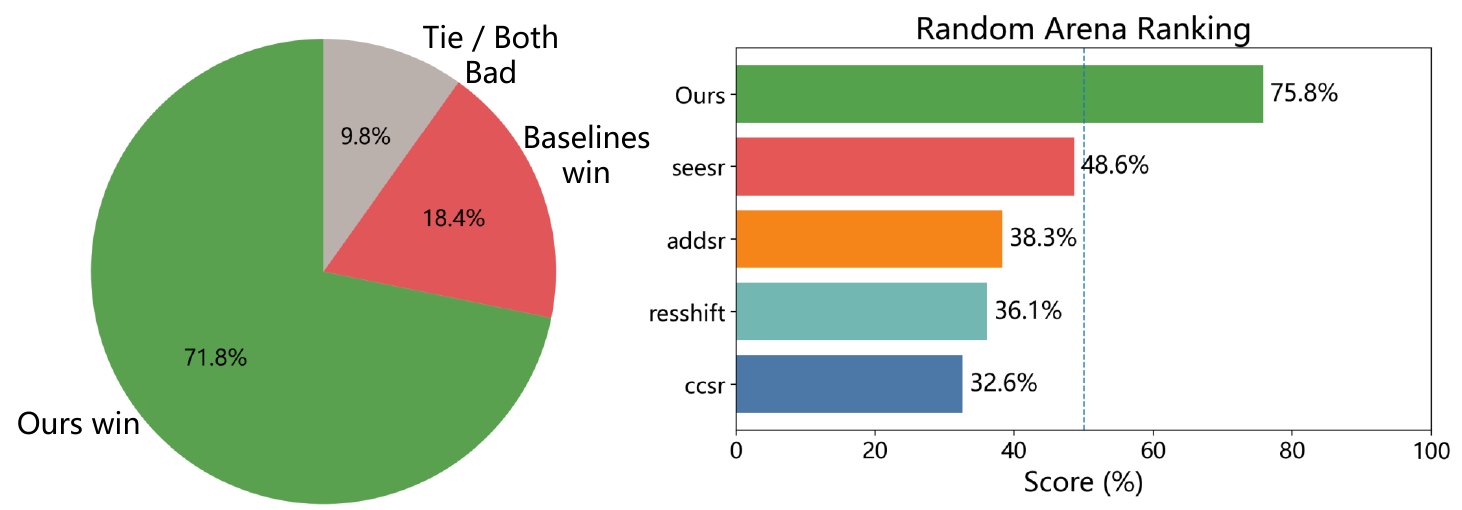}
    \caption{Pairwise Arena user study results. Left: preference distribution when comparing ScaleResfusion against other methods, where ScaleResfusion wins 71.8\% of comparisons. Right: Random Arena ranking, where ScaleResfusion achieves the highest overall score of 75.8\%, indicating consistently stronger user preference.}
    \label{fig:arena}
\end{figure}

\noindent
{\bf User Study.} We conduct a user study following a pairwise \emph{model arena} protocol: in each round, a participant is shown an LQ input together with the restored results of two anonymized models and selects the preferred one (or marks a tie / both bad). The test images cover benchmarks with ground truth (e.g., LSDIR) as well as in-the-wild images without ground truth (WebPhoto-Test), the latter probing out-of-distribution robustness. In total, 48 valid independent evaluators contribute 1,756 pairwise comparisons. As shown in Fig.~\ref{fig:arena}, in cross valid comparisons, ScaleResfusion is preferred in 71.8\% of cases, compared with 18.4\% for baselines and 9.8\% for Tie/BothBad. It also achieves a 75.8\% overall score in the Random Arena ranking, clearly outperforming all baselines. Participant feedback further shows that users mainly attend to text correctness, detail richness, object consistency, artifacts, and material realism, whereas color consistency is less noticed, since the restored images remain close to the ground truth under strong degradation.

\section{Ablation Studies}
\label{sec:ablation}

\begin{figure}[!t]
\centering
\includegraphics[width=0.9\columnwidth]{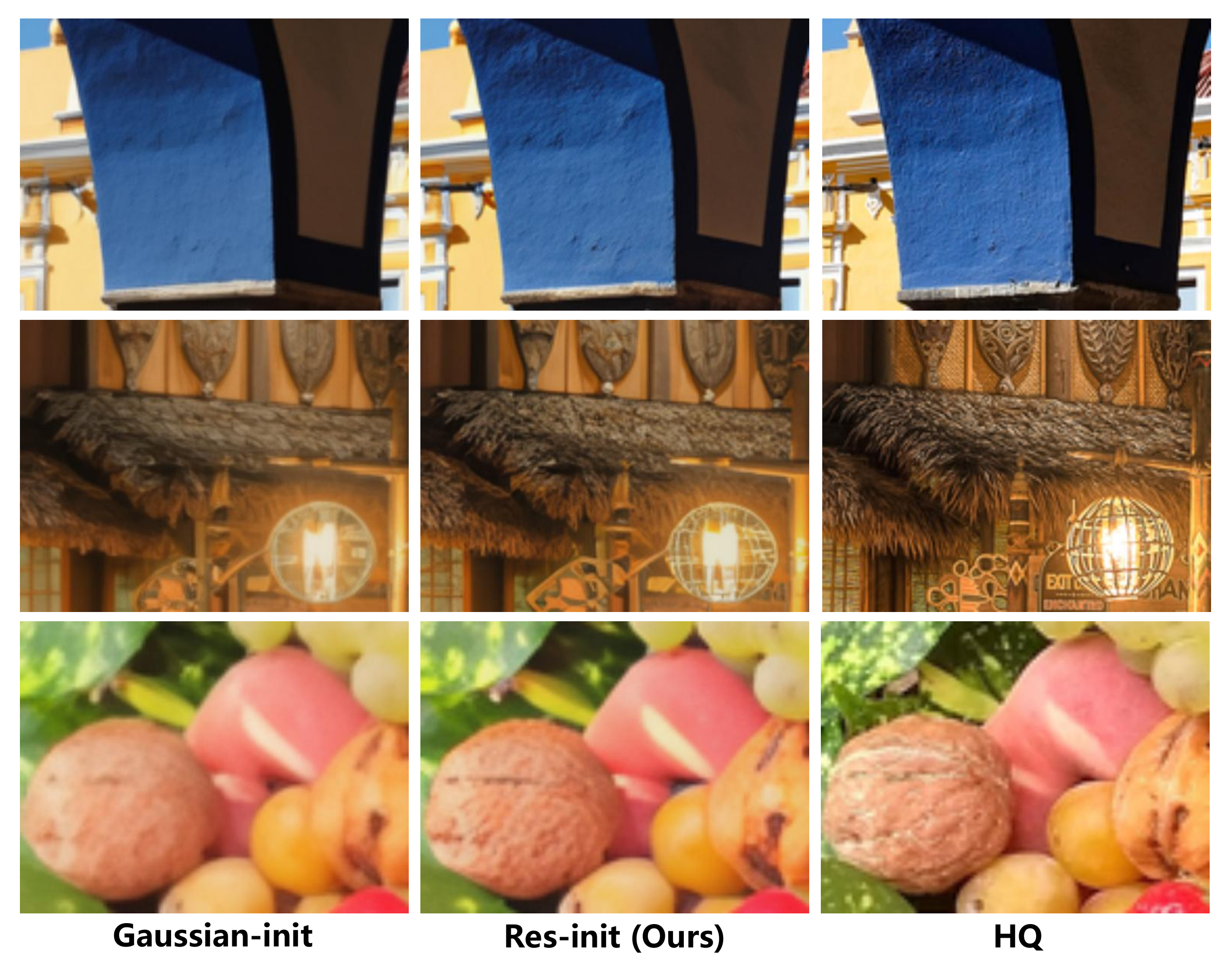}
\caption{Visual ablation on residual initialization. Compared with task-agnostic Gaussian initialization, residual initialization starts from a more restoration-aligned state and better preserves the observed image structure.}
\Description{Visual comparison between Gaussian initialization and residual initialization.}
\label{fig:ablation_resinit}
\end{figure}
\textbf{Effectiveness of RRF.} We first isolate the transport formulation itself. Under identical FLUX2-4B settings (same backbone, data, conditioning, distillation pipeline, and 4-step sampling), we train three variants that differ only in the forward path: Standard RF, which transports Gaussian noise to HQ images; a direct LQ$\rightarrow$HQ flow, which deterministically transports the LQ image to its HQ counterpart; and the proposed RRF. As shown in Table~\ref{tab:ablation_rrf}, RRF achieves the best or tied-best results on every fidelity, perceptual-similarity, and distribution metric across the three benchmarks, e.g., $+2.94$ dB PSNR and $-19.40$ FID over Standard RF on DRealSR. Standard RF must synthesize both global structure and details from a task-agnostic state, weakening detail consistency; the deterministic LQ$\rightarrow$HQ flow anchors the start to the observation but removes the stochastic component, reducing diversity and causing over-smoothing, as reflected by its clearly lower MUSIQ---empirically confirming the degeneration into a plain LQ--HQ mapping discussed in Sec.~\ref{sec:intro}. RRF combines both advantages: the observation anchors the structure while the noise retains generative sampling. The conclusion also transfers across backbones: with SD3, replacing Gaussian initialization by the residual start improves PSNR from 26.92 to 28.77 on DRealSR and from 17.87 to 20.22 on LSDIR-Val (visual comparisons in Fig.~\ref{fig:ablation_resinit}).

\begin{table*}[!t]\tiny
\setlength{\abovecaptionskip}{0.1cm}
\caption{Module ablation of ScaleResfusion on the FLUX2-Klein (4B) backbone, with results reported on DIV2K-Val. The best and second-best results of each metric are highlighted in \textcolor{red}{red} and \textcolor{blue}{blue}.}
\label{tab:ablation_module}
\centering
\renewcommand{\arraystretch}{1.08}
\setlength{\tabcolsep}{3.8pt}
\resizebox{\textwidth}{!}{%
\begin{tabular}{@{}cccc|cccccccc@{}}
\toprule
\multicolumn{4}{c|}{\textbf{Module}} & \multirow{2}{*}{\textbf{PSNR $\uparrow$}} & \multirow{2}{*}{\textbf{SSIM $\uparrow$}} & \multirow{2}{*}{\textbf{LPIPS $\downarrow$}} & \multirow{2}{*}{\textbf{DISTS $\downarrow$}} & \multirow{2}{*}{\textbf{FID $\downarrow$}} & \multirow{2}{*}{\textbf{NIQE $\downarrow$}} & \multirow{2}{*}{\textbf{MUSIQ $\uparrow$}} & \multirow{2}{*}{\textbf{MANIQA $\uparrow$}} \\
\textbf{Residual initialization} & \textbf{Feature Extractor} & \textbf{DMD Regularizer} & \textbf{GAN Finetune} &  &  &  &  &  &  &  &  \\ \midrule
$\times$ & $\checkmark$ & $\checkmark$ & $\times$ & 22.60 & 0.55 & 0.35 & 0.23 & 37.58 & {\color[HTML]{4E83FD} \textbf{4.96}} & 64.48 & 0.61 \\
$\checkmark$ & $\times$ & $\checkmark$ & $\times$ & 23.52 & {\color[HTML]{4E83FD} \textbf{0.60}} & 0.28 & 0.21 & 23.93 & 5.28 & 65.20 & 0.61 \\
$\checkmark$ & $\checkmark$ & $\times$ & $\times$ & 21.90 & 0.51 & 0.38 & 0.24 & 51.25 & 6.16 & 63.88 & 0.59 \\
$\checkmark$ & $\checkmark$ & $\checkmark$ & $\times$ & {\color[HTML]{D83931} \textbf{24.90}} & {\color[HTML]{D83931} \textbf{0.64}} & {\color[HTML]{4E83FD} \textbf{0.27}} & {\color[HTML]{4E83FD} \textbf{0.18}} & {\color[HTML]{4E83FD} \textbf{22.90}} & 5.09 & {\color[HTML]{4E83FD} \textbf{66.10}} & {\color[HTML]{4E83FD} \textbf{0.63}} \\
$\checkmark$ & $\checkmark$ & $\checkmark$ & $\checkmark$ & {\color[HTML]{4E83FD} \textbf{23.63}} & {\color[HTML]{4E83FD} \textbf{0.60}} & {\color[HTML]{D83931} \textbf{0.26}} & {\color[HTML]{D83931} \textbf{0.17}} & {\color[HTML]{D83931} \textbf{19.47}} & {\color[HTML]{D83931} \textbf{4.44}} & {\color[HTML]{D83931} \textbf{69.30}} & {\color[HTML]{D83931} \textbf{0.68}} \\ \bottomrule
\end{tabular}%
}
\end{table*}

\begin{figure*}[t]
\centering
\includegraphics[width=\textwidth]{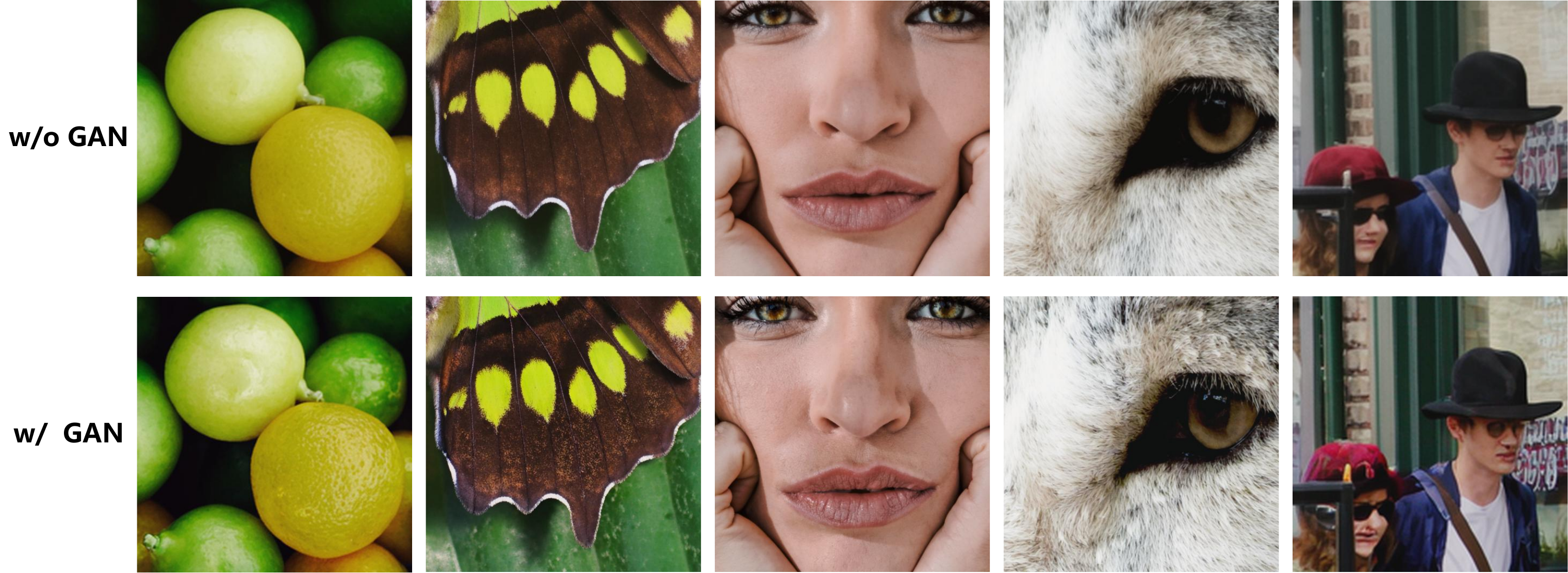}
\caption{Visual ablation on GAN fine-tuning. GAN fine-tuning enhances perceptual sharpness and local details, while occasionally reducing content consistency with the input or reference.}
\Description{Visual comparison of restoration results with and without GAN fine-tuning.}
\label{fig:ablation_gan}
\end{figure*}

\textbf{Module Ablation.} Table~\ref{tab:ablation_module} evaluates the RRF residual initialization together with the components built around it (Feature Extractor, DMD, GAN) on FLUX2 (4B). Removing the DMD regularizer causes the largest degradation (24.90 to 21.90 in PSNR, 22.90 to 51.25 in FID): for few-step generators adapted from pre-trained diffusion models, distribution-level regularization is a generic training requirement rather than a component specific to RRF (Sec.~\ref{sec: training pipeline}); we analyze regularizer choices below. Removing residual initialization causes the second-largest drop (24.90 to 22.60 in PSNR)---this variant is exactly the Standard RF transport in Table~\ref{tab:ablation_rrf}, so the effect is isolated from the regularizer. Removing the Feature Extractor also degrades PSNR, SSIM, and perceptual metrics, showing the benefit of injecting LQ reference features. With all three enabled, ScaleResfusion achieves the best distortion-oriented performance, and GAN fine-tuning further improves perceptual and distribution quality at the expected perception--distortion cost (Appendix Fig.~\ref{fig:ablation_gan}).

\begin{table*}[!t]\tiny
\setlength{\abovecaptionskip}{0.1cm}
\caption{Ablation on the training regularizer with the FLUX2-4B backbone. Res-init and the feature extractor are enabled for all variants, and GAN fine-tuning is disabled. \emph{None} removes the distribution-level regularizer, \emph{GAN} uses only a DINOv2-based discriminator (22.1M parameters) as the regularizer, and \emph{DMD} is our default setting. Within each benchmark, the best and second-best results of each metric are highlighted in \textcolor{red}{red} and \textcolor{blue}{blue}.}
\label{tab:ablation_regularizer}
\centering
\renewcommand{\arraystretch}{1.08}
\setlength{\tabcolsep}{11pt}
\resizebox{\textwidth}{!}{%
\begin{tabular}{@{}c|l|cccccccc@{}}
\toprule
\textbf{Dataset} & \textbf{Regularizer} & \textbf{PSNR $\uparrow$} & \textbf{SSIM $\uparrow$} & \textbf{LPIPS $\downarrow$} & \textbf{DISTS $\downarrow$} & \textbf{FID $\downarrow$} & \textbf{NIQE $\downarrow$} & \textbf{MUSIQ $\uparrow$} & \textbf{MANIQA $\uparrow$} \\ \midrule
 & None & 26.45 & 0.70 & 0.37 & 0.26 & 145.30 & 7.96 & 57.20 & 0.55 \\
 & GAN & {\color[HTML]{4E83FD} \textbf{28.79}} & {\color[HTML]{4E83FD} \textbf{0.78}} & {\color[HTML]{4E83FD} \textbf{0.29}} & {\color[HTML]{D83931} \textbf{0.20}} & {\color[HTML]{4E83FD} \textbf{125.70}} & {\color[HTML]{D83931} \textbf{6.27}} & {\color[HTML]{D83931} \textbf{62.67}} & {\color[HTML]{4E83FD} \textbf{0.58}} \\
\multirow{-3}{*}{DRealSR} & DMD & {\color[HTML]{D83931} \textbf{29.77}} & {\color[HTML]{D83931} \textbf{0.82}} & {\color[HTML]{D83931} \textbf{0.25}} & {\color[HTML]{D83931} \textbf{0.20}} & {\color[HTML]{D83931} \textbf{118.18}} & {\color[HTML]{4E83FD} \textbf{6.95}} & {\color[HTML]{4E83FD} \textbf{62.42}} & {\color[HTML]{D83931} \textbf{0.61}} \\ \midrule
 & None & 24.98 & 0.72 & 0.28 & 0.22 & 116.85 & 6.92 & 62.40 & 0.60 \\
 & GAN & {\color[HTML]{4E83FD} \textbf{26.29}} & {\color[HTML]{4E83FD} \textbf{0.75}} & {\color[HTML]{4E83FD} \textbf{0.25}} & {\color[HTML]{D83931} \textbf{0.19}} & {\color[HTML]{D83931} \textbf{102.53}} & {\color[HTML]{D83931} \textbf{5.37}} & {\color[HTML]{4E83FD} \textbf{67.21}} & {\color[HTML]{4E83FD} \textbf{0.63}} \\
\multirow{-3}{*}{RealSR} & DMD & {\color[HTML]{D83931} \textbf{27.05}} & {\color[HTML]{D83931} \textbf{0.78}} & {\color[HTML]{D83931} \textbf{0.24}} & {\color[HTML]{4E83FD} \textbf{0.20}} & {\color[HTML]{4E83FD} \textbf{104.74}} & {\color[HTML]{4E83FD} \textbf{6.21}} & {\color[HTML]{D83931} \textbf{67.25}} & {\color[HTML]{D83931} \textbf{0.64}} \\ \midrule
 & None & 21.90 & 0.51 & 0.38 & 0.24 & 51.25 & 6.16 & 63.88 & 0.59 \\
 & GAN & {\color[HTML]{4E83FD} \textbf{24.03}} & {\color[HTML]{4E83FD} \textbf{0.61}} & {\color[HTML]{D83931} \textbf{0.27}} & {\color[HTML]{D83931} \textbf{0.17}} & {\color[HTML]{D83931} \textbf{20.38}} & {\color[HTML]{D83931} \textbf{4.56}} & {\color[HTML]{D83931} \textbf{67.25}} & {\color[HTML]{4E83FD} \textbf{0.61}} \\
\multirow{-3}{*}{DIV2K-Val} & DMD & {\color[HTML]{D83931} \textbf{24.90}} & {\color[HTML]{D83931} \textbf{0.64}} & {\color[HTML]{D83931} \textbf{0.27}} & {\color[HTML]{4E83FD} \textbf{0.18}} & {\color[HTML]{4E83FD} \textbf{22.90}} & {\color[HTML]{4E83FD} \textbf{5.09}} & {\color[HTML]{4E83FD} \textbf{66.10}} & {\color[HTML]{D83931} \textbf{0.63}} \\ \bottomrule
\end{tabular}%
}
\end{table*}

\textbf{Training Regularizer.} We compare regularizer choices under identical conditions (residual initialization and LQ conditioning enabled, GAN fine-tuning disabled): \emph{None}, a lightweight DINOv2-based \emph{GAN} discriminator (22.1M parameters), and \emph{DMD}. Table~\ref{tab:ablation_regularizer} shows two findings. First, distribution-level regularization is indispensable for few-step adaptation: removing it degrades every metric on all three benchmarks, since unregularized few-step outputs become over-smoothed and overfit perceptual similarity metrics. Second, the benefit is not tied to DMD: the GAN-only regularizer recovers most of the gap and is stronger on some perceptual metrics, while DMD attains the best fidelity on all three benchmarks. We therefore adopt DMD as the overall strongest choice, while the RRF formulation itself remains compatible with different regularizers (Appendix Fig.~\ref{fig:ablation_vsd}).

\begin{figure*}[!t]
\centering
\includegraphics[width=0.85\textwidth]{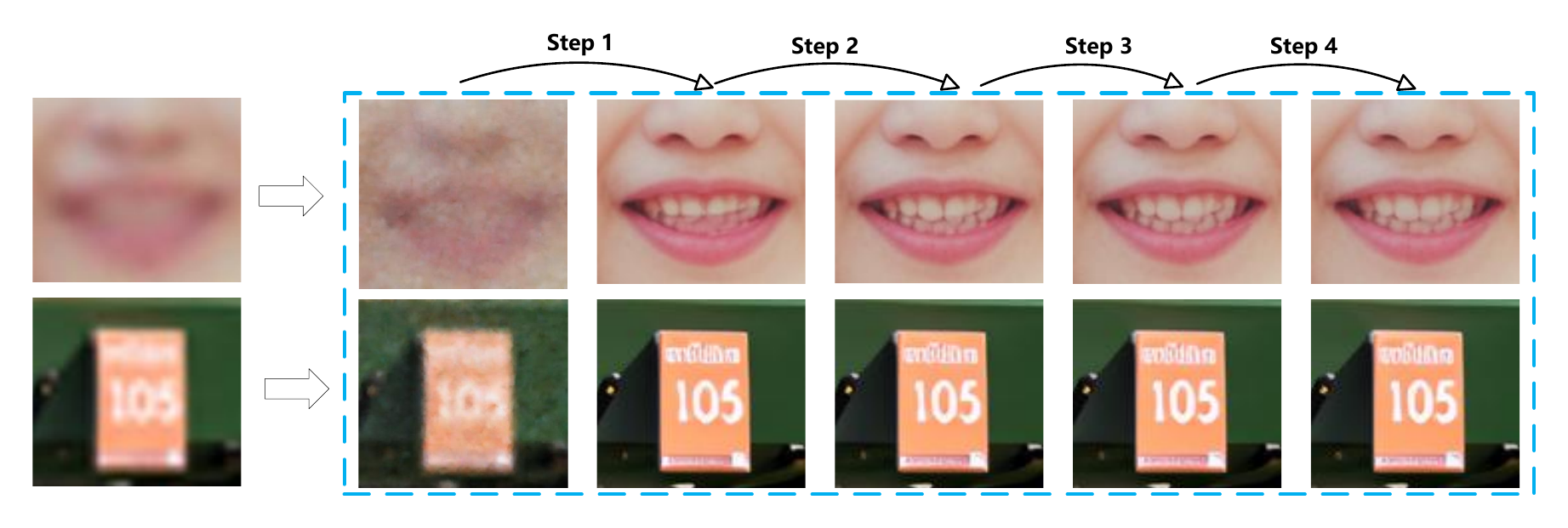}
\caption{Intermediate restoration results from the same multi-step model. As the number of function evaluations (NFE) increases during sampling, ScaleResfusion progressively updates the residual state, leading to clearer structures and more refined image details.}
\Description{Intermediate ScaleResfusion restoration results showing that image details are progressively refined as the number of function evaluations increases.}
\label{fig:nfe_intermediate}
\end{figure*}
\begin{table}[!t]\tiny
\setlength{\abovecaptionskip}{0.1cm}
\caption{Ablation on the number of function evaluations (NFE) with the FLUX2-4B backbone on DRealSR. We compare separately trained 4-step, 2-step, and 1-step models. The best and second-best results of each metric are highlighted in \textcolor{red}{red} and \textcolor{blue}{blue}.}
\label{tab:ablation_nfe}
\centering
\renewcommand{\arraystretch}{1.08}
\setlength{\tabcolsep}{5pt}
\resizebox{\columnwidth}{!}{%
\begin{tabular}{@{}c|c|cccc@{}}
\toprule
\textbf{Dataset} & \textbf{NFE} & \textbf{PSNR $\uparrow$} & \textbf{SSIM $\uparrow$} & \textbf{LPIPS $\downarrow$} & \textbf{FID $\downarrow$} \\ \midrule
 & 4-step & {\color[HTML]{D83931} \textbf{29.77}} & {\color[HTML]{D83931} \textbf{0.82}} & {\color[HTML]{D83931} \textbf{0.25}} & 118.18 \\
 & 2-step & {\color[HTML]{4E83FD} \textbf{29.03}} & {\color[HTML]{4E83FD} \textbf{0.79}} & {\color[HTML]{4E83FD} \textbf{0.27}} & {\color[HTML]{D83931} \textbf{115.43}} \\
\multirow{-3}{*}{DRealSR} & 1-step & 28.93 & 0.78 & 0.28 & {\color[HTML]{4E83FD} \textbf{117.56}} \\ \bottomrule
\end{tabular}%
}
\end{table}

\textbf{Sampling Steps.} Table~\ref{tab:ablation_nfe} compares 1-, 2-, and 4-step models trained separately under the same pipeline. More steps consistently improve fidelity, and 4-step sampling yields the most stable distortion and perceptual metrics (FID varies only marginally across step counts), so we adopt it by default; the same trend holds for the SD3 backbone and on LSDIR-Val (Appendix Table~\ref{tab:ablation_nfe_full}), and the 1- and 2-step variants remain competitive, indicating that the RRF acceleration point provides a strong initialization even under extreme budgets. Moreover, the intermediate predictions of a single multi-step model show that early steps recover the global structure while later steps progressively refine details and correct local inconsistencies (Fig.~\ref{fig:nfe_intermediate} and Appendix Fig.~\ref{fig:ablation_nfe}). Together with the over-smoothing of the deterministic LQ$\rightarrow$HQ flow in Table~\ref{tab:ablation_rrf}, which uses the same 4-step budget, this explains why we retain few-step \emph{stochastic} sampling: reducing the number of steps costs fidelity, while removing the stochastic component causes over-smoothing even at the same sampling budget.

\begin{table}[!t]\tiny
\setlength{\abovecaptionskip}{0.1cm}
\caption{Ablation on LoRA rank with the FLUX2-4B backbone on DRealSR. The best and second-best results of each metric are highlighted in \textcolor{red}{red} and \textcolor{blue}{blue}.}
\label{tab:ablation_lora_rank}
\centering
\renewcommand{\arraystretch}{1.08}
\setlength{\tabcolsep}{6pt}
\resizebox{\columnwidth}{!}{%
\begin{tabular}{@{}c|c|cccc@{}}
\toprule
\textbf{Dataset} & \textbf{Rank} & \textbf{PSNR $\uparrow$} & \textbf{SSIM $\uparrow$} & \textbf{LPIPS $\downarrow$} & \textbf{FID $\downarrow$} \\ \midrule
 & 32 & {\color[HTML]{D83931} \textbf{29.77}} & {\color[HTML]{D83931} \textbf{0.82}} & {\color[HTML]{D83931} \textbf{0.25}} & {\color[HTML]{D83931} \textbf{118.18}} \\
 & 16 & {\color[HTML]{4E83FD} \textbf{29.42}} & {\color[HTML]{4E83FD} \textbf{0.80}} & {\color[HTML]{4E83FD} \textbf{0.26}} & {\color[HTML]{4E83FD} \textbf{123.64}} \\
 & 8 & 29.05 & 0.79 & 0.28 & 131.37 \\
\multirow{-4}{*}{DRealSR} & 4 & 28.63 & 0.77 & 0.30 & 140.82 \\ \bottomrule
\end{tabular}%
}
\end{table}

\textbf{LoRA Rank.} Table~\ref{tab:ablation_lora_rank} ablates the capacity of the trainable residual update. Restoration quality improves monotonically from rank 4 to rank 32 on DRealSR, with the same trend on LSDIR-Val (Appendix Table~\ref{tab:ablation_lora_full}), so we adopt rank 32 by default; notably, even rank 4 already achieves reasonable restoration quality (e.g., 28.63 PSNR on DRealSR). This graceful degradation supports the residual formulation: since the residual vector field differs from the pre-trained target only by the residual offset, the adaptation does not need to relearn the full image--noise transport and can exploit a very compact LoRA subspace.

\begin{table}[!t]\tiny
\setlength{\abovecaptionskip}{0.1cm}
\caption{Ablation of ScaleResfusion on backbone generality on DRealSR. The upper four rows report the \emph{w/o GAN} variants and the shaded rows the \emph{w/ GAN} variants. Within each setting, the best and second-best results of each metric are highlighted in \textcolor{red}{red} and \textcolor{blue}{blue}. The full comparison on all four benchmarks is provided in Appendix Table~\ref{tab:ablation_backbone_full}.}
\label{tab:ablation_backbone}
\centering
\renewcommand{\arraystretch}{1.08}
\setlength{\tabcolsep}{3.8pt}
\resizebox{\columnwidth}{!}{%
\begin{tabular}{@{}c|l|ccccc@{}}
\toprule
\textbf{Dataset} & \textbf{Backbone} & \textbf{PSNR $\uparrow$} & \textbf{SSIM $\uparrow$} & \textbf{LPIPS $\downarrow$} & \textbf{DISTS $\downarrow$} & \textbf{FID $\downarrow$} \\ \midrule
 & SD3 (2B) & 28.77 & 0.79 & 0.30 & 0.23 & 149.67 \\
 & FLUX2-Klein (4B) & {\color[HTML]{4E83FD} \textbf{29.77}} & {\color[HTML]{D83931} \textbf{0.82}} & {\color[HTML]{D83931} \textbf{0.25}} & {\color[HTML]{4E83FD} \textbf{0.20}} & {\color[HTML]{4E83FD} \textbf{118.18}} \\
 & Z-Image (6B) & 29.35 & {\color[HTML]{4E83FD} \textbf{0.80}} & {\color[HTML]{4E83FD} \textbf{0.28}} & 0.22 & 132.31 \\
 & FLUX2-Klein (9B) & {\color[HTML]{D83931} \textbf{30.17}} & {\color[HTML]{D83931} \textbf{0.82}} & {\color[HTML]{D83931} \textbf{0.25}} & {\color[HTML]{D83931} \textbf{0.19}} & {\color[HTML]{D83931} \textbf{109.28}} \\
\rowcolor[gray]{0.95}
 & SD3 (2B) & 27.86 & 0.76 & 0.32 & 0.22 & 146.93 \\
\rowcolor[gray]{0.95}
 & FLUX2-Klein (4B) & 28.26 & {\color[HTML]{4E83FD} \textbf{0.78}} & {\color[HTML]{4E83FD} \textbf{0.29}} & {\color[HTML]{4E83FD} \textbf{0.21}} & {\color[HTML]{4E83FD} \textbf{124.03}} \\
\rowcolor[gray]{0.95}
 & Z-Image (6B) & {\color[HTML]{4E83FD} \textbf{28.34}} & {\color[HTML]{4E83FD} \textbf{0.78}} & {\color[HTML]{D83931} \textbf{0.28}} & {\color[HTML]{4E83FD} \textbf{0.21}} & 127.40 \\
\rowcolor[gray]{0.95}
\multirow{-8}{*}{DRealSR} & FLUX2-Klein (9B) & {\color[HTML]{D83931} \textbf{28.77}} & {\color[HTML]{D83931} \textbf{0.79}} & {\color[HTML]{D83931} \textbf{0.28}} & {\color[HTML]{D83931} \textbf{0.19}} & {\color[HTML]{D83931} \textbf{116.69}} \\ \bottomrule
\end{tabular}%
}
\par\medskip
\includegraphics[width=\columnwidth]{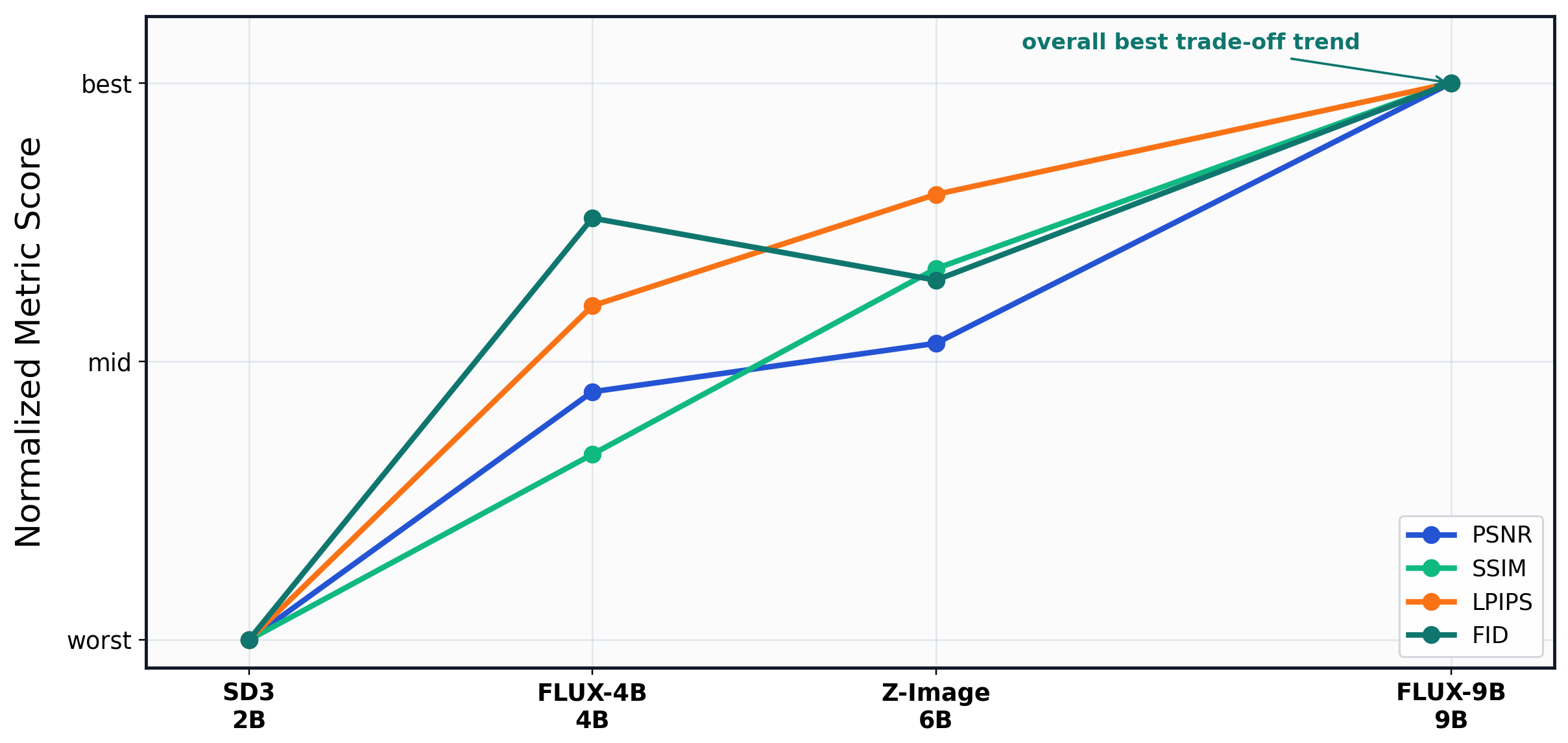}
\captionof{figure}{Consistent transfer of ScaleResfusion across pre-trained rectified-flow backbones (DRealSR \emph{w/ GAN} results from Table~\ref{tab:ablation_backbone}). Despite differences in backbone architecture and pre-training, fidelity and perceptual quality improve from SD3 (2B) to FLUX2 (9B), while FLUX2-4B provides the best quality-efficiency trade-off.}
\label{fig:ablation_scaling}
\end{table}

\textbf{Backbone Generality.} We train ScaleResfusion on pre-trained rectified-flow backbones that differ in both architecture and parameter scale---SD3 (2B), FLUX2 (4B/9B), and Z-Image (6B)---to validate the generality of the residual rectified-flow adaptation rather than to benchmark any single backbone. Table~\ref{tab:ablation_backbone} and Fig.~\ref{fig:ablation_scaling} show consistent transfer across these backbones under both the \emph{w/o GAN} and \emph{w/ GAN} settings, and the same holds on the remaining benchmarks (Appendix Table~\ref{tab:ablation_backbone_full}): on all four benchmarks, fidelity and perceptual similarity improve from SD3 (2B) to FLUX2 (9B), and FLUX2 (9B) achieves the best or tied-best results on every metric, while the ranking among the smaller backbones varies, suggesting that different pre-trained priors emphasize different aspects of restoration quality. Since the backbones differ in architecture, pre-training data, and objectives, we interpret this trend as consistent transfer across heterogeneous pre-trained rectified-flow models rather than as a controlled scaling study. FLUX2 (4B) offers the best practical quality-efficiency trade-off (Table~\ref{tab:efficiency}) and is therefore used as our default backbone. More ablation studies are provided in Appendix~\ref{sec:appendix_ablation}.

\section{Conclusion}
We presented ScaleResfusion, which rewrites residual restoration as a scheduler-independent adaptation interface for pre-trained rectified-flow models. Its core, Residual Rectified Flow, starts sampling from a noisy LQ state at an exact acceleration point with explicit SNR control through the residual ratio $\gamma$, and preserves the linear transport structure of Rectified Flow, so that its residual-vector-field target differs from the pre-trained objective only by the residual offset $\gamma R$ and a frozen billion-scale backbone can be adapted with LoRA-only training and knowledge distillation. Experiments, user studies, and ablations on real-world super-resolution across multiple benchmarks show state-of-the-art restoration quality with only 4 sampling steps, clear subjective preference, and consistent transfer across pre-trained rectified-flow backbones. When the LQ and HQ images share a representation in which the residual $R$ is well-defined and the two endpoints remain close, RRF is in principle applicable to other restoration settings.

\bibliographystyle{IEEEtran}
\bibliography{11_Reference}

\clearpage
\appendices
\clearpage
\makeatletter
\if@twocolumn
  \onecolumn
  \def\AppendixRestoreColumns{\clearpage\twocolumn}
\else
  \def\AppendixRestoreColumns{\clearpage}
\fi
\makeatother
\begingroup
\setlength{\parindent}{0pt}
\setlength{\parskip}{0pt}
\newcommand{\AppendixTOCSection}[2]{%
  \noindent\makebox[1.6em][l]{\textbf{\textcolor{blue}{\ref{#1}}}}%
  \hyperref[#1]{\textbf{\textcolor{blue}{#2}}}%
  \leaders\hbox{\kern .35em.\raisebox{.35ex}{.}\kern .35em}\hfill%
  \textbf{\pageref{#1}}\par\vspace{0.78em}}
\newcommand{\AppendixTOCSubsection}[2]{%
  \noindent\hspace{1.6em}\makebox[2.6em][l]{\textcolor{blue}{\ref{#1}}}%
  \hyperref[#1]{\textcolor{blue}{#2}}%
  \leaders\hbox{\kern .35em.\raisebox{.35ex}{.}\kern .35em}\hfill%
  \pageref{#1}\par\vspace{0.42em}}
\null\vskip 0.04\textheight
\begin{center}
{\Large\bfseries ScaleResfusion: Residual Rectified Flow based on Residual Vector Field\par}
\vskip 0.45em
{\large\itshape Supplementary Materials\par}
\end{center}
\vskip 2.1em
{\Large\bfseries Contents\par}
\vskip 1.25em
\small
\AppendixTOCSection{sec:residual_transport_theoretical_justification}{Residual Initialization Leads to a Better Transport Path}
\AppendixTOCSection{sec:appendix_experimental_setting_details}{Experimental Setting Details}
\AppendixTOCSection{sec:appendix_no_reference_metrics}{Caution on No-Reference Metrics}
\AppendixTOCSection{sec:appendix_ablation}{Extra Ablation Study}
\AppendixTOCSubsection{sec:ablation_on_rsr}{Ablation on Residual Ratio}
\AppendixTOCSubsection{sec:ablation_on_dape}{Ablation on DAPE}
\AppendixTOCSubsection{sec:ablation_on_feature_extractor}{Ablation on Feature Extractor}
\AppendixTOCSection{sec:general_residual_flow_derivation}{Theoretical Justification of Residual Rectified Flow}
\AppendixTOCSection{sec:appendix_more_visual_results}{More Visual Results}
\vfill
\hrule
\vskip 0.18em
\hrule
\endgroup
\AppendixRestoreColumns
\section{Residual Initialization Leads to a Better Transport Path}
\label{sec:residual_transport_theoretical_justification}

In this section, we provide a theoretical justification for residual initialization from the perspective of measure differential equations. Our key goal is to show that, for a fixed observed LQ image and a fixed learned residual velocity field, the reverse process is stable with respect to its initial measure. Consequently, an initialization that is closer to a task-aware reference admits a tighter upper bound on terminal discrepancy. This conditional comparison does not assert a strict ordering of actual terminal distributions. We further show that the residual parameterization provides a task-aware structured initialization while retaining conditional randomness through noise.

Throughout this section, we fix the observed LQ image $\hat{x}_0$. All velocity fields, flow maps, and probability measures below are understood conditionally on this observation; to keep notation compact, that conditioning is suppressed except where displayed.

\paragraph{\textbf{From sample trajectories to measure flows.}}
After the acceleration point $t^\star=\frac{1}{1+\gamma}$, the reverse process of Residual Rectified Flow is governed by the ODE
\begin{equation}
\label{eq:theory_sample_ode_en}
\frac{dX_t}{dt}=v_\theta(X_t,t), \qquad t\in[0,t^\star],
\end{equation}
where $v_\theta$ denotes the learned residual velocity field and $X_{t^\star}$ is a random initial state. Let
\[
\mu_t := \mathrm{Law}(X_t)
\]
be the probability distribution of $X_t$ at time $t$. The measures considered below have finite second moments. Assume also that $v_\theta$ is continuous in time and uniformly globally Lipschitz in the state variable: there exists a constant $L\ge 0$ such that
\begin{equation}
\label{eq:theory_lipschitz_en}
\|v_\theta(x,t)-v_\theta(y,t)\|\le L\|x-y\|,\qquad \forall x,y,\ \forall t\in[0,t^\star].
\end{equation}
These conditions give a unique global ODE flow on $[0,t^\star]$ and preserve finite second moments. Consequently, the family of measures $\{\mu_t\}_{t\in[0,t^\star]}$ satisfies the continuity equation
\begin{equation}
\label{eq:theory_continuity_en}
\partial_t \mu_t + \nabla \cdot \big(v_\theta(\cdot,t)\mu_t\big)=0.
\end{equation}
Let $\Phi_{t,s}$ denote the flow map induced by Eq.~\eqref{eq:theory_sample_ode_en}, namely, for $0\le t\le s\le t^\star$,
\[
X_t=\Phi_{t,s}(X_s).
\]
Then the corresponding measure solution can be written in pushforward form as
\begin{equation}
\label{eq:theory_pushforward_en}
\mu_t = (\Phi_{t,s})_\# \mu_s.
\end{equation}
Therefore, the reverse generation process can be interpreted as a measure transport operator
\[
\mathcal{T}_{t\leftarrow s}(\mu_s):=(\Phi_{t,s})_\#\mu_s,
\]
which transports an initial measure at time $s$ to a terminal measure at time $t$ along the learned residual flow.

\paragraph{\textbf{Stability of the reverse measure flow.}}
To quantify the effect of initialization, consider two finite-second-moment initial measures $\mu_s$ and $\nu_s$, and let
\[
\mu_t=(\Phi_{t,s})_\#\mu_s,\qquad \nu_t=(\Phi_{t,s})_\#\nu_s
\]
be the corresponding evolved measures under the same fixed velocity field $v_\theta$. Let $X_t$ and $Y_t$ be two characteristic trajectories. For the reverse direction, set
\[
\begin{aligned}
\tau&=s-t,\qquad Z_\tau=X_{s-\tau},\\
\widetilde Z_\tau&=Y_{s-\tau},\\
w(z,\tau)&=-v_\theta(z,s-\tau).
\end{aligned}
\]
Thus $\tau\ge 0$ and
\[
\frac{dZ_\tau}{d\tau}=w(Z_\tau,\tau),\qquad
\frac{d\widetilde Z_\tau}{d\tau}=w(\widetilde Z_\tau,\tau).
\]
The reverse velocity $w$ has the same Lipschitz constant as $v_\theta$, so
\[
\|Z_\tau-\widetilde Z_\tau\|
\le \|Z_0-\widetilde Z_0\|+L\int_0^\tau\|Z_r-\widetilde Z_r\|\,dr.
\]
Applying Gr\"onwall's inequality at $\tau=s-t$ gives
\begin{equation}
\label{eq:theory_sample_stability_en}
\|X_t-Y_t\|\le e^{L(s-t)}\|X_s-Y_s\|.
\end{equation}
Evolving any coupling of $\mu_s$ and $\nu_s$ by the same flow and taking the infimum over couplings yields
\begin{equation}
\label{eq:theory_measure_stability_en}
W_2(\mu_t,\nu_t)\le e^{L(s-t)}W_2(\mu_s,\nu_s).
\end{equation}
Equation~\eqref{eq:theory_measure_stability_en} shows that the reverse flow depends continuously on its initial measure. Since $L\ge 0$ and $s-t\ge 0$, its factor is at least one (equal to one when $L=0$); this stability estimate does not in general establish non-expansiveness or contraction.

\paragraph{\textbf{Conditional comparison of initializations.}}
The stability result controls terminal discrepancy for a fixed observed LQ image and a fixed velocity field. Let $\mu_{t^\star}^{\dagger}$ denote an ideal task-aware reference measure, $\mu_{t^\star}^{G}$ a task-agnostic Gaussian initialization, and $\mu_{t^\star}^{R}$ our residual-based initialization. Suppose, as an explicit task-dependent assumption, that
\begin{equation}
\label{eq:theory_initial_discrepancy_en}
W_2\!\left(\mu_{t^\star}^{R},\mu_{t^\star}^{\dagger}\right)
<
W_2\!\left(\mu_{t^\star}^{G},\mu_{t^\star}^{\dagger}\right),
\end{equation}
where both initializations are compared under the same fixed velocity field. Define their terminal measures, including the reference terminal measure, by $\mu_0^j=(\Phi_{0,t^\star})_\#\mu_{t^\star}^j$ for $j\in\{R,G,\dagger\}$. With $C_L=\exp(Lt^\star)$, Eq.~\eqref{eq:theory_measure_stability_en} gives
\begin{equation}
\label{eq:theory_terminal_discrepancy_en}
W_2\!\left(\mu_{0}^{R},\mu_{0}^{\dagger}\right)
\le
C_L
W_2\!\left(\mu_{t^\star}^{R},\mu_{t^\star}^{\dagger}\right)
<
C_L
W_2\!\left(\mu_{t^\star}^{G},\mu_{t^\star}^{\dagger}\right),
\end{equation}
The strict inequality is an assumption about the initial measures, not a consequence of stability. Equation~\eqref{eq:theory_terminal_discrepancy_en} compares propagated upper bounds and does not assert a strict ordering of the actual terminal discrepancies. To relate the reference terminal measure to the actual HQ target, let $\pi_0:=\pi_0(\cdot\mid\hat{x}_0)$ and define the common model error $E_\theta:=W_2(\mu_0^\dagger,\pi_0)$. If the ideal reference is transported to the target, then $E_\theta=0$; otherwise, the triangle inequality and Eq.~\eqref{eq:theory_measure_stability_en} give
\begin{equation}
\label{eq:theory_target_discrepancy_en}
W_2\!\left(\mu_0^j,\pi_0\right)
\le
C_L W_2\!\left(\mu_{t^\star}^j,\mu_{t^\star}^\dagger\right)+E_\theta,
\qquad j\in\{R,G\}.
\end{equation}
The term $E_\theta$ is common to both initializations. Stability alone neither proves the initial closeness assumption in Eq.~\eqref{eq:theory_initial_discrepancy_en} nor implies exact target matching by the learned ODE. Thus the comparison supplies a conditional, task-aligned motivation for an observation-aware residual initialization, rather than an unconditional ordering of separately trained models.

\paragraph{\textbf{Residual as a task-aware parameterization.}}
We now explain why the residual $R$ is a natural structured variable for restoration. Let $\hat{x}_0$ and $x_0$ denote the LQ and HQ images, respectively, and define
\begin{equation}
\label{eq:theory_residual_def_en}
R=\hat{x}_0-x_0.
\end{equation}
Then
\begin{equation}
\label{eq:theory_residual_inverse_en}
x_0=\hat{x}_0-R.
\end{equation}
For each fixed observed image $\hat{x}_0$, the target image $x_0$ and the residual variable $R$ are in one-to-one correspondence. Recovering the HQ image can therefore be parameterized as recovering a correction term rather than regenerating the full image from scratch.

Moreover, the target velocity in Residual Rectified Flow is
\begin{equation}
\label{eq:theory_residual_velocity_en}
u_t = \gamma R + \epsilon - x_0.
\end{equation}
Substituting Eq.~\eqref{eq:theory_residual_inverse_en} into Eq.~\eqref{eq:theory_residual_velocity_en} yields
\begin{equation}
\label{eq:theory_residual_dominant_en}
u_t = (\gamma+1)R + \epsilon - \hat{x}_0.
\end{equation}
Equation~\eqref{eq:theory_residual_dominant_en} expresses the task-specific transport direction using the observed image, residual, and noise. The degraded image $\hat{x}_0$ provides coarse structure and semantic layout, while $\epsilon$ supplies stochasticity and $R$ encodes the correction toward the clean target. This makes the residual a direct, task-aligned parameterization for constructing an observation-aware initialization.

\paragraph{\textbf{Deterministic residual-only initialization.}}
The true residual $R=\hat{x}_0-x_0$ remains random conditioned on $\hat{x}_0$ when the target conditional law is non-degenerate. To study a deterministic residual-only initialization, let $\hat{R}(\hat{x}_0)$ be a deterministic predicted residual depending only on the observation, distinct from the true residual $R$, and let $g$ be deterministic. Define
\[
\begin{aligned}
h(\hat{x}_0)&:=g\!\left(\hat{x}_0,\hat{R}(\hat{x}_0)\right),\\
X_{t^\star}^{R\text{-only}}&=h(\hat{x}_0).
\end{aligned}
\]
Conditioned on $\hat{x}_0$, the induced initial measure is therefore a Dirac mass:
\[
\mu_{t^\star}^{R\text{-only}}(\cdot \mid \hat{x}_0)=\delta_{h(\hat{x}_0)}.
\]
Since the reverse dynamics is also a deterministic ODE flow, its pushforward remains a Dirac measure:
\[
\begin{aligned}
\mu_{0}^{R\text{-only}}(\cdot \mid \hat{x}_0)
&=(\Phi_{0,t^\star})_\# \delta_{h(\hat{x}_0)}\\
&=\delta_{\Phi_{0,t^\star}(h(\hat{x}_0))}.
\end{aligned}
\]
Thus this deterministic initialization together with a deterministic ODE cannot generate conditional variability by itself.

This property becomes problematic when the target conditional HQ distribution is non-degenerate, i.e.,
\[
\pi_0(\cdot \mid \hat{x}_0)
\]
has finite covariance $\Sigma_\pi\neq 0$. If its mean is $m_\pi$, then any deterministic terminal point $x$ satisfies
\[
W_2^2\!\left(\delta_x,\pi_0(\cdot \mid \hat{x}_0)\right)
=
\|x-m_\pi\|^2+\operatorname{Tr}(\Sigma_\pi),
\]
so the mismatch is bounded below by $\operatorname{Tr}(\Sigma_\pi)>0$. The same limitation on conditional variability applies to the deterministic LQ$\rightarrow$HQ flow in Table~\ref{tab:ablation_rrf}.

In contrast, the proposed residual-based initialization preserves task structure while retaining stochastic flexibility. For any finite $\gamma>0$ and noise $\epsilon\sim\mathcal N(0,I)$ independent of $\hat{x}_0$, it is
\[
X_{t^\star}
=
\frac{\gamma}{1+\gamma}\hat{x}_0
+
\frac{1}{1+\gamma}\epsilon.
\]
Its conditional measure, which is the same $\mu_{t^\star}^{R}$ used in the comparison above, is therefore non-degenerate:
\[
\mu_{t^\star}^{R}(\cdot \mid \hat{x}_0)
=
\mathcal N\left(
\frac{\gamma}{1+\gamma}\hat{x}_0,\,
\frac{1}{(1+\gamma)^2}I
\right),
\]
which is computable from the LQ image and sampled noise and does not require ground-truth $R$. This non-degeneracy retains conditional randomness, but by itself does not prove a better terminal distribution or coverage; the tighter-bound statement remains conditional on Eq.~\eqref{eq:theory_initial_discrepancy_en} and the same fixed velocity field.

\paragraph{\textbf{Implications for ScaleResfusion.}}
The residual parameterization also motivates low-rank parameter-efficient adaptation as a modeling choice. In this restoration setting, $\hat{x}_0$ carries coarse content while $R$ encodes restoration-specific corrections such as missing high-frequency details, denoising compensation, and deblurring offsets. This suggests concentrating adaptation on residual-related behavior can be practical. However, image-domain residual complexity does not imply a low-rank parameter update. Rank suitability is assessed empirically in Table~\ref{tab:ablation_lora_rank}; the analysis here motivates testing LoRA as a modeling choice.

\begin{figure*}[t]
\centering
\includegraphics[width=\textwidth]{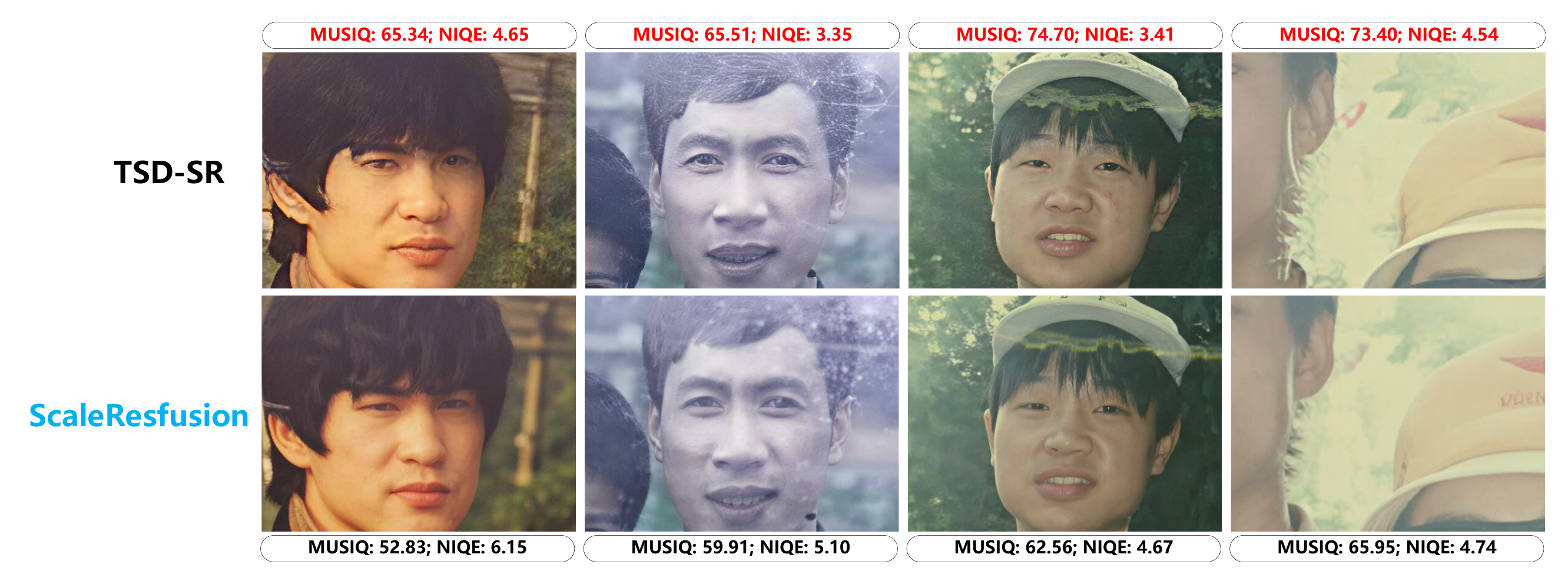}
\caption{No-reference metric comparison on examples without ground truth. Although TSD-SR still obtains better MUSIQ/NIQE scores than ScaleResfusion, its visual quality is clearly worse, with stronger semantic drift and local artifacts.}
\Description{Visual comparison between TSD-SR and ScaleResfusion on examples without ground truth, showing that better no-reference scores can still correspond to worse visual quality.}
\label{fig:exp_tradeoff_nogt}
\end{figure*}
In summary, the theoretical justification for residual initialization has three parts. First, the reverse process is a measure differential equation driven by the learned residual velocity field, and its well-posed reverse flow is stable under the stated finite-moment and regularity assumptions. Second, for the same fixed velocity field, an explicitly assumed closer residual initialization yields a tighter upper bound on propagated reference discrepancy, with the common target error captured by $E_\theta$. Third, conditioned on the LQ observation, $R$ is in bijection with the target HQ image and gives an explicit task-aware correction parameterization. A deterministic predicted-residual initialization with a deterministic ODE remains a Dirac law, whereas the proposed LQ-plus-noise initialization retains conditional randomness. Together, these observations motivate an observation-aware initialization that retains stochasticity.

\section{Experimental Setting Details}
\label{sec:appendix_experimental_setting_details}
We train all models with PyTorch and the Diffusers framework, using the AdamW optimizer~\cite{loshchilov2017decoupled} with default hyperparameters. For ScaleResfusion with SD3 (2B) or Z-Image (6B) as the backbone, we use a ReferenceNet as the feature extractor and inject LQ features via Ref Attention. With FLUX2-Klein (4B or 9B) as the backbone, we instead concatenate the LQ features with the input features along the feature dimension, preserving FLUX2’s native feature extraction capabilities for image editing. We train the SD3- and FLUX2-Klein-4B-based models on 8 NVIDIA GeForce RTX 5090 GPUs, and the Z-Image- and FLUX2-Klein-9B-based models on 8 NVIDIA RTX Pro 6000 GPUs.

We train the \emph{w/o GAN} variant for 100K iterations with a learning rate of $5 \times 10^{-5}$, and then fine-tune it with a learning rate of $1 \times 10^{-5}$ to obtain \emph{w/ GAN}. We adopt the Dynamic Distribution Guidance (DynaDG) and Dynamic Renoise Sampling (DynaRS) following~\cite{jiang2025distribution} for cold start during the first 10K iterations. By default, we use 4-step sampling and the residual ratio $\gamma=1$.
All testing and inference-time measurements are conducted on a single NVIDIA RTX A6000 GPU. For inference time evaluation, we test all methods at $512\times512$ resolution by processing 3,000 images and computing the average inference time per image after subtracting the model loading time from the total time. We employ RAM-based~\cite{zhang2024recognize} DAPE following OSEDiff~\cite{wu2024one}. We use a DINOv2-based~\cite{oquab2023dinov2} GAN discriminator identical to AddSR~\cite{xie2024addsr}.

During training, we also observe a training--inference mismatch similar to the multi-step generator issue discussed in DMD2~\cite{yin2024improved}. In the RRF training process, the $x_0$ component in the observed state $x_t$ comes from the real HQ image. During inference, this component must instead be supplied by the generated prediction $x_0^{pred}$. With very few (1--4) inference steps, the $x_0$ component changes rapidly between adjacent sampling states, which further amplifies the input-distribution mismatch. This mismatch is presumably less harmful for relatively deterministic restoration tasks such as low-light enhancement and deraining. However, it becomes critical for highly ill-posed problems such as super-resolution, where the generated $x_0^{pred}$ can deviate substantially from the real $x_0$, leading to over-saturated and hallucinated textures. Following DMD2, we replace noisy real training states with noisy synthetic states produced by the current generator after several sampling steps.

\begin{figure}[!t]
\centering
\includegraphics[width=\columnwidth]{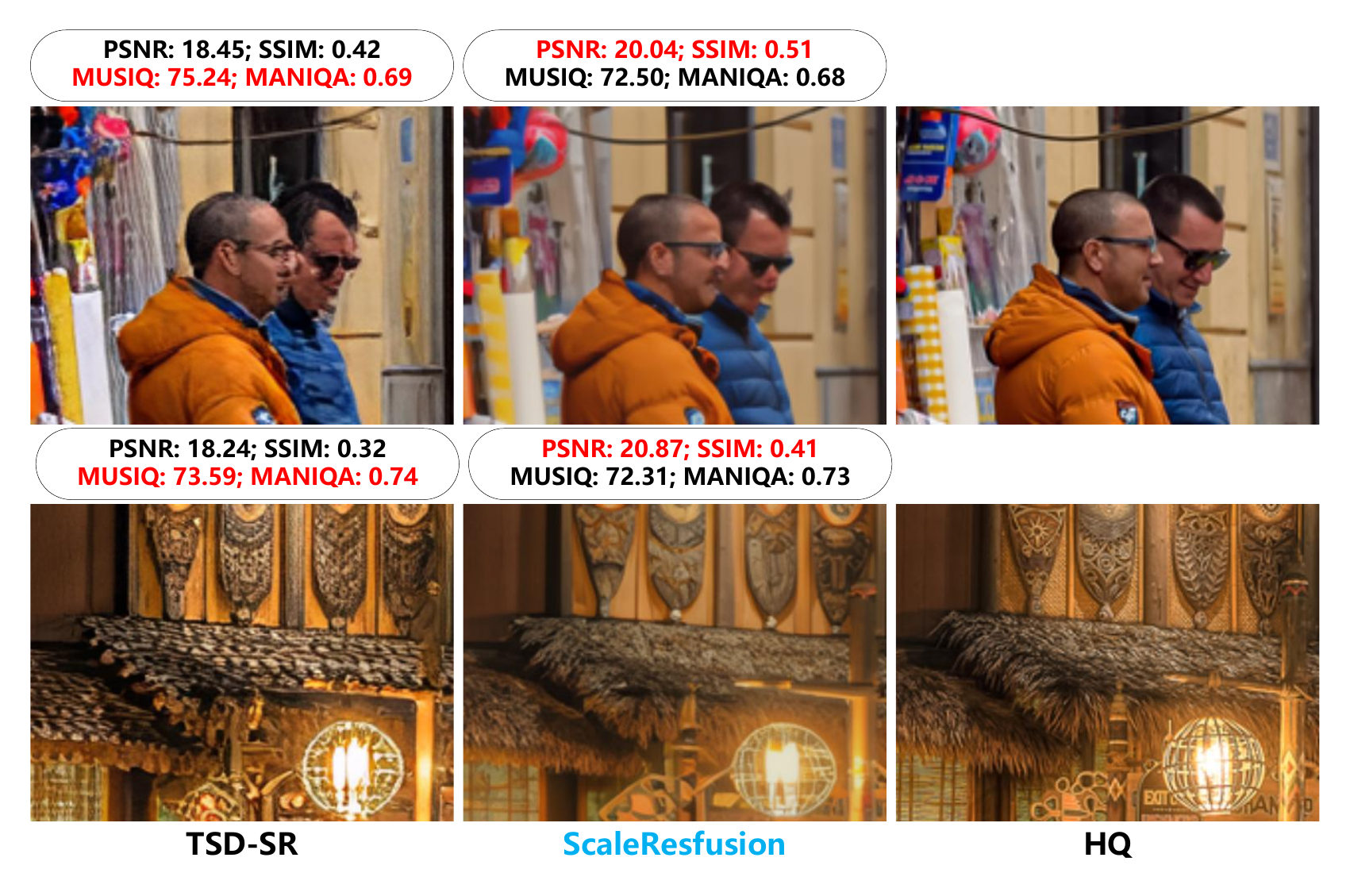}
\caption{Examples where no-reference metrics do not fully reflect visual quality. TSD-SR obtains higher MUSIQ/MANIQA scores, but introduces stronger artifacts and less faithful structures than ScaleResfusion.}
\Description{Visual comparison between TSD-SR, ScaleResfusion, and ground truth, showing that higher no-reference scores can correspond to worse visual fidelity.}
\label{fig:exp_tradeoff}
\end{figure}
\section{Caution on No-Reference Metrics}
\label{sec:appendix_no_reference_metrics}
No-reference metrics are useful when ground-truth images are unavailable, but current metrics still contain biases toward certain low-level statistics, such as sharpness, contrast, and texture richness. As a result, they may assign higher scores to images with over-sharpened details, semantic drift, or local artifacts, even when these artifacts reduce restoration fidelity. Fig.~\ref{fig:exp_tradeoff_nogt} considers in-the-wild examples without ground-truth images, where reference-based metrics such as PSNR/SSIM cannot be computed and evaluation naturally relies more on no-reference scores. Although TSD-SR reports better no-reference scores in these cases, the visual comparison reveals facial semantic drift, unstable exposure and color appearance, and structural degradation, while ScaleResfusion better preserves plausible structures and overall appearance---suggesting that the bias of no-reference metrics becomes more visible when they are used in isolation.

Fig.~\ref{fig:exp_tradeoff} further shows paired examples with ground truth, where TSD-SR obtains higher MUSIQ/MANIQA scores than ScaleResfusion, yet its outputs have lower reference fidelity and visibly stronger artifacts, confirming that no-reference scores can be inconsistent with both ground-truth-based metrics and human visual judgment. Together, these observations motivate a more balanced evaluation protocol: when the gap in no-reference metrics is not substantial, consistency-oriented metrics (PSNR, SSIM, LPIPS, DISTS) and distribution-level metrics (FID) should receive more weight whenever they are available, since they better capture whether the restored image remains faithful to the input content and aligned with the target image distribution. We thus treat no-reference scores as complementary indicators rather than the sole criterion for evaluating Real-IR quality.

\section{Extra Ablation Study}
\label{sec:appendix_ablation}
\begin{figure}[t]
\centering
\includegraphics[width=\columnwidth]{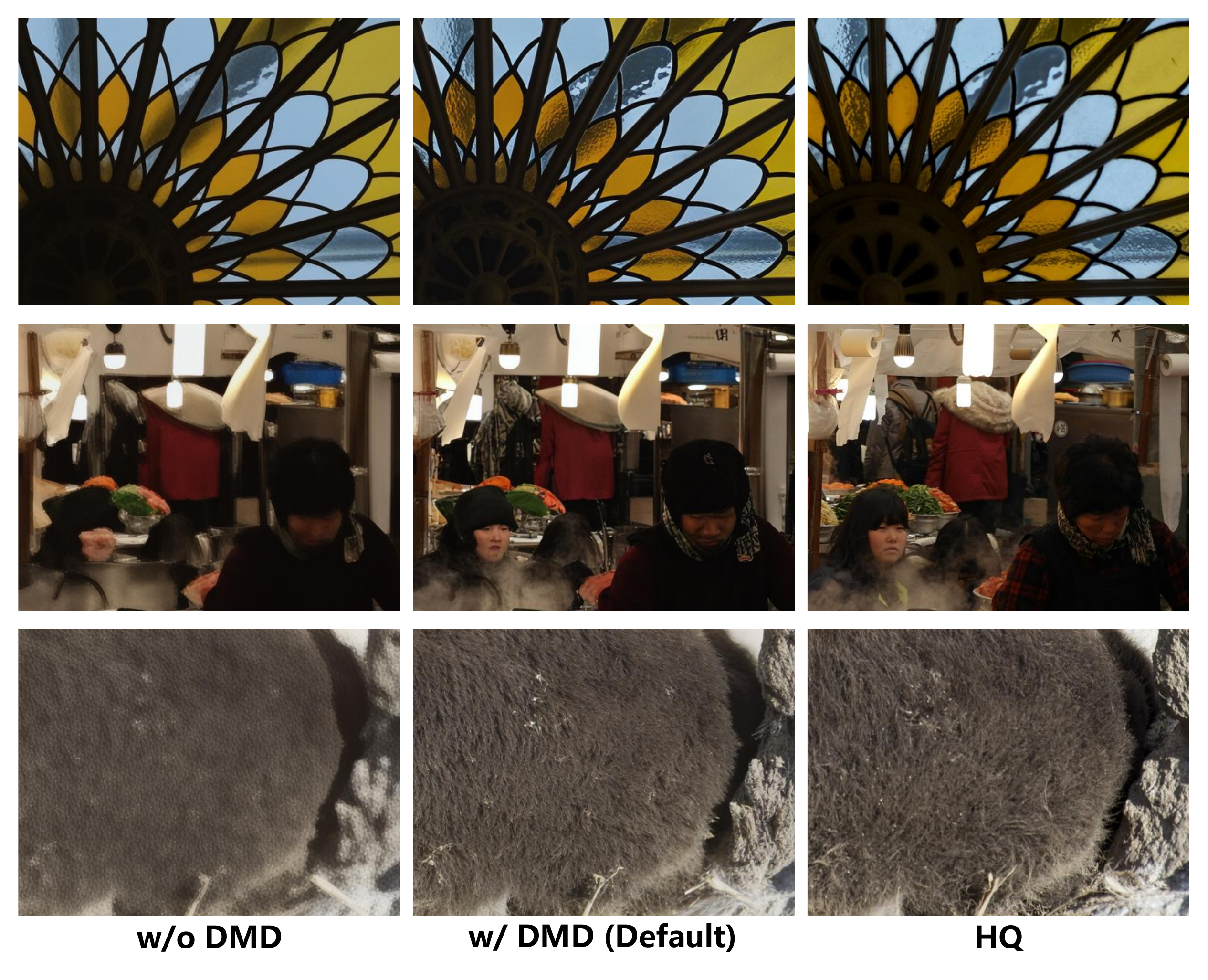}
\caption{Visual ablation on the training regularizer. Adding a distribution-level regularizer (here DMD) suppresses unstable textures and improves both structural fidelity and perceptual detail quality.}
\Description{Visual comparison of restoration results with and without the DMD regularizer.}
\label{fig:ablation_vsd}
\end{figure}

\begin{figure}[t]
\centering
\includegraphics[width=\columnwidth]{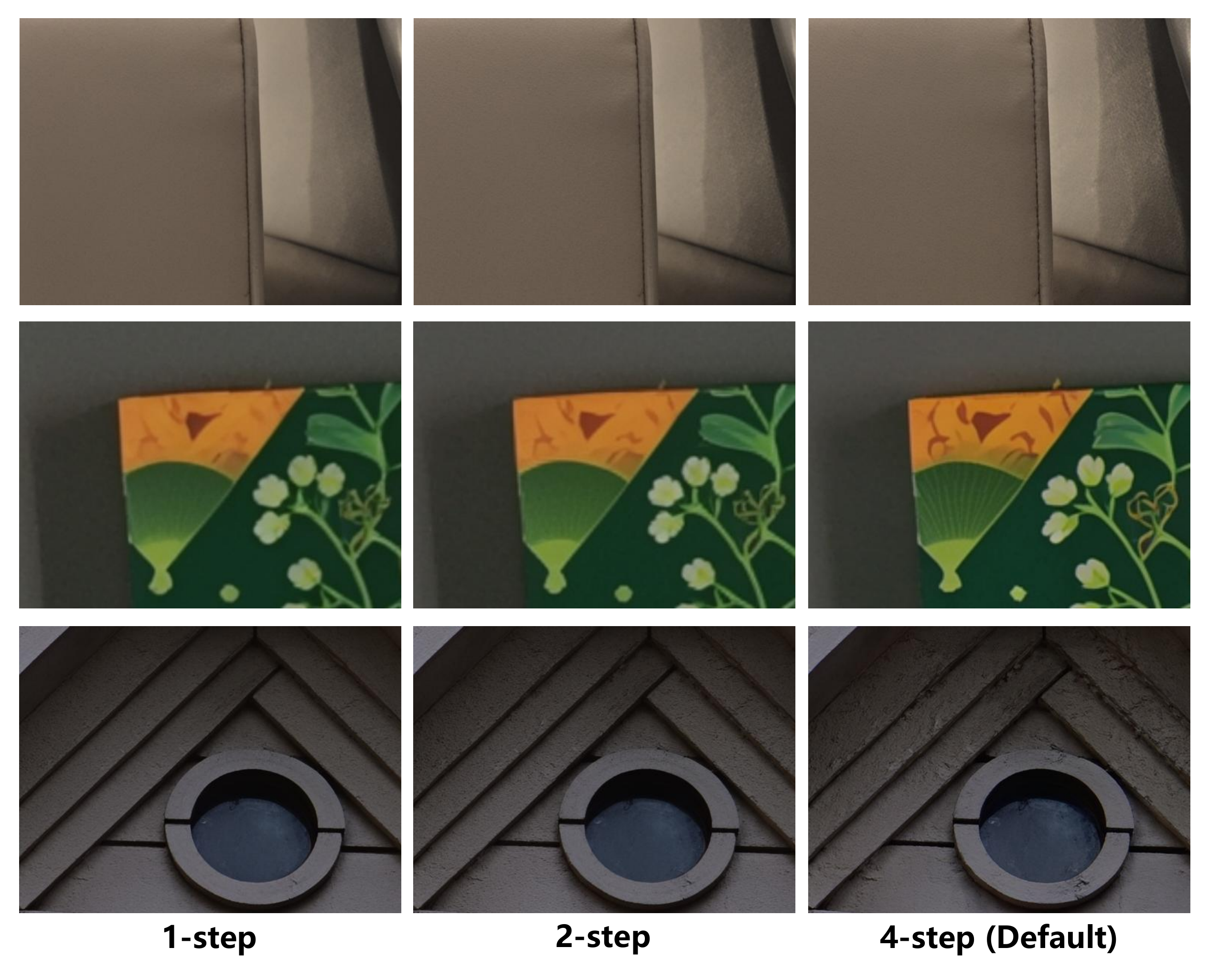}
\caption{Visual ablation on the number of function evaluations. Increasing the number of sampling steps generally improves restoration fidelity and detail consistency, while the proposed residual transport remains effective even with very few steps.}
\Description{Visual comparison of ScaleResfusion restoration results under different numbers of function evaluations.}
\label{fig:ablation_nfe}
\end{figure}

In this section, we provide additional ablation studies on the residual ratio, DAPE, and feature extraction. We follow the same evaluation protocol as the main experiments and use the FLUX2-4B variant as the default setting unless otherwise specified. Figs.~\ref{fig:ablation_vsd} and~\ref{fig:ablation_nfe} provide visual comparisons for the main-text ablations on the training regularizer and sampling steps. Table~\ref{tab:ablation_nfe_full} reports the full NFE comparison including the SD3 backbone and LSDIR-Val, Table~\ref{tab:ablation_lora_full} the full LoRA-rank comparison, and Table~\ref{tab:ablation_backbone_full} the full backbone-generality comparison on all four benchmarks.
\begin{table}[t]\tiny
\centering
\caption{Full ablation on the number of function evaluations (NFE), including both SD3 and FLUX2-4B backbones on DRealSR and LSDIR-Val.}
\label{tab:ablation_nfe_full}
\setlength{\tabcolsep}{3.5pt}
\renewcommand{\arraystretch}{1.05}
\resizebox{\linewidth}{!}{%
\begin{tabular}{@{}l|l|c|cccc@{}}
\toprule
\textbf{Dataset} & \textbf{Backbone} & \textbf{NFE} & \textbf{PSNR $\uparrow$} & \textbf{SSIM $\uparrow$} & \textbf{LPIPS $\downarrow$} & \textbf{FID $\downarrow$} \\
\midrule
 & & 4-step & 28.77 & 0.79 & 0.30 & 149.67 \\
 & & 2-step & 28.61 & 0.78 & 0.31 & 150.90 \\
 \multirow{-3}{*}{DRealSR} & \multirow{-3}{*}{SD3} & 1-step & 28.23 & 0.77 & 0.31 & 153.65 \\
\midrule
 & & 4-step & 29.77 & 0.82 & 0.25 & 118.18 \\
 & & 2-step & 29.03 & 0.79 & 0.27 & 115.43 \\
 \multirow{-3}{*}{DRealSR} & \multirow{-3}{*}{FLUX2-4B} & 1-step & 28.93 & 0.78 & 0.28 & 117.56 \\
\midrule
 & & 4-step & 20.22 & 0.52 & 0.25 & 63.84 \\
 & & 2-step & 19.56 & 0.49 & 0.26 & 65.07 \\
 \multirow{-3}{*}{LSDIR-Val} & \multirow{-3}{*}{SD3} & 1-step & 19.18 & 0.48 & 0.26 & 67.82 \\
\midrule
 & & 4-step & 21.35 & 0.57 & 0.23 & 41.69 \\
 & & 2-step & 20.61 & 0.54 & 0.25 & 38.94 \\
 \multirow{-3}{*}{LSDIR-Val} & \multirow{-3}{*}{FLUX2-4B} & 1-step & 20.51 & 0.53 & 0.26 & 41.07 \\
\bottomrule
\end{tabular}%
}
\end{table}
\begin{table}[t]\tiny
\centering
\caption{Full ablation on LoRA rank with the FLUX2-4B backbone on DRealSR and LSDIR-Val.}
\label{tab:ablation_lora_full}
\setlength{\tabcolsep}{4.2pt}
\renewcommand{\arraystretch}{1.05}
\resizebox{\linewidth}{!}{%
\begin{tabular}{@{}l|c|cccc@{}}
\toprule
\textbf{Dataset} & \textbf{Rank} & \textbf{PSNR $\uparrow$} & \textbf{SSIM $\uparrow$} & \textbf{LPIPS $\downarrow$} & \textbf{FID $\downarrow$} \\
\midrule
 & 32 & 29.77 & 0.82 & 0.25 & 118.18 \\
 & 16 & 29.42 & 0.80 & 0.26 & 123.64 \\
 & 8 & 29.05 & 0.79 & 0.28 & 131.37 \\
 \multirow{-4}{*}{DRealSR} & 4 & 28.63 & 0.77 & 0.30 & 140.82 \\
\midrule
 & 32 & 21.35 & 0.57 & 0.23 & 41.69 \\
 & 16 & 21.00 & 0.55 & 0.24 & 47.15 \\
 & 8 & 20.63 & 0.54 & 0.26 & 54.88 \\
 \multirow{-4}{*}{LSDIR-Val} & 4 & 20.21 & 0.52 & 0.28 & 64.33 \\
\bottomrule
\end{tabular}%
}
\end{table}
\begin{table}[t]\tiny
\centering
\caption{Full backbone-generality comparison of ScaleResfusion across four benchmarks. For each benchmark, the upper four rows report the \emph{w/o GAN} variants and the shaded rows the \emph{w/ GAN} variants. Within each benchmark and setting, the best and second-best results of each metric are highlighted in \textcolor{red}{red} and \textcolor{blue}{blue}.}
\label{tab:ablation_backbone_full}
\renewcommand{\arraystretch}{1.08}
\setlength{\tabcolsep}{3.8pt}
\resizebox{\columnwidth}{!}{%
\begin{tabular}{@{}c|l|ccccc@{}}
\toprule
\textbf{Dataset} & \textbf{Backbone} & \textbf{PSNR $\uparrow$} & \textbf{SSIM $\uparrow$} & \textbf{LPIPS $\downarrow$} & \textbf{DISTS $\downarrow$} & \textbf{FID $\downarrow$} \\ \midrule
 & SD3 (2B) & 28.77 & 0.79 & 0.30 & 0.23 & 149.67 \\
 & FLUX2-Klein (4B) & {\color[HTML]{4E83FD} \textbf{29.77}} & {\color[HTML]{D83931} \textbf{0.82}} & {\color[HTML]{D83931} \textbf{0.25}} & {\color[HTML]{4E83FD} \textbf{0.20}} & {\color[HTML]{4E83FD} \textbf{118.18}} \\
 & Z-Image (6B) & 29.35 & {\color[HTML]{4E83FD} \textbf{0.80}} & {\color[HTML]{4E83FD} \textbf{0.28}} & 0.22 & 132.31 \\
 & FLUX2-Klein (9B) & {\color[HTML]{D83931} \textbf{30.17}} & {\color[HTML]{D83931} \textbf{0.82}} & {\color[HTML]{D83931} \textbf{0.25}} & {\color[HTML]{D83931} \textbf{0.19}} & {\color[HTML]{D83931} \textbf{109.28}} \\
\rowcolor[gray]{0.95}
 & SD3 (2B) & 27.86 & 0.76 & 0.32 & 0.22 & 146.93 \\
\rowcolor[gray]{0.95}
 & FLUX2-Klein (4B) & 28.26 & {\color[HTML]{4E83FD} \textbf{0.78}} & {\color[HTML]{4E83FD} \textbf{0.29}} & {\color[HTML]{4E83FD} \textbf{0.21}} & {\color[HTML]{4E83FD} \textbf{124.03}} \\
\rowcolor[gray]{0.95}
 & Z-Image (6B) & {\color[HTML]{4E83FD} \textbf{28.34}} & {\color[HTML]{4E83FD} \textbf{0.78}} & {\color[HTML]{D83931} \textbf{0.28}} & {\color[HTML]{4E83FD} \textbf{0.21}} & 127.40 \\
\rowcolor[gray]{0.95}
\multirow{-8}{*}{DRealSR} & FLUX2-Klein (9B) & {\color[HTML]{D83931} \textbf{28.77}} & {\color[HTML]{D83931} \textbf{0.79}} & {\color[HTML]{D83931} \textbf{0.28}} & {\color[HTML]{D83931} \textbf{0.19}} & {\color[HTML]{D83931} \textbf{116.69}} \\ \midrule
 & SD3 (2B) & 25.67 & 0.73 & 0.29 & 0.22 & 126.75 \\
 & FLUX2-Klein (4B) & {\color[HTML]{4E83FD} \textbf{27.05}} & {\color[HTML]{D83931} \textbf{0.78}} & {\color[HTML]{4E83FD} \textbf{0.24}} & {\color[HTML]{4E83FD} \textbf{0.20}} & {\color[HTML]{4E83FD} \textbf{104.74}} \\
 & Z-Image (6B) & 26.29 & {\color[HTML]{4E83FD} \textbf{0.75}} & 0.27 & 0.21 & 114.35 \\
 & FLUX2-Klein (9B) & {\color[HTML]{D83931} \textbf{27.30}} & {\color[HTML]{D83931} \textbf{0.78}} & {\color[HTML]{D83931} \textbf{0.23}} & {\color[HTML]{D83931} \textbf{0.19}} & {\color[HTML]{D83931} \textbf{96.38}} \\
\rowcolor[gray]{0.95}
 & SD3 (2B) & 24.92 & 0.71 & 0.30 & 0.22 & 126.53 \\
\rowcolor[gray]{0.95}
 & FLUX2-Klein (4B) & {\color[HTML]{4E83FD} \textbf{25.78}} & {\color[HTML]{D83931} \textbf{0.75}} & {\color[HTML]{4E83FD} \textbf{0.26}} & {\color[HTML]{4E83FD} \textbf{0.20}} & 106.22 \\
\rowcolor[gray]{0.95}
 & Z-Image (6B) & 25.44 & {\color[HTML]{4E83FD} \textbf{0.73}} & 0.27 & 0.21 & {\color[HTML]{4E83FD} \textbf{104.37}} \\
\rowcolor[gray]{0.95}
\multirow{-8}{*}{RealSR} & FLUX2-Klein (9B) & {\color[HTML]{D83931} \textbf{26.12}} & {\color[HTML]{D83931} \textbf{0.75}} & {\color[HTML]{D83931} \textbf{0.25}} & {\color[HTML]{D83931} \textbf{0.18}} & {\color[HTML]{D83931} \textbf{98.30}} \\ \midrule
 & SD3 (2B) & 23.90 & 0.60 & 0.31 & 0.21 & 31.83 \\
 & FLUX2-Klein (4B) & {\color[HTML]{4E83FD} \textbf{24.90}} & {\color[HTML]{4E83FD} \textbf{0.64}} & {\color[HTML]{4E83FD} \textbf{0.27}} & {\color[HTML]{D83931} \textbf{0.18}} & {\color[HTML]{4E83FD} \textbf{22.90}} \\
 & Z-Image (6B) & 23.99 & 0.62 & 0.30 & {\color[HTML]{4E83FD} \textbf{0.20}} & 28.55 \\
 & FLUX2-Klein (9B) & {\color[HTML]{D83931} \textbf{25.18}} & {\color[HTML]{D83931} \textbf{0.65}} & {\color[HTML]{D83931} \textbf{0.26}} & {\color[HTML]{D83931} \textbf{0.18}} & {\color[HTML]{D83931} \textbf{21.67}} \\
\rowcolor[gray]{0.95}
 & SD3 (2B) & 23.13 & 0.58 & 0.32 & 0.20 & 27.80 \\
\rowcolor[gray]{0.95}
 & FLUX2-Klein (4B) & 23.63 & {\color[HTML]{4E83FD} \textbf{0.60}} & {\color[HTML]{4E83FD} \textbf{0.26}} & {\color[HTML]{4E83FD} \textbf{0.17}} & {\color[HTML]{4E83FD} \textbf{19.47}} \\
\rowcolor[gray]{0.95}
 & Z-Image (6B) & {\color[HTML]{4E83FD} \textbf{23.65}} & {\color[HTML]{4E83FD} \textbf{0.60}} & 0.28 & 0.19 & 22.96 \\
\rowcolor[gray]{0.95}
\multirow{-8}{*}{DIV2K-Val} & FLUX2-Klein (9B) & {\color[HTML]{D83931} \textbf{23.97}} & {\color[HTML]{D83931} \textbf{0.62}} & {\color[HTML]{D83931} \textbf{0.25}} & {\color[HTML]{D83931} \textbf{0.16}} & {\color[HTML]{D83931} \textbf{17.93}} \\ \midrule
 & SD3 (2B) & 20.22 & {\color[HTML]{4E83FD} \textbf{0.52}} & 0.25 & 0.15 & 63.84 \\
 & FLUX2-Klein (4B) & {\color[HTML]{4E83FD} \textbf{21.35}} & {\color[HTML]{D83931} \textbf{0.57}} & {\color[HTML]{4E83FD} \textbf{0.23}} & {\color[HTML]{4E83FD} \textbf{0.14}} & {\color[HTML]{4E83FD} \textbf{41.69}} \\
 & Z-Image (6B) & 20.26 & {\color[HTML]{4E83FD} \textbf{0.52}} & {\color[HTML]{4E83FD} \textbf{0.23}} & {\color[HTML]{4E83FD} \textbf{0.14}} & 54.33 \\
 & FLUX2-Klein (9B) & {\color[HTML]{D83931} \textbf{21.60}} & {\color[HTML]{D83931} \textbf{0.57}} & {\color[HTML]{D83931} \textbf{0.21}} & {\color[HTML]{D83931} \textbf{0.13}} & {\color[HTML]{D83931} \textbf{34.43}} \\
\rowcolor[gray]{0.95}
 & SD3 (2B) & 19.93 & 0.51 & 0.26 & 0.16 & 60.61 \\
\rowcolor[gray]{0.95}
 & FLUX2-Klein (4B) & {\color[HTML]{4E83FD} \textbf{20.64}} & {\color[HTML]{D83931} \textbf{0.55}} & {\color[HTML]{4E83FD} \textbf{0.21}} & {\color[HTML]{4E83FD} \textbf{0.14}} & {\color[HTML]{4E83FD} \textbf{43.56}} \\
\rowcolor[gray]{0.95}
 & Z-Image (6B) & 20.41 & {\color[HTML]{4E83FD} \textbf{0.53}} & 0.24 & 0.15 & 51.57 \\
\rowcolor[gray]{0.95}
\multirow{-8}{*}{LSDIR-Val} & FLUX2-Klein (9B) & {\color[HTML]{D83931} \textbf{20.75}} & {\color[HTML]{D83931} \textbf{0.55}} & {\color[HTML]{D83931} \textbf{0.20}} & {\color[HTML]{D83931} \textbf{0.13}} & {\color[HTML]{D83931} \textbf{38.34}} \\ \bottomrule
\end{tabular}%
}
\end{table}

\begin{figure}[t]
\centering
\includegraphics[width=\columnwidth]{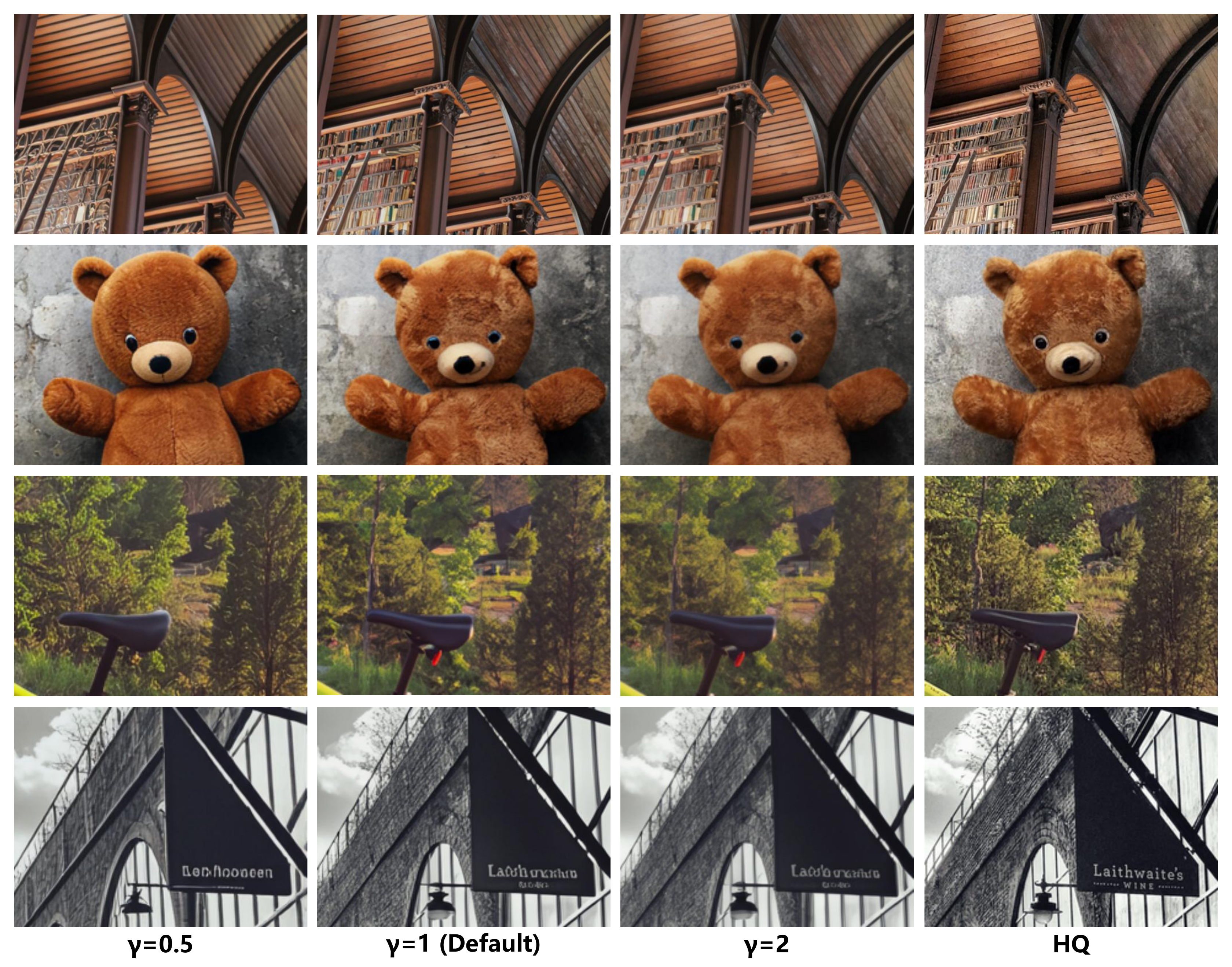}
\caption{Visual ablation on the residual ratio $\gamma$. The residual ratio controls the balance between deterministic LQ guidance and stochastic restoration, leading to different fidelity--realism trade-offs.}
\Description{Visual comparison of ScaleResfusion restoration results under different residual ratios.}
\label{fig:ablation_rsr}
\end{figure}
\begin{table}[t]\tiny
\centering
\caption{Ablation on the residual ratio $\gamma$. We evaluate three residual ratios under two backbones on DRealSR and LSDIR-Val.}
\label{tab:ablation_rsr}
\setlength{\tabcolsep}{3.5pt}
\renewcommand{\arraystretch}{1.05}
\resizebox{\linewidth}{!}{%
\begin{tabular}{@{}l|l|c|cccc@{}}
\toprule
\textbf{Dataset} & \textbf{Backbone} & \textbf{$\gamma$} & \textbf{PSNR $\uparrow$} & \textbf{SSIM $\uparrow$} & \textbf{LPIPS $\downarrow$} & \textbf{FID $\downarrow$} \\
\midrule
 & & 2.0 & 29.53 & 0.81 & 0.27 & 139.44 \\
 & & 1.0 & 28.77 & 0.79 & 0.30 & 149.67 \\
 \multirow{-3}{*}{DRealSR} & \multirow{-3}{*}{SD3} & 0.5 & 28.64 & 0.78 & 0.30 & 148.19 \\
\midrule
 & & 2.0 & 30.49 & 0.83 & 0.26 & 120.03 \\
 & & 1.0 & 29.77 & 0.82 & 0.25 & 118.18 \\
 \multirow{-3}{*}{DRealSR} & \multirow{-3}{*}{FLUX2-4B} & 0.5 & 29.18 & 0.79 & 0.28 & 132.45 \\
\midrule
 & & 2.0 & 20.48 & 0.52 & 0.22 & 53.61 \\
 & & 1.0 & 20.22 & 0.52 & 0.25 & 63.84 \\
 \multirow{-3}{*}{LSDIR-Val} & \multirow{-3}{*}{SD3} & 0.5 & 19.59 & 0.49 & 0.25 & 62.36 \\
\midrule
 & & 2.0 & 22.07 & 0.58 & 0.24 & 43.54 \\
 & & 1.0 & 21.35 & 0.57 & 0.23 & 41.69 \\
 \multirow{-3}{*}{LSDIR-Val} & \multirow{-3}{*}{FLUX2-4B} & 0.5 & 20.76 & 0.54 & 0.26 & 55.96 \\
\bottomrule
\end{tabular}%
}
\end{table}
\subsection{Ablation on Residual Ratio}
\label{sec:ablation_on_rsr}
We further ablate the residual ratio $\gamma$ introduced in Sec.~\ref{sec: residual rectified flow}. As discussed in Sec.~\ref{sec: residual rectified flow}, $\gamma$ directly determines the signal-to-noise ratio (SNR) of the RRF acceleration point: a larger $\gamma$ makes the initial state closer to the LQ-related residual signal, while a smaller $\gamma$ injects stronger Gaussian randomness. This parameter therefore controls how much the sampling process relies on deterministic LQ initialization versus stochastic generative refinement.

Table~\ref{tab:ablation_rsr} shows that increasing $\gamma$ generally improves fidelity-oriented metrics. For example, moving from $\gamma=0.5$ to $\gamma=2.0$ consistently improves PSNR and SSIM on both DRealSR and LSDIR-Val, suggesting that a stronger residual signal helps preserve the observed image structure. The visual comparison in Fig.~\ref{fig:ablation_rsr} follows the same trend: small $\gamma$ values leave more room for stochastic restoration but can introduce unstable details, while larger $\gamma$ values raise the SNR of the starting state and therefore provide stronger structural guidance.

However, the largest $\gamma$ is not uniformly the best choice, and the optimum is backbone-dependent. On SD3, $\gamma=2.0$ improves or matches all four metrics, whereas on the stronger FLUX2-4B backbone it sacrifices LPIPS and FID on both datasets despite the PSNR/SSIM gains, suggesting that once the generative prior is strong enough, excessive reliance on the LQ-related signal limits realistic detail synthesis and distribution alignment. Since FLUX2-4B is our default configuration, we adopt $\gamma=1.0$, which offers the best perception--distortion balance for the default model while retaining enough stochasticity for realistic restoration.
\begin{figure}[t]
\centering
\includegraphics[width=\columnwidth]{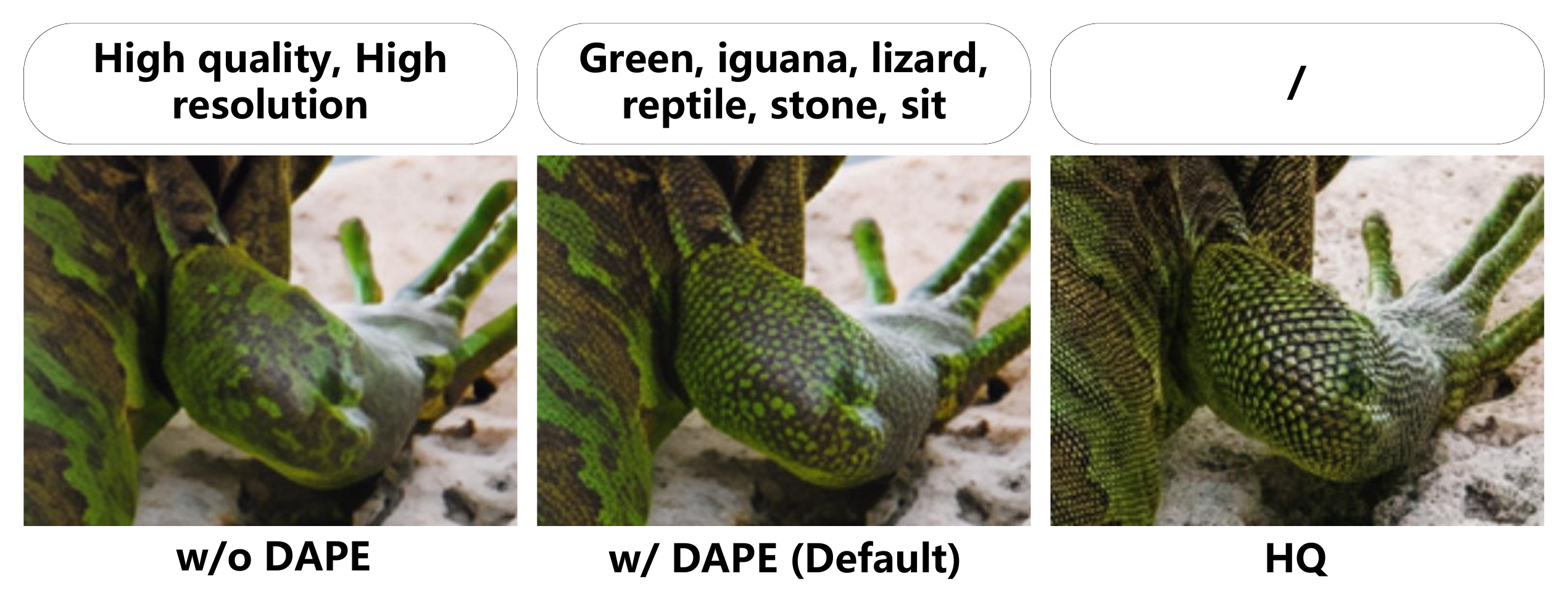}
\caption{Visual ablation on DAPE. Replacing DAPE with a fixed prompt simplifies the pipeline but loses image-specific semantic and degradation-aware guidance, while DAPE better preserves structures and degradation-specific details.}
\Description{Visual comparison between fixed-prompt conditioning and DAPE, showing the effect of image-specific textual guidance.}
\label{fig:ablation_dape}
\end{figure}
\begin{table}[t]\tiny
\centering
\caption{Ablation on replacing DAPE with a fixed prompt using the FLUX2-4B backbone on DRealSR and LSDIR-Val.}
\label{tab:ablation_dape}
\setlength{\tabcolsep}{3.8pt}
\renewcommand{\arraystretch}{1.05}
\resizebox{\linewidth}{!}{%
\begin{tabular}{@{}l|l|cccc@{}}
\toprule
\textbf{Dataset} & \textbf{Variant} & \textbf{PSNR $\uparrow$} & \textbf{SSIM $\uparrow$} & \textbf{LPIPS $\downarrow$} & \textbf{FID $\downarrow$} \\
\midrule
 & Fixed prompt & 28.39 & 0.78 & 0.28 & 130.98 \\
 \multirow{-2}{*}{DRealSR} & DAPE & 29.77 & 0.82 & 0.25 & 118.18 \\
\midrule
 & Fixed prompt & 20.83 & 0.55 & 0.23 & 47.95 \\
 \multirow{-2}{*}{LSDIR-Val} & DAPE & 21.35 & 0.57 & 0.23 & 41.69 \\
\bottomrule
\end{tabular}%
}
\end{table}
\begin{figure*}[t]
\centering
\includegraphics[width=0.95\textwidth]{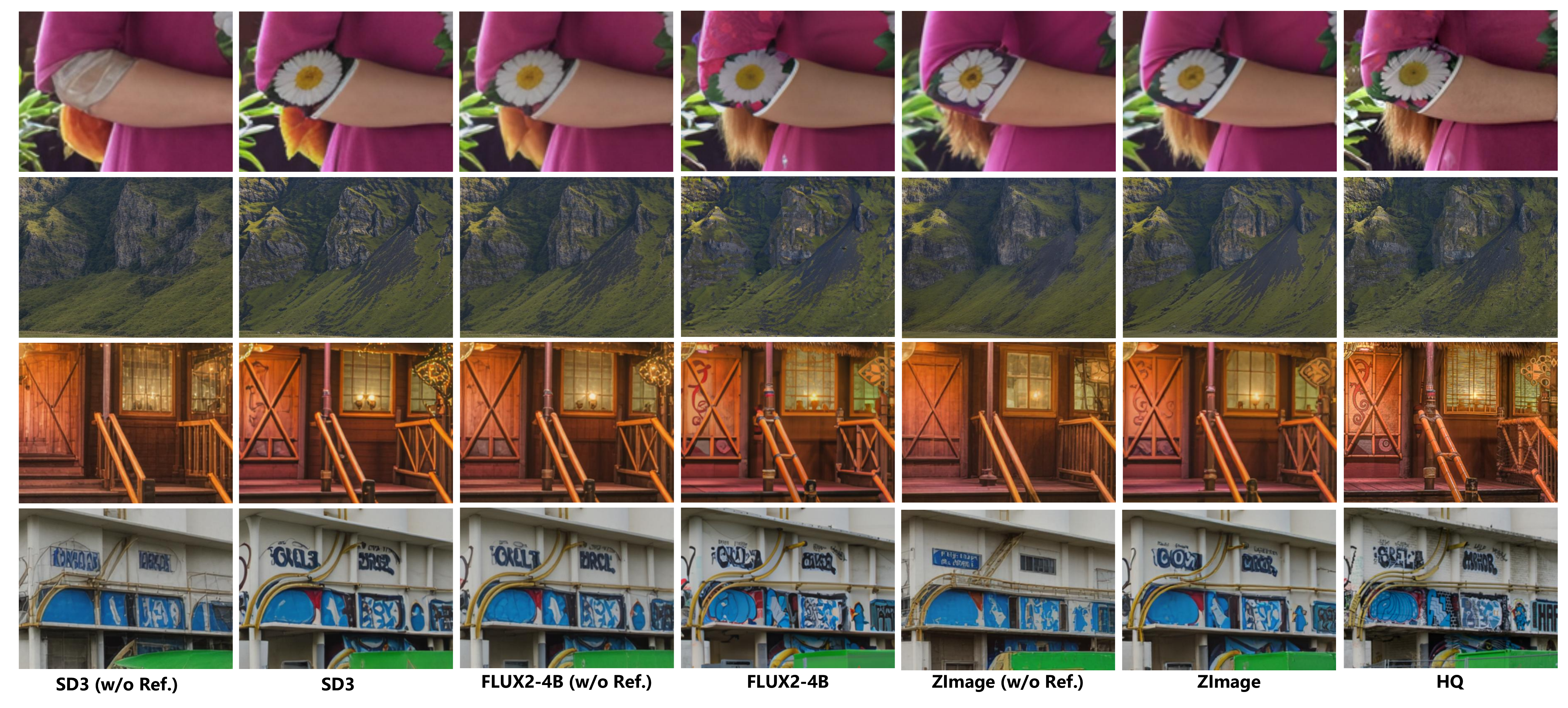}
\caption{Visual ablation on LQ feature conditioning. Removing LQ conditioning allows the model to synthesize more diverse details, but it also weakens input consistency and reconstruction fidelity.}
\Description{Visual comparison of restoration results with and without LQ feature conditioning, showing the trade-off between detail diversity and input fidelity.}
\label{fig:ablation_nolq}
\end{figure*}
\subsection{Ablation on DAPE}
\label{sec:ablation_on_dape}
We analyze whether the image-specific semantic information provided by DAPE can be replaced by a fixed restoration prompt, which would remove the extra DAPE module and simplify inference. This is a practical trade-off: the LQ conditioning branch already conveys the image content to the generator, so image-specific textual guidance may be partially redundant.

Table~\ref{tab:ablation_dape} and Fig.~\ref{fig:ablation_dape} show that the fixed-prompt variant is not invalid; it still produces reasonable restoration results under the same FLUX2-4B backbone. However, DAPE consistently improves PSNR, SSIM, LPIPS, and FID on DRealSR, and also improves PSNR, SSIM, and FID on LSDIR-Val. The visual comparison further suggests that image-specific textual guidance helps recover more faithful structures and degradation-specific details. Therefore, replacing DAPE with a fixed prompt is a viable lightweight option when simplicity is preferred, but the default DAPE setting offers a better quality--robustness trade-off.
\begin{table}[t]\tiny
\centering
\caption{Ablation on the feature extractor with the FLUX2-4B backbone on DRealSR and LSDIR-Val.}
\label{tab:ablation_feature_extractor}
\setlength{\tabcolsep}{3.8pt}
\renewcommand{\arraystretch}{1.05}
\resizebox{\linewidth}{!}{%
\begin{tabular}{@{}l|l|cccc@{}}
\toprule
\textbf{Dataset} & \textbf{Variant} & \textbf{PSNR $\uparrow$} & \textbf{SSIM $\uparrow$} & \textbf{LPIPS $\downarrow$} & \textbf{FID $\downarrow$} \\
\midrule
 & w/o Feature Extractor & 28.93 & 0.79 & 0.28 & 129.86 \\
 \multirow{-2}{*}{DRealSR} & w/ Feature Extractor & 29.77 & 0.82 & 0.25 & 118.18 \\
\midrule
 & w/o Feature Extractor & 20.51 & 0.54 & 0.26 & 53.37 \\
 \multirow{-2}{*}{LSDIR-Val} & w/ Feature Extractor & 21.35 & 0.57 & 0.23 & 41.69 \\
\bottomrule
\end{tabular}%
}
\end{table}
\subsection{Ablation on Feature Extractor}
\label{sec:ablation_on_feature_extractor}

\textbf{Effect of LQ Feature Conditioning.}
We compare ScaleResfusion with and without the feature extractor. The feature extractor provides explicit LQ feature injection as an additional conditioning signal, helping the generator align restored structures with the degraded input. Removing this conditioning gives the generative prior more freedom to synthesize diverse details, but it also weakens the constraint from the observation, which may reduce fidelity and input consistency.

Table~\ref{tab:ablation_feature_extractor} shows this trade-off quantitatively. Adding the feature extractor improves PSNR from 28.93 to 29.77 and SSIM from 0.79 to 0.82 on DRealSR, and improves PSNR from 20.51 to 21.35 and SSIM from 0.54 to 0.57 on LSDIR-Val. The same setting also reduces LPIPS and FID on both datasets, suggesting that LQ feature conditioning improves not only distortion fidelity but also perceptual and distribution-level quality.

The visual comparison in Fig.~\ref{fig:ablation_nolq} further illustrates the role of the feature extractor. Without LQ conditioning, the model can generate richer or more varied local details, but these details are less tightly anchored to the input image and can deviate from the original structure. With feature extraction, ScaleResfusion better preserves the observed content while still benefiting from the generative prior, leading to a more reliable fidelity--diversity trade-off.

\textbf{Layers of Feature Extractor.}
We further ablate the number of LoRA-branch feature extractor layers on SD3 to study how much LQ information should be injected into the residual generator. As shown in Table~\ref{tab:ablation_feature_layers}, using more feature extractor layers consistently improves restoration quality. Increasing the number of layers from 4 to 23 improves PSNR/SSIM and reduces LPIPS/FID on both DRealSR and LSDIR-Val, indicating that deeper feature extraction provides more complete conditioning.

This trend is consistent with the role of LQ feature injection. Shallow feature settings mainly provide coarse structural cues, which help anchor the global layout but are less effective at transferring fine textures, local degradation patterns, and high-frequency details. Deeper settings expose the residual generator to richer multi-level LQ representations, allowing it to better preserve input content while producing natural HQ details. We therefore use the full 23-layer feature extractor for SD3, as it provides the strongest and most complete conditioning signal.
\begin{table}[t]\tiny
\centering
\caption{Ablation on the number of feature extractor layers with SD3 backbone on DRealSR and LSDIR-Val.}
\label{tab:ablation_feature_layers}
\setlength{\tabcolsep}{4.0pt}
\renewcommand{\arraystretch}{1.05}
\resizebox{\linewidth}{!}{%
\begin{tabular}{@{}l|c|cccc@{}}
\toprule
\textbf{Dataset} & \textbf{\# Layers} & \textbf{PSNR $\uparrow$} & \textbf{SSIM $\uparrow$} & \textbf{LPIPS $\downarrow$} & \textbf{FID $\downarrow$} \\
\midrule
 & 23 & 28.77 & 0.79 & 0.30 & 149.67 \\
 & 16 & 28.41 & 0.78 & 0.31 & 153.82 \\
 & 8 & 28.02 & 0.76 & 0.33 & 161.35 \\
 \multirow{-4}{*}{DRealSR} & 4 & 27.68 & 0.75 & 0.35 & 168.90 \\
\midrule
 & 23 & 20.22 & 0.52 & 0.25 & 63.84 \\
 & 16 & 19.36 & 0.49 & 0.26 & 67.99 \\
 & 8 & 18.97 & 0.47 & 0.28 & 75.52 \\
 \multirow{-4}{*}{LSDIR-Val} & 4 & 18.63 & 0.46 & 0.30 & 83.07 \\
\bottomrule
\end{tabular}%
}
\end{table}

\section{Theoretical Justification of Residual Rectified Flow}
\label{sec:general_residual_flow_derivation}

We justify Residual Rectified Flow (RRF) by following the marginal-preserving argument of nonlinear Rectified Flow~\cite{liu2022flow}. Different from the canonical Rectified Flow, which constructs a linear interpolation between two endpoint distributions, our residual construction defines an observation-dependent stochastic path whose intermediate state admits an exact acceleration point. Therefore, RRF inherits the marginal-preserving property of nonlinear Rectified Flow, while its acceleration point comes from clean-image coefficient cancellation in the residual parameterization.

Let $X_0\in\mathbb R^d$ denote the clean high-quality image and let $C=\hat X_0\in\mathbb R^d$ denote the image-space condition obtained from the low-quality observation, such as an upsampled or preliminary restored image. Let $\epsilon\sim\mathcal N(0,\Sigma)$ be independent Gaussian noise. For a fixed residual strength $\gamma>0$, define the residual endpoint
\begin{equation}
Y_{\gamma}
=
\gamma(C-X_0)+\epsilon .
\end{equation}
We then construct the residual interpolation
\begin{equation}
X_t=(1-t)X_0+tY_{\gamma},
\qquad t\in[0,1].
\end{equation}
Equivalently,
\begin{equation}
X_t
=
(1-t)X_0+t\gamma(C-X_0)+t\epsilon ,
\end{equation}
and hence
\begin{equation}
X_t
=
[1-(1+\gamma)t]X_0+\gamma t C+t\epsilon .
\end{equation}
The pathwise velocity is
\begin{equation}
\dot X_t
=
Y_{\gamma}-X_0
=
\gamma(C-X_0)+\epsilon-X_0 .
\end{equation}
We denote $U_t:=\dot X_t$.

Following Definition~3.1 of original Rectified Flow~\cite{liu2022flow}, which defines the expected velocity as $v^X(x,t)=\mathbb E[\dot X_t\mid X_t=x]$ for a pathwise differentiable process, we define the conditional expected velocity by
\begin{equation}
v^X(x,t,c)
=
\mathbb E[\dot X_t\mid X_t=x,C=c].
\end{equation}
In practice, $v^X$ is approximated by a neural network $v_\theta(x,t,c)$. Following Eq.~(6) of original Rectified Flow~\cite{liu2022flow}, which fits the nonlinear Rectified Flow velocity by matching $v(X_t,t)$ to $\dot X_t$, we use the least-squares objective
\begin{equation}
\min_\theta
\int_0^1
\mathbb E
\left[
\left\|
v_\theta(X_t,t,C)-\dot X_t
\right\|_2^2
\right]dt .
\end{equation}
Here we take the weight $w_t=1$ in Eq.~(6) of original Rectified Flow~\cite{liu2022flow} and add the condition $C$ as input. Equivalently,
\begin{equation}
\min_\theta
\int_0^1
\mathbb E
\left[
\left\|
v_\theta(X_t,t,C)
-
\bigl(\gamma(C-X_0)+\epsilon-X_0\bigr)
\right\|_2^2
\right]dt .
\end{equation}

\paragraph{\textbf{Definition 1: Conditional residual rectified flow.}}
For each fixed condition $C=c$, define
\begin{equation}
X_t^c
=
[1-(1+\gamma)t]X_0+\gamma t c+t\epsilon .
\end{equation}
Its expected velocity is
\begin{equation}
v^X(x,t,c)
=
\mathbb E[\dot X_t\mid X_t=x,C=c].
\end{equation}
We call $X=\{X_t:t\in[0,1]\}$ conditionally rectifiable if $v^X(\cdot,t,c)$ is locally bounded and the integral equation
\begin{equation}
Z_t
=
Z_0+
\int_0^t
v^X(Z_s,s,c)\,ds
\end{equation}
admits a unique solution for each fixed $c$.

Following Eq.~(9) of original Rectified Flow~\cite{liu2022flow}, where the rectified flow is defined by the integral equation driven by $v^X$, the above equation is its conditional counterpart after fixing $C=c$.

\paragraph{\textbf{Theorem 1: Conditional marginal preservation.}}
Assume that, for each fixed $C=c$, the residual interpolation $X_t$ is conditionally rectifiable. Let $Z_t$ solve
\begin{equation}
dZ_t
=
v^X(Z_t,t,c)\,dt,
\qquad
Z_0\mid C=c\sim X_0\mid C=c .
\end{equation}
Then
\begin{equation}
\operatorname{Law}(Z_t\mid C=c)
=
\operatorname{Law}(X_t\mid C=c),
\qquad
\forall t\in[0,1].
\end{equation}

\emph{Proof.}
Fix $C=c$. The proof is the conditional version of Theorem~3.3 in~\cite{liu2022flow}. Let $\pi_t^c:=\operatorname{Law}(X_t\mid C=c)$. Following Eq.~(10) of original Rectified Flow~\cite{liu2022flow}, for any compactly supported continuously differentiable test function $h:\mathbb R^d\to\mathbb R$, the chain rule gives
\begin{equation}
\frac{d}{dt}
\mathbb E[h(X_t)\mid C=c]
=
\mathbb E[
\nabla h(X_t)^\top \dot X_t
\mid C=c
].
\end{equation}
Following the conditional expected-velocity substitution in Eq.~(10) of original Rectified Flow~\cite{liu2022flow}, we use
\begin{equation}
v^X(X_t,t,c)
=
\mathbb E[\dot X_t\mid X_t,C=c],
\end{equation}
and obtain
\begin{equation}
\mathbb E[
\nabla h(X_t)^\top \dot X_t
\mid C=c
]
=
\mathbb E[
\nabla h(X_t)^\top v^X(X_t,t,c)
\mid C=c
].
\end{equation}
This is Eq.~(10) of original Rectified Flow~\cite{liu2022flow} with the unconditional velocity $v^X(X_t,t)$ replaced by the conditional velocity $v^X(X_t,t,c)$.
Therefore,
\begin{equation}
\frac{d}{dt}
\mathbb E[h(X_t)\mid C=c]
=
\mathbb E[
\nabla h(X_t)^\top v^X(X_t,t,c)
\mid C=c
].
\end{equation}
Following Eq.~(11) of original Rectified Flow~\cite{liu2022flow}, the above weak identity is equivalently written as the continuity equation for $\pi_t^c$ in the sense of distributions:
\begin{equation}
\dot\pi_t^c
+
\nabla\cdot
\left(
v^X(\cdot,t,c)\pi_t^c
\right)
=
0 .
\end{equation}
Following the paragraph below Eq.~(11) of original Rectified Flow~\cite{liu2022flow}, this weak equivalence follows by multiplying the continuity equation by $h$ and integrating by parts. Since $Z_t$ is driven by the same velocity field $v^X(\cdot,t,c)$, its conditional law $q_t^c:=\operatorname{Law}(Z_t\mid C=c)$ also satisfies
\begin{equation}
\dot q_t^c
+
\nabla\cdot
\left(
v^X(\cdot,t,c)q_t^c
\right)
=
0 .
\end{equation}
Moreover, $q_0^c=\pi_0^c$ by the initialization assumption. By the uniqueness of the weak solution to the continuity equation, we obtain
\begin{equation}
q_t^c=\pi_t^c,
\qquad
\forall t\in[0,1].
\end{equation}
Thus,
\begin{equation}
\operatorname{Law}(Z_t\mid C=c)
=
\operatorname{Law}(X_t\mid C=c).
\end{equation}
This proves the claim.

Following Theorem~3.3 of original Rectified Flow~\cite{liu2022flow}, the stochastic interpolation $X_t$ and the deterministic ODE $Z_t$ solve the same continuity equation in Eq.~(11), so their one-time marginals coincide.

\paragraph{\textbf{Theorem 2: Exact acceleration point.}}
For the residual interpolation
\begin{equation}
X_t
=
[1-(1+\gamma)t]X_0+\gamma t C+t\epsilon ,
\end{equation}
there exists a unique time
\begin{equation}
t^\star=\frac{1}{1+\gamma}
\end{equation}
at which the clean-image component $X_0$ is exactly cancelled. At this time,
\begin{equation}
X_{t^\star}
=
\frac{\gamma}{1+\gamma}C
+
\frac{1}{1+\gamma}\epsilon .
\end{equation}

\emph{Proof.}
The coefficient of $X_0$ in $X_t$ is $1-(1+\gamma)t$. Setting this coefficient to zero gives
\begin{equation}
1-(1+\gamma)t^\star=0 .
\end{equation}
Therefore,
\begin{equation}
t^\star=\frac{1}{1+\gamma}.
\end{equation}
Substituting $t^\star$ into the residual interpolation gives
\begin{equation}
X_{t^\star}
=
[1-(1+\gamma)t^\star]X_0
+
\gamma t^\star C
+
t^\star\epsilon .
\end{equation}
Since $1-(1+\gamma)t^\star=0$, we have
\begin{equation}
X_{t^\star}
=
\gamma t^\star C+t^\star\epsilon .
\end{equation}
Using $t^\star=1/(1+\gamma)$, we obtain
\begin{equation}
X_{t^\star}
=
\frac{\gamma}{1+\gamma}C
+
\frac{1}{1+\gamma}\epsilon .
\end{equation}
Thus, $X_{t^\star}$ depends only on the condition $C$ and Gaussian noise $\epsilon$, but not on the unknown clean image $X_0$.

If $\epsilon\sim\mathcal N(0,\Sigma)$, then
\begin{equation}
X_{t^\star}\mid C=c
\sim
\mathcal N
\left(
\frac{\gamma}{1+\gamma}c,
\frac{1}{(1+\gamma)^2}\Sigma
\right).
\end{equation}
Hence, the reverse ODE can be initialized from
\begin{equation}
Z_{t^\star}
=
\frac{\gamma}{1+\gamma}C
+
\frac{1}{1+\gamma}\epsilon ,
\qquad
\epsilon\sim\mathcal N(0,\Sigma).
\end{equation}
By Theorem 1, under the population velocity field and exact ODE integration,
\begin{equation}
Z_0\mid C=c
\sim
X_0\mid C=c .
\end{equation}

\paragraph{\textbf{Training objective of RRF.}}
Since
\begin{equation}
\dot X_t
=
\gamma(C-X_0)+\epsilon-X_0,
\end{equation}
the population RRF objective is obtained by following Eq.~(6) of original Rectified Flow~\cite{liu2022flow} and substituting the residual path velocity $\dot X_t=\gamma(C-X_0)+\epsilon-X_0$:
\begin{equation}
\small
\mathcal L_{\mathrm{RRF}}(\theta)
=
\int_0^1
\mathbb E
\left[
\left\|
v_\theta(X_t,t,C)
-
\left(
\gamma(C-X_0)+\epsilon-X_0
\right)
\right\|_2^2
\right]dt .
\end{equation}
Equivalently, if training is restricted to the accelerated interval $[0,t^\star]$, one may use
\begin{equation}
\small
\mathcal L_{\mathrm{RRF}}(\theta)
=
\int_0^{t^\star}
\mathbb E
\left[
\left\|
v_\theta(X_t,t,C)
-
\left(
\gamma(C-X_0)+\epsilon-X_0
\right)
\right\|_2^2
\right]dt .
\end{equation}
Following Eq.~(2) of original Rectified Flow~\cite{liu2022flow}, where the canonical population minimizer is $v^X(x,t)=\mathbb E[X_1-X_0\mid X_t=x]$, the RRF population minimizer is
\begin{equation}
v^\star(x,t,c)
=
\mathbb E[
\gamma(C-X_0)+\epsilon-X_0
\mid X_t=x,C=c
].
\end{equation}

\paragraph{\textbf{Relation to Rectified Flow.}}
RRF preserves the linear transport form of standard Rectified Flow.
Recall that canonical Rectified Flow constructs the straight interpolation
\begin{equation}
X_t=(1-t)X_0+tX_1
\end{equation}
and learns the velocity field associated with the path derivative
$X_1-X_0$. In RRF, we replace the standard endpoint $X_1$ with the
residual endpoint
\begin{equation}
Y_\gamma=\gamma(C-X_0)+\epsilon,
\end{equation}
which gives
\begin{equation}
X_t=(1-t)X_0+tY_\gamma .
\end{equation}
Therefore, conditioned on $C$, RRF is still a linear interpolation between
two endpoints, and its pathwise velocity is
\begin{equation}
\dot X_t=Y_\gamma-X_0
=\gamma(C-X_0)+\epsilon-X_0 .
\end{equation}

Following the Rectified Flow argument, the population velocity field is
given by the conditional expectation
\begin{equation}
v^\star(x,t,c)
=
\mathbb{E}\!\left[
Y_\gamma-X_0
\,\middle|\,
X_t=x, C=c
\right].
\end{equation}
At the population level, with this exact velocity field and exact ODE
integration, the induced flow preserves the conditional interpolation
marginals:
\begin{equation}
\mathrm{Law}(Z_t\mid C=c)
=
\mathrm{Law}(X_t\mid C=c),
\qquad t\in[0,1].
\end{equation}
Thus, RRF inherits the marginal-preserving property of Rectified Flow while
introducing an observation-aware residual endpoint.

The residual endpoint further yields an exact acceleration point. Since
\begin{equation}
X_t=[1-(1+\gamma)t]X_0+\gamma t C+t\epsilon ,
\end{equation}
the coefficient of the unknown clean image $X_0$ vanishes at
\begin{equation}
t^\star=\frac{1}{1+\gamma}.
\end{equation}
Hence,
\begin{equation}
X_{t^\star}
=
\frac{\gamma}{1+\gamma}C
+
\frac{1}{1+\gamma}\epsilon ,
\end{equation}
which depends only on the observed condition $C$ and Gaussian noise. This
allows RRF to start sampling from a noisy LQ-related state and integrate
only over the shortened interval $[0,t^\star]$.

Consequently, the RRF training objective is obtained by substituting the
residual path velocity into the standard Rectified Flow objective:
\begin{equation}
\small
\mathcal{L}_{\mathrm{RRF}}(\theta)
=
\int_{0}^{t^\star}
\mathbb{E}
\left[
\left\|
v_\theta(X_t,t,C)
-
\left(\gamma(C-X_0)+\epsilon-X_0\right)
\right\|_2^2
\right]dt .
\end{equation}
In this sense, RRF keeps the linear interpolation and marginal-preserving
structure of Rectified Flow, but shifts the endpoint from pure Gaussian
noise to a residual-aware stochastic endpoint. This endpoint change is what
enables noisy-LQ initialization and accelerated sampling.

\section{More Visual Results}
\label{sec:appendix_more_visual_results}
We provide additional visual comparisons to complement the quantitative results. These examples cover real-world benchmarks, synthetic benchmarks, and challenging face restoration cases, showing the behavior of ScaleResfusion under diverse degradation patterns. Overall, ScaleResfusion better preserves input structures while recovering realistic local details, and it avoids the over-smoothed textures, semantic drift, or hallucinated artifacts that can appear in competing restoration methods.
\newpage
\begin{figure*}[t]
\center
\includegraphics[width=0.95\textwidth]{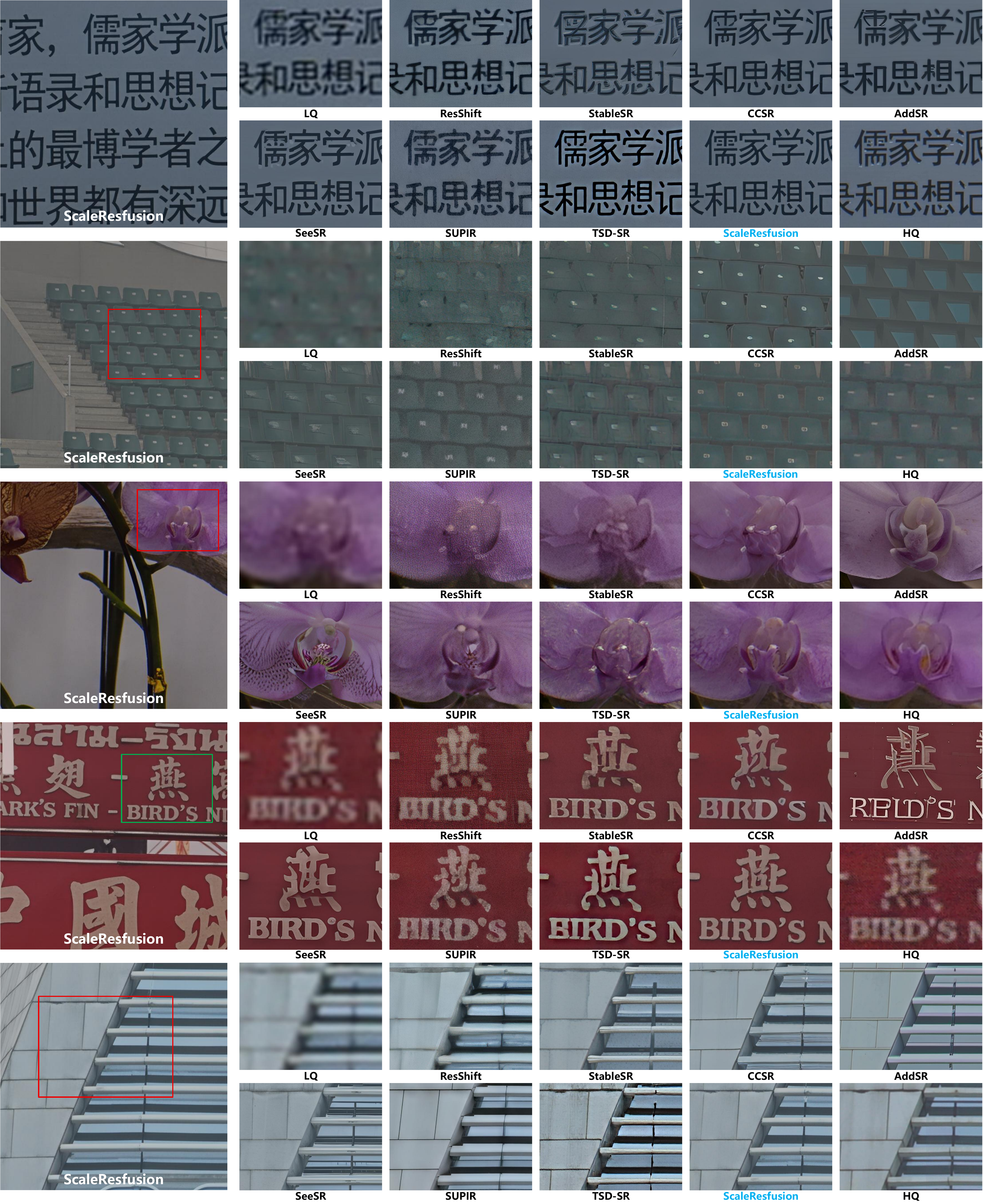}
\caption{More visual comparison on DRealSR with existing diffusion-based restoration methods. The left side shows the results of ScaleResfusion, while the right side compares local crops. Overall, ScaleResfusion better preserves the global structure while recovering more natural details without the over-smoothing or hallucinated patterns observed in competing methods.}
\label{fig:dreal_1}
\end{figure*}

\begin{figure*}[t]
\center
\includegraphics[width=0.95\textwidth]{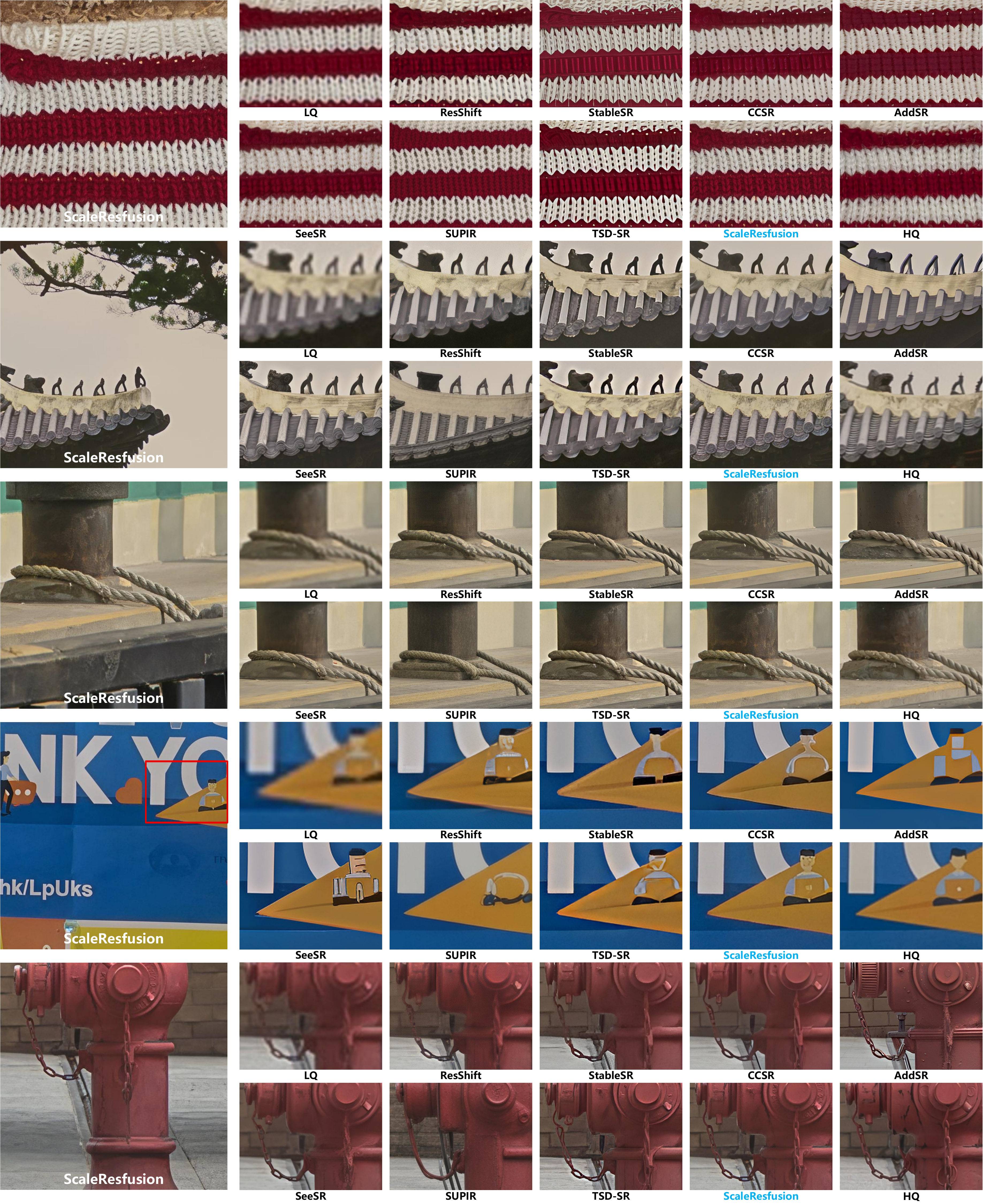}
\caption{More visual comparison on RealSR with existing diffusion-based restoration methods. The left side shows the results of ScaleResfusion, while the right side compares local crops. Overall, ScaleResfusion better preserves the global structure while recovering more natural details without the over-smoothing or hallucinated patterns observed in competing methods.}
\label{fig:Rreal_1}
\end{figure*}

\begin{figure*}[t]
\center
\includegraphics[width=0.95\textwidth]{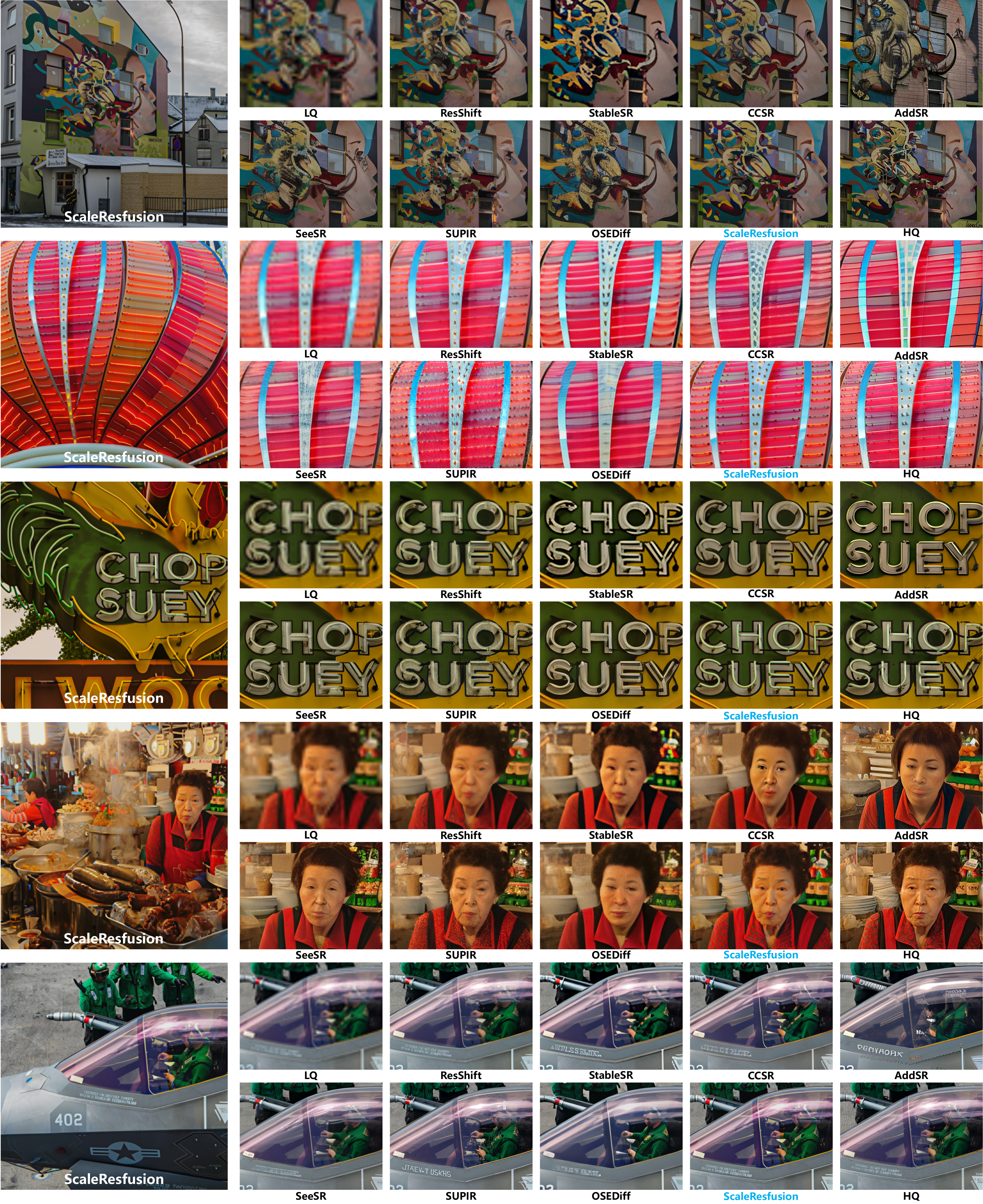}
\caption{More visual comparison with existing diffusion-based restoration methods. The left side shows the results of ScaleResfusion, while the right side compares local crops. Overall, ScaleResfusion better preserves the global structure while recovering more natural details without the over-smoothing or hallucinated patterns observed in competing methods.}
\label{fig:lsdir_2}
\end{figure*}

\begin{figure*}[t]
\center
\includegraphics[width=\textwidth]{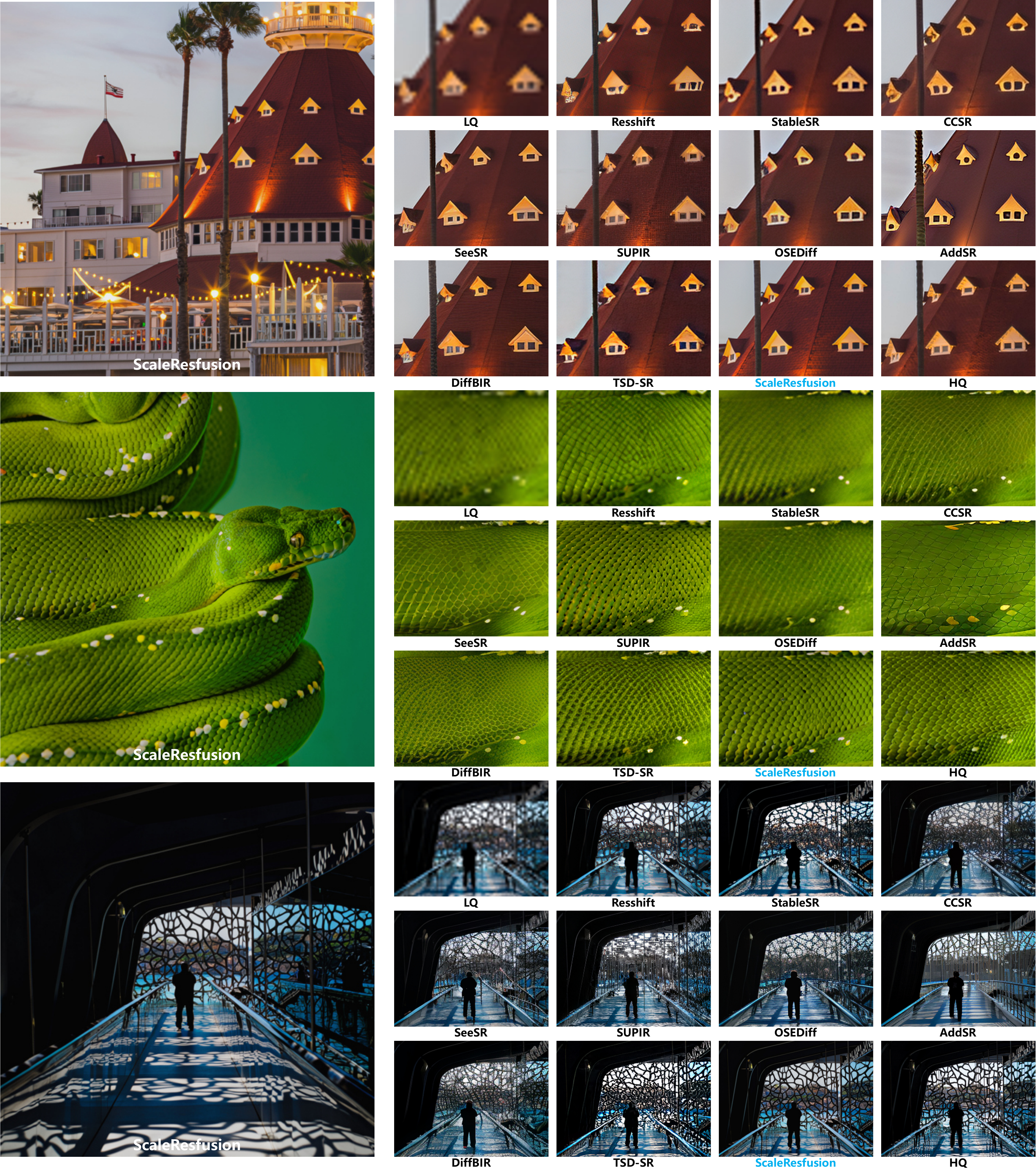}
\caption{More visual comparison on LSDIR-Val with existing diffusion-based restoration methods. The left side shows the results of ScaleResfusion, while the right side compares local crops. Overall, ScaleResfusion better preserves the global structure while recovering more natural details without the over-smoothing or hallucinated patterns observed in competing methods.}
\label{fig:lsdir_3}
\end{figure*}

\begin{figure*}[t]
\center
\includegraphics[width=\textwidth]{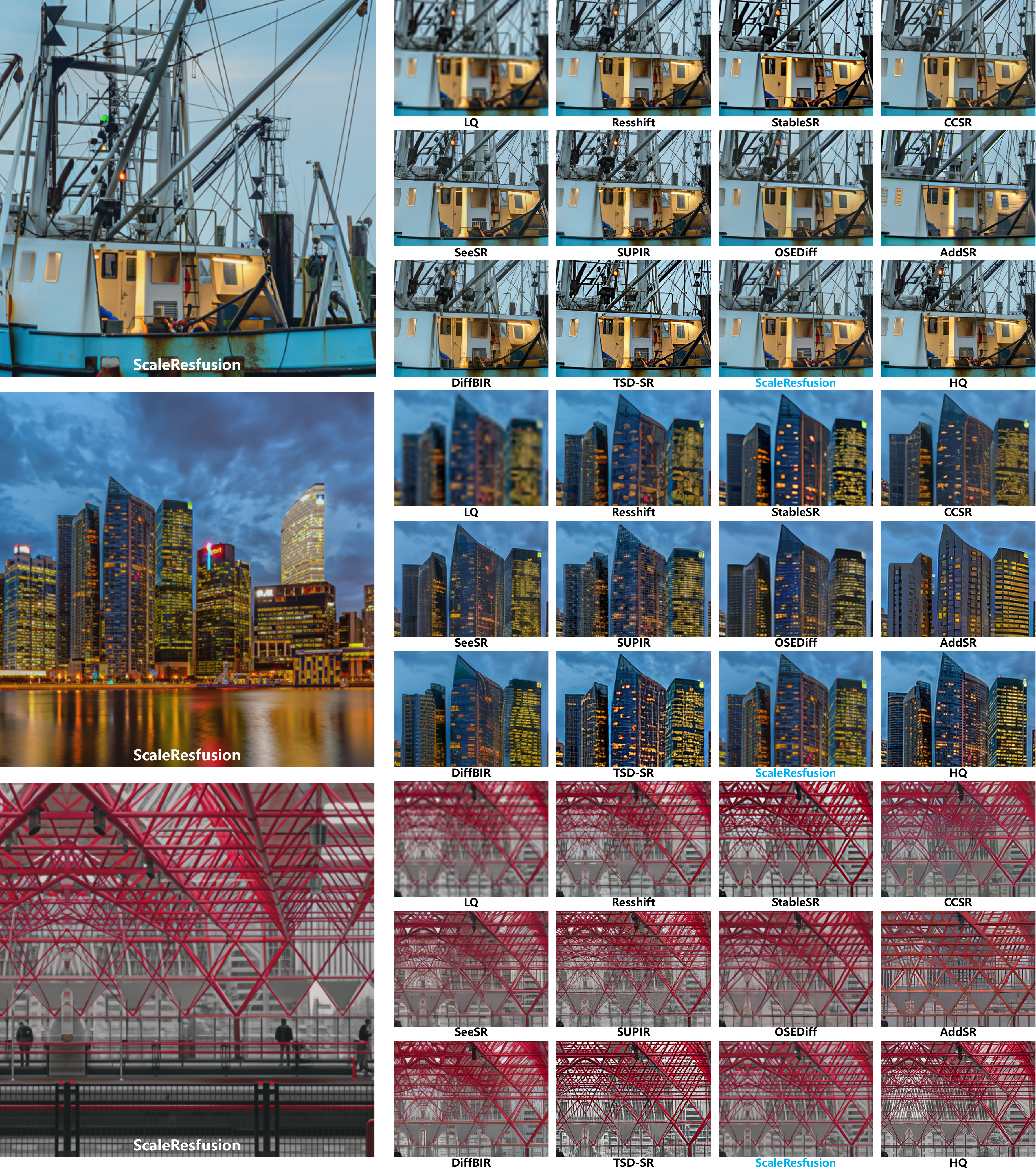}
\caption{More visual comparison on LSDIR-Val with existing diffusion-based restoration methods. The left side shows the results of ScaleResfusion, while the right side compares local crops. Overall, ScaleResfusion better preserves the global structure while recovering more natural details without the over-smoothing or hallucinated patterns observed in competing methods.}
\label{fig:lsdir_4}
\end{figure*}

\begin{figure*}[t]
\center
\includegraphics[width=\textwidth]{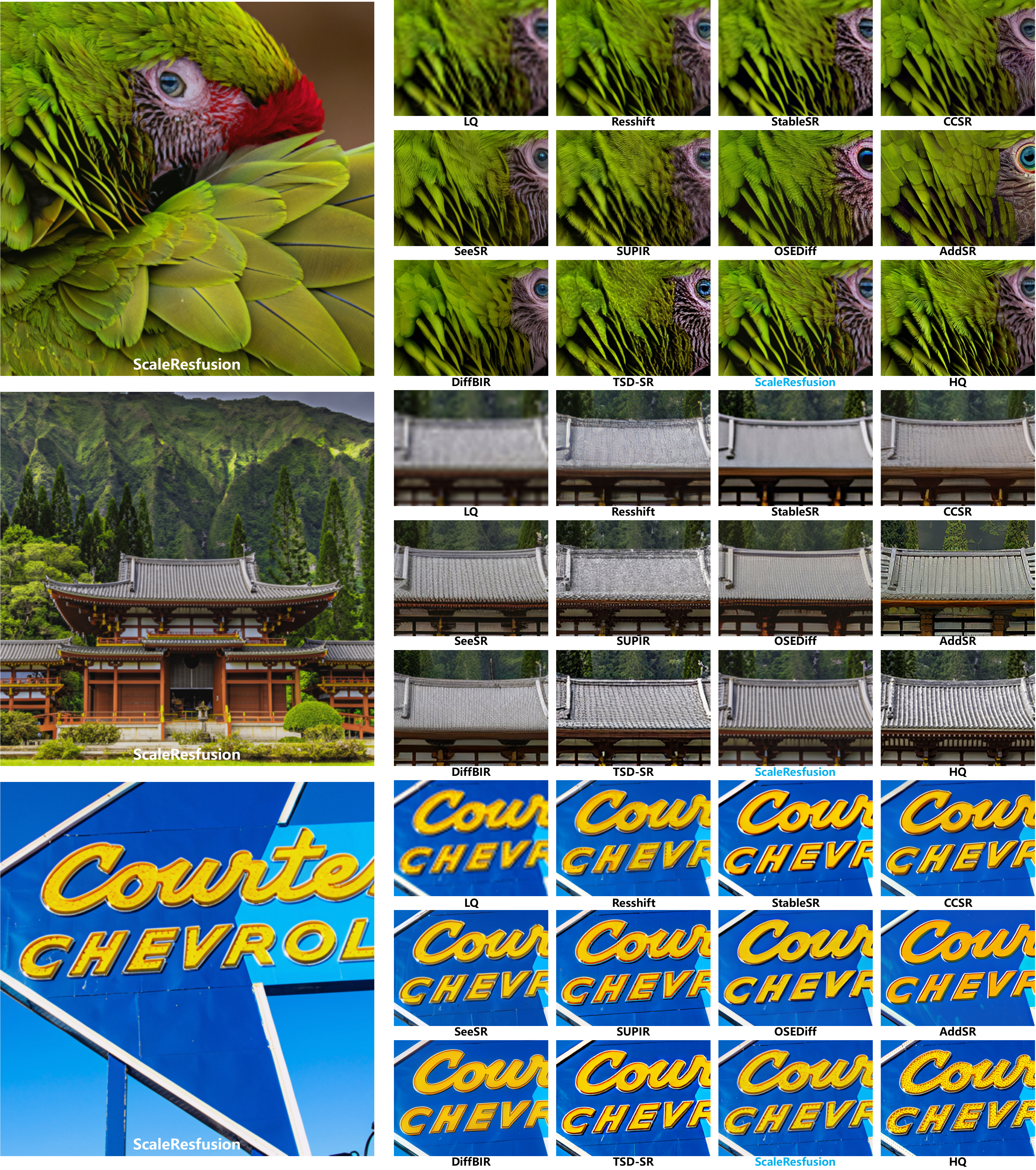}
\caption{More visual comparison on LSDIR-Val with existing diffusion-based restoration methods. The left side shows the results of ScaleResfusion, while the right side compares local crops. Overall, ScaleResfusion better preserves the global structure while recovering more natural details without the over-smoothing or hallucinated patterns observed in competing methods.}
\label{fig:lsdir_5}
\end{figure*}

\begin{figure*}[t]
\center
\includegraphics[width=\textwidth]{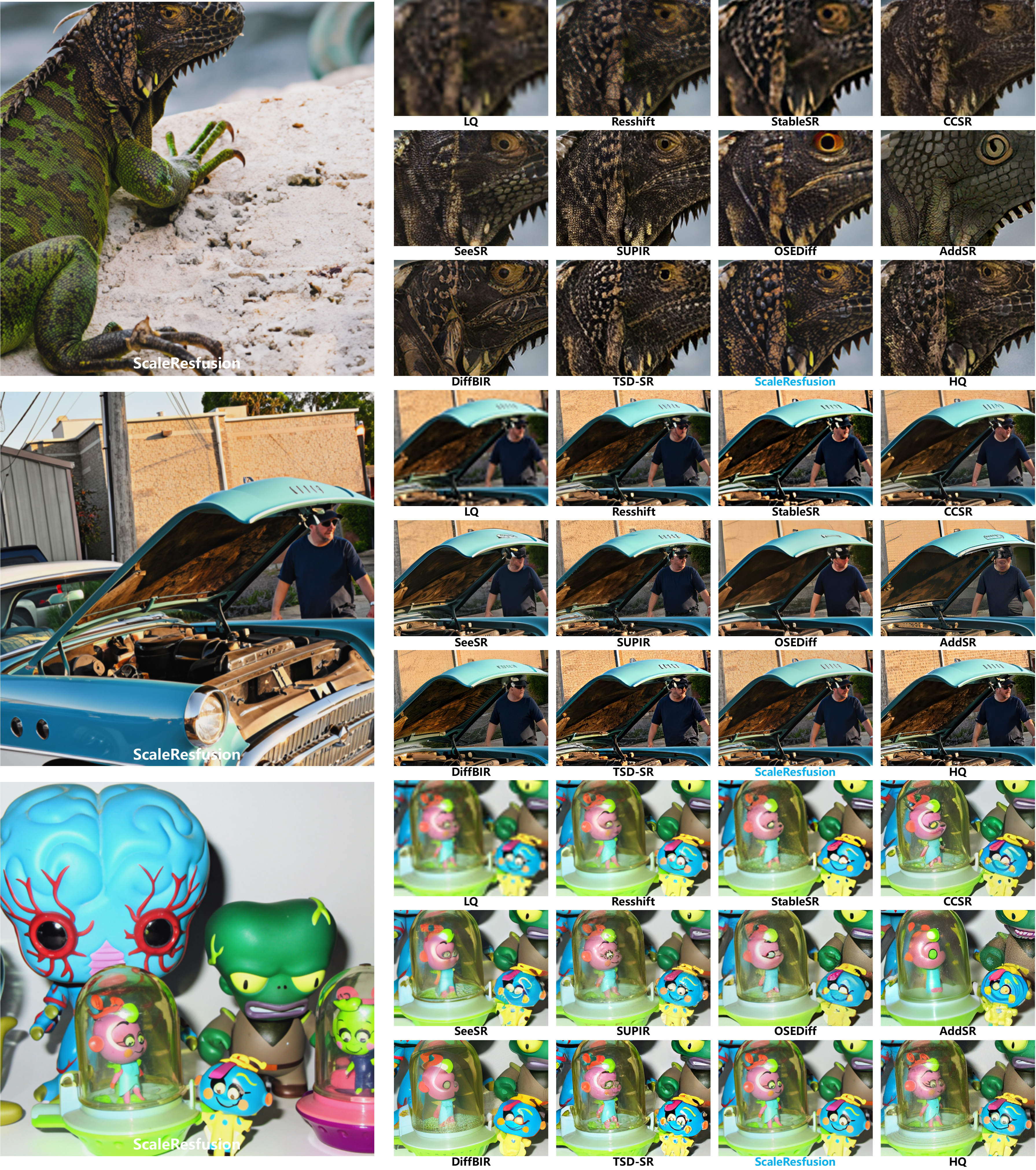}
\caption{More visual comparison on LSDIR-Val with existing diffusion-based restoration methods. The left side shows the results of ScaleResfusion, while the right side compares local crops. Overall, ScaleResfusion better preserves the global structure while recovering more natural details without the over-smoothing or hallucinated patterns observed in competing methods.}
\label{fig:lsdir_6}
\end{figure*}

\begin{figure*}[t]
\center
\includegraphics[width=\textwidth]{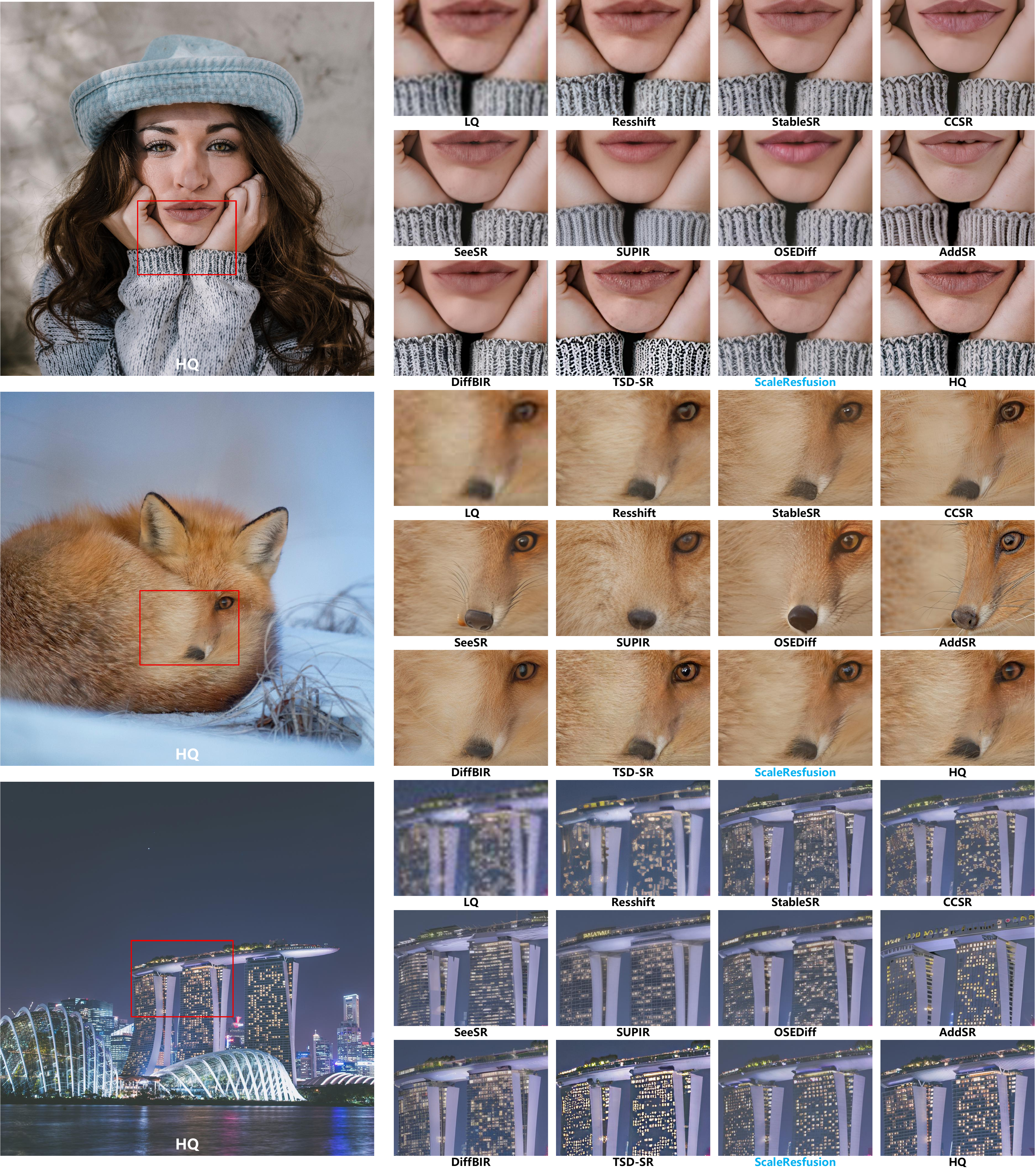}
\caption{More visual comparison on DIV2K-Val with existing diffusion-based restoration methods. The left side shows the HQ reference images, while the right side compares local crops. Overall, ScaleResfusion recovers more natural details while remaining faithful to the HQ references, avoiding the over-smoothing or hallucinated patterns observed in competing methods.}
\label{fig:dev2k_1}
\end{figure*}

\begin{figure*}[t]
\center
\includegraphics[width=\textwidth]{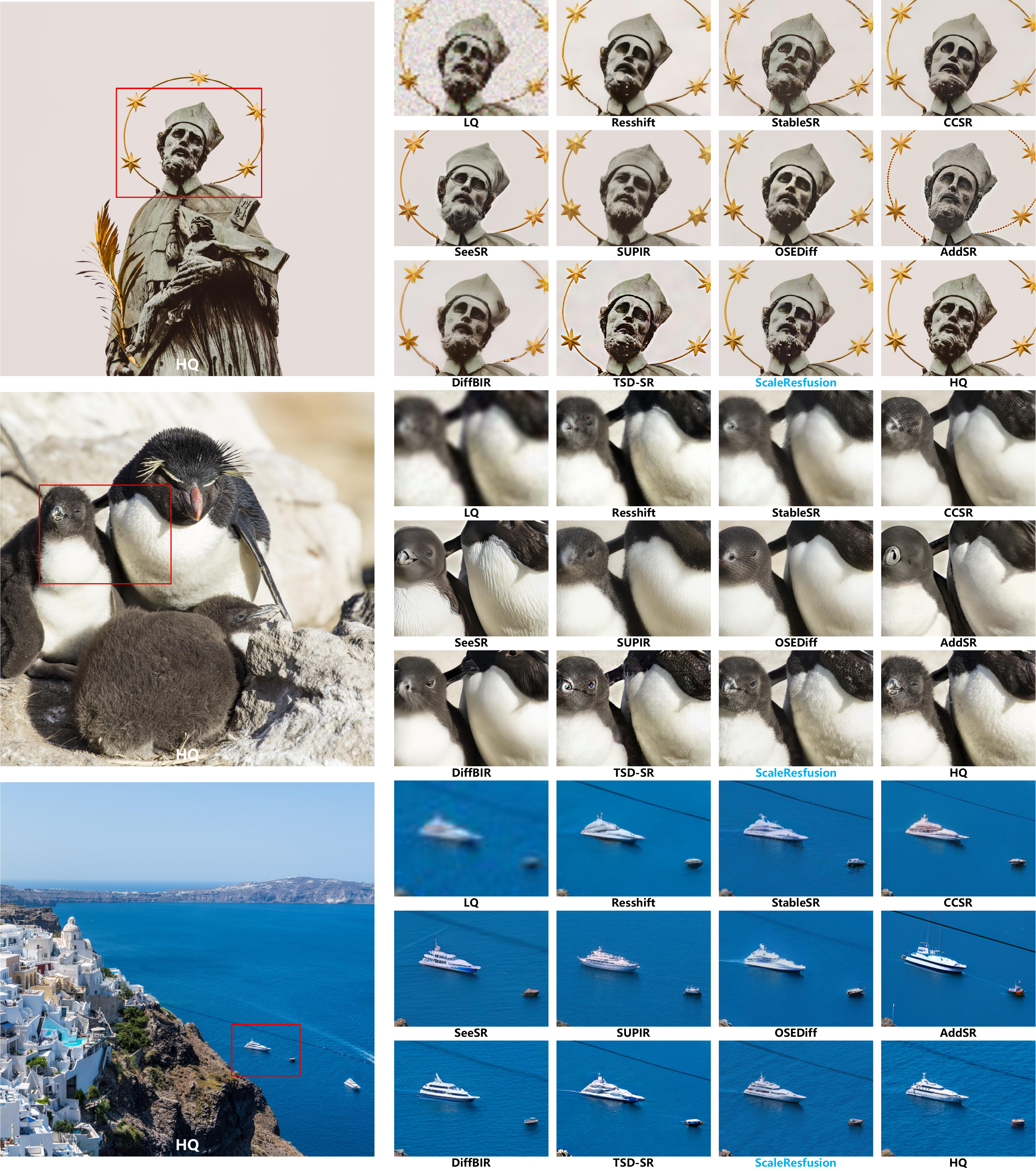}
\caption{More visual comparison on DIV2K-Val with existing diffusion-based restoration methods. The left side shows the HQ reference images, while the right side compares local crops. Overall, ScaleResfusion recovers more natural details while remaining faithful to the HQ references, avoiding the over-smoothing or hallucinated patterns observed in competing methods.}
\label{fig:div2k_2}
\end{figure*}

\begin{figure*}[t]
\center
\includegraphics[width=0.95\textwidth]{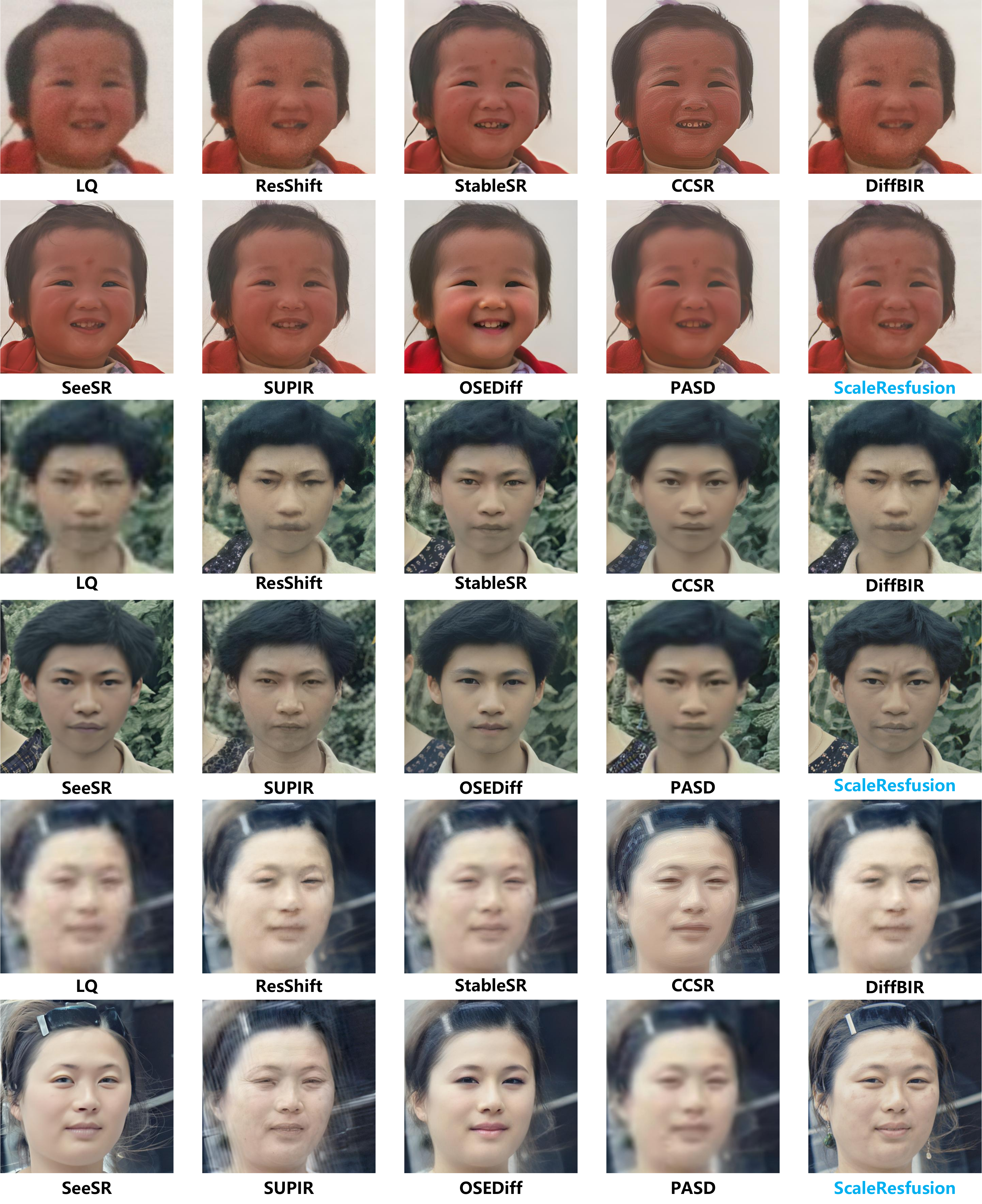}
\caption{Visual comparison on WebPhoto-Test~\cite{wang2021towards}, an in-the-wild face restoration benchmark without paired ground truth. In this zero-shot setting, ScaleResfusion better handles out-of-distribution degradations and restores more natural facial details than competing methods.}
\label{fig:face_1}
\end{figure*}

\begin{figure*}[t]
\center
\includegraphics[width=0.75\textwidth]{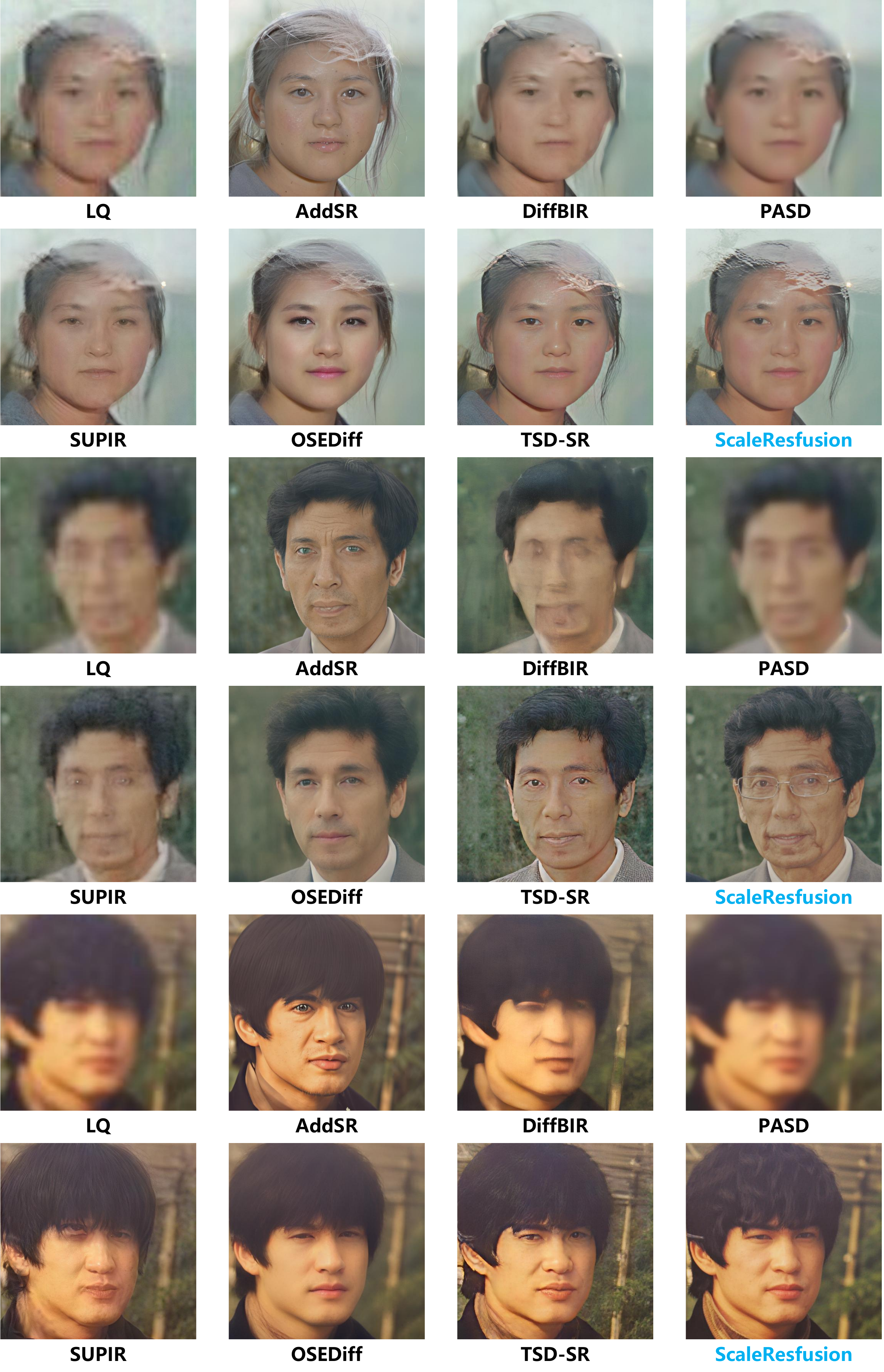}
\caption{More visual comparison on WebPhoto-Test~\cite{wang2021towards}, an in-the-wild face restoration benchmark without paired ground truth. In this zero-shot setting, ScaleResfusion better handles out-of-distribution degradations and restores more natural facial details than competing methods.}
\label{fig:face_2}
\end{figure*}

\begin{figure*}[t]
\center
\includegraphics[width=0.75\textwidth]{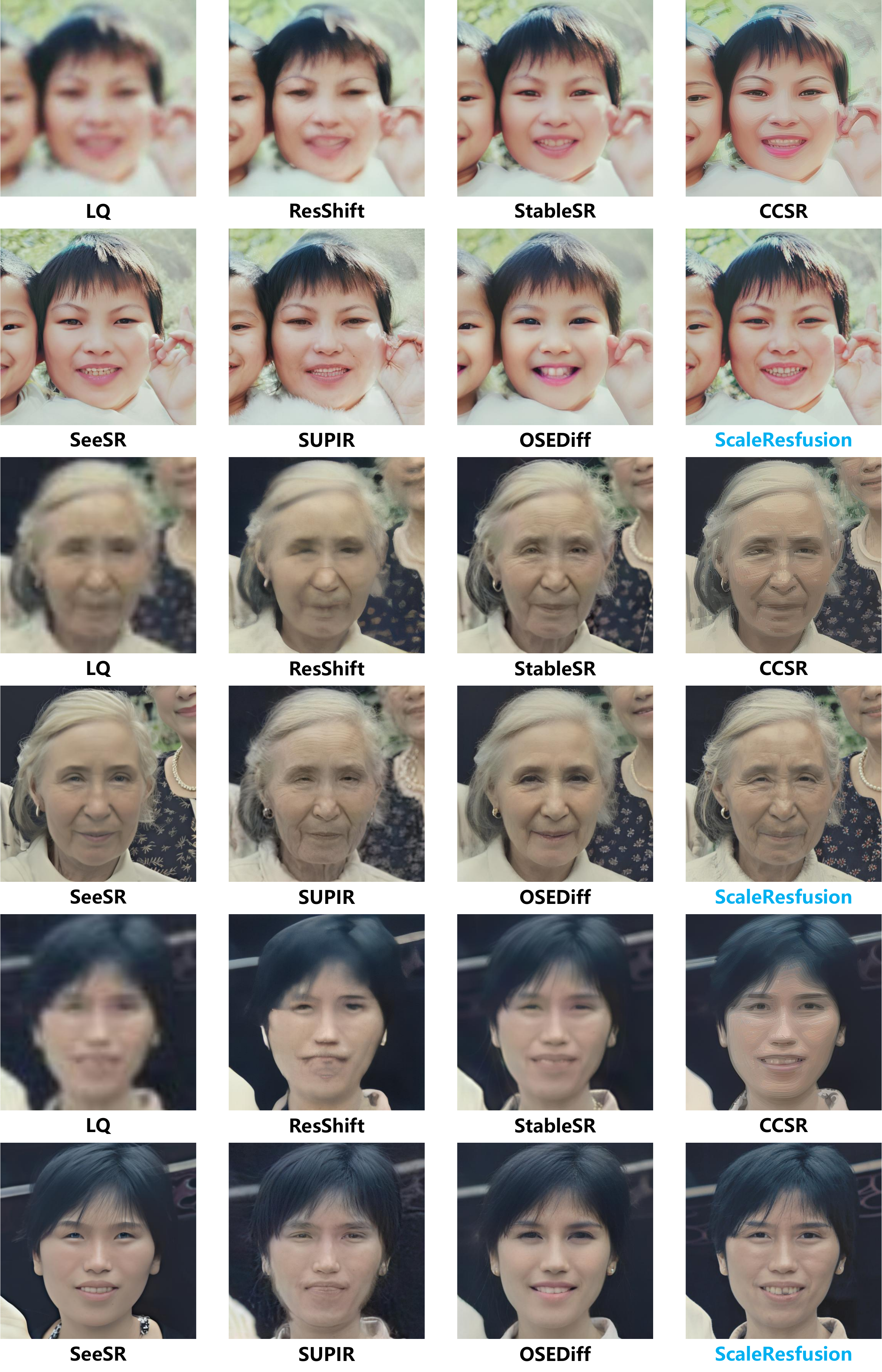}
\caption{More visual comparison on WebPhoto-Test~\cite{wang2021towards}, an in-the-wild face restoration benchmark without paired ground truth. In this zero-shot setting, ScaleResfusion better handles out-of-distribution degradations and restores more natural facial details than competing methods.}
\label{fig:face_3}
\end{figure*}

\begin{figure*}[t]
\center
\includegraphics[width=0.75\textwidth]{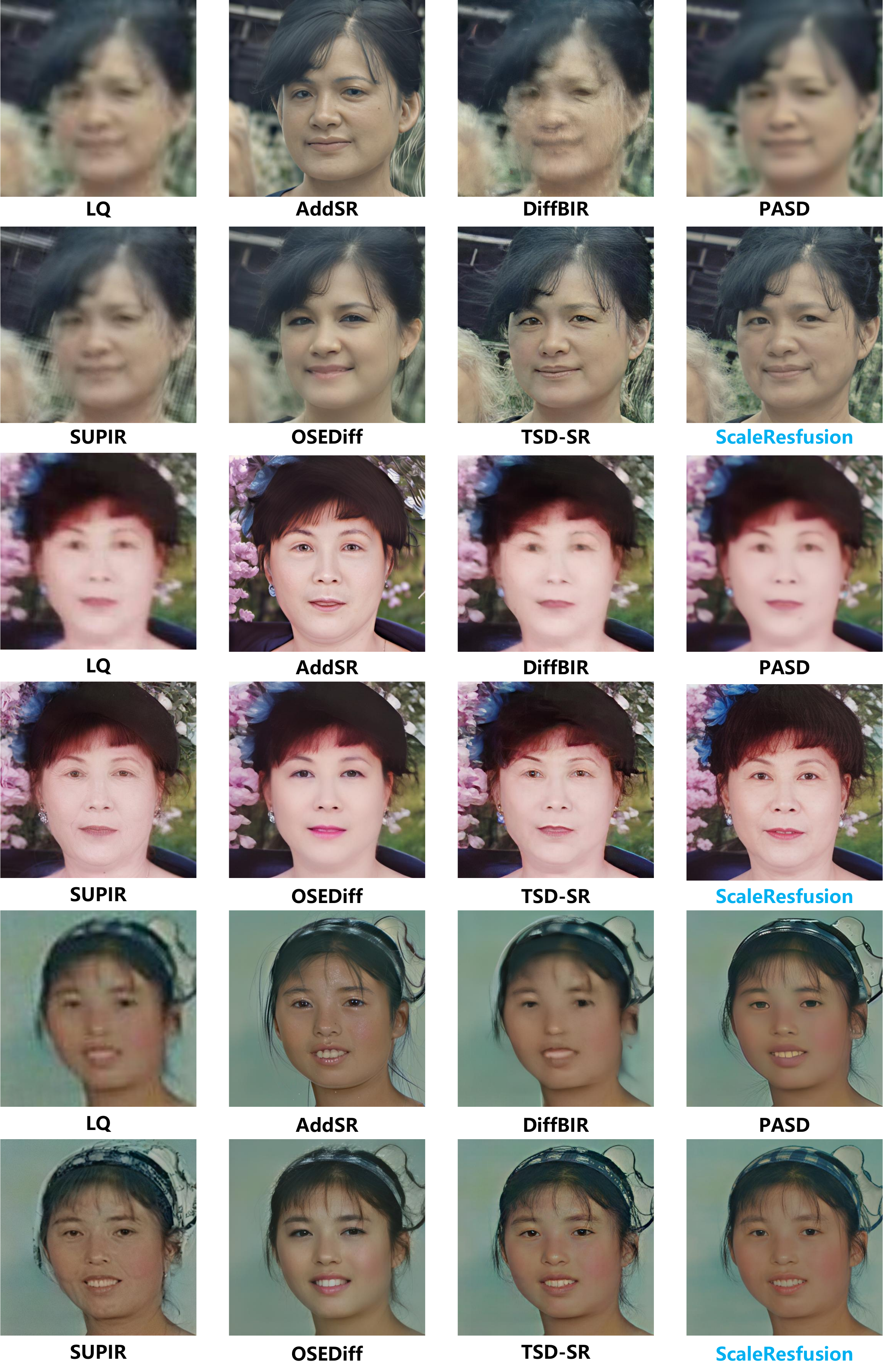}
\caption{More visual comparison on WebPhoto-Test~\cite{wang2021towards}, an in-the-wild face restoration benchmark without paired ground truth. In this zero-shot setting, ScaleResfusion better handles out-of-distribution degradations and restores more natural facial details than competing methods.}
\label{fig:face_4}
\end{figure*}

\begin{figure*}[t]
\center
\includegraphics[width=0.75\textwidth]{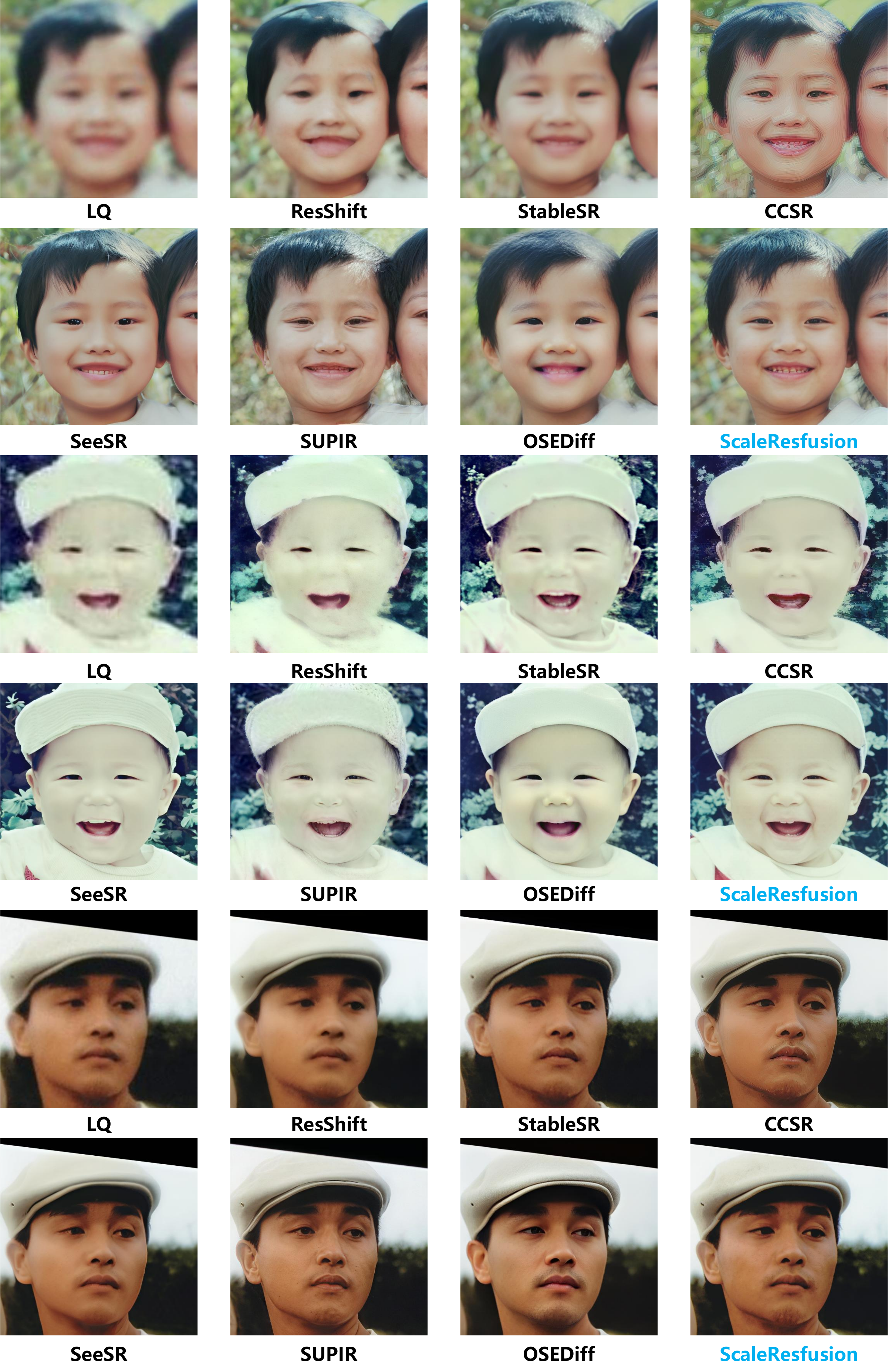}
\caption{More visual comparison on WebPhoto-Test~\cite{wang2021towards}, an in-the-wild face restoration benchmark without paired ground truth. In this zero-shot setting, ScaleResfusion better handles out-of-distribution degradations and restores more natural facial details than competing methods.}
\label{fig:face_5}
\end{figure*}

\end{document}